\pdfoutput=1
\documentclass[11pt]{article}

\usepackage[final]{acl}

\usepackage{times}
\usepackage{latexsym}
\usepackage{amsmath}
\usepackage{amssymb}
\usepackage{booktabs}
\usepackage{graphicx}
\usepackage[table,xcdraw]{xcolor}
\usepackage{hyperref}
\usepackage{ulem}

\usepackage[T1]{fontenc}
\usepackage[utf8]{inputenc}

\usepackage{microtype}

\usepackage{inconsolata}
\usepackage{multirow}

\usepackage{subcaption}

\usepackage{graphicx}

\usepackage{CJKutf8}
\usepackage[breakable,skins]{tcolorbox}
\usepackage{colortbl}
\usepackage[ruled,vlined]{algorithm2e}

\usepackage{listings}
\definecolor{codegreen}{rgb}{0,0.6,0}
\definecolor{codegray}{rgb}{0.5,0.5,0.5}
\definecolor{codepurple}{rgb}{0.58,0,0.82}
\definecolor{backcolour}{rgb}{0.98,0.98,0.98}

\lstdefinestyle{pythonstyle}{
    language=Python,
    backgroundcolor=\color{backcolour},   
    commentstyle=\color{codegreen}\itshape,
    keywordstyle=\color{blue}\bfseries,
    numberstyle=\tiny\color{codegray},
    stringstyle=\color{codepurple},
    basicstyle=\ttfamily\scriptsize, % ACL 双栏排版建议用 scriptsize 或 footnotesize
    breakatwhitespace=false,         
    breaklines=true,                 % 自动换行（关键）
    captionpos=b,                    % 标题在底部
    keepspaces=true,                 
    numbers=left,                    % 左侧显示行号
    numbersep=5pt,                  
    showspaces=false,                
    showstringspaces=false,
    showtabs=false,                  
    tabsize=4,
    extendedchars=false              % 避免部分中文字符导致编译报错
}

\newcommand{\bench}[1]{\textsc{HoWTo}\textsc{Bench}}
\newcommand{\redx}[1]{\textcolor{red}{X}}
\newcommand{\method}[1]{\textbf{ToW}}
\newcommand{\equalcontri}[1]{$^\ast$}
\newcommand{\one}[1]{$^\alpha$}
\newcommand{\two}[1]{$^\beta$}

\newtcolorbox[auto counter,number within=chapter,]{PromptBox}[3][]{
arc=5mm,
lower separated=false,
fonttitle=\bfseries,
colbacktitle=blue!10,
coltitle=blue!50!black,
enhanced,
attach boxed title to top left={xshift=0.5cm,
        yshift=-2mm},
colframe=blue!50!black,
colback=blue!10,
overlay={
\node[draw=blue!50!black,thick,
fill= blue!10,rounded corners=1mm, 
yshift=0pt, 
xshift=-0.5cm, 
left, 
text=blue!50!black,
anchor=east,
font=\bfseries] 
at (frame.north east) {#3};},
overlay={
\node[draw=blue!50!black,thick,
fill= blue!10,rounded corners=1mm, 
yshift=+1.2mm, %hier geaendert
xshift=-0.5cm, 
left, 
text=blue!50!black,
anchor=east,
font=\bfseries] 
at (frame.north east) {#3};},
title=#2 \thetcbcounter,#1,breakable}

\title{HoWToBench: \underline{Ho}listic Evaluation for LLM's Capability in Human-level \underline{W}riting using \underline{T}ree \underline{o}f Writing}
\author{
    Andrew Zhuoer Feng\equalcontri{}\one{}$^{\ddagger}$, 
    Cunxiang Wang\equalcontri{}\two{}\one{}$^{\dagger}$, 
    Yu Luo\one{}, 
    Lin Fan\two{},
    Yilin Zhou\two{}, \\
    \textbf{
    Zikang Wang\two{},
    Xiaotao Gu\two{},
    Jie Tang\one{},
    Hongning Wang\one{}, 
    Minlie Huang\one{}$^{\dagger}$
    } \\
    \one{}Department of Computer Science and Technology, Tsinghua University, \two{}Z.ai\\
    \texttt{\{fze22, aihuang\}@tsinghua.edu.cn} \quad \texttt{wangcunxiang303@gmail.com}
}

\begin{document}

\begin{CJK*}{UTF8}{gbsn}

\maketitle

{
    \let\thefootnote\relax\footnotetext{
    \begin{tabular}{@{}l@{\hspace{0.5em}}l@{}}
        \equalcontri{} & Equal contribution.  $^{\dagger}$  Corresponding authors.  \\
        $^{\ddagger}$ & Work done when A. Z. Feng interned at Z.ai. \\
    \end{tabular}
    }
}

\begin{abstract}
Evaluating the writing capabilities of large language models (LLMs) remains a significant challenge due to the multidimensional nature of writing skills and the limitations of existing metrics. LLM's performance in thousand-words level and open-ended writing is inadequately assessed by traditional reference-based metrics or modern LLM-as-a-judge methods. We propose Tree-of-Writing (\method{}), to resolve the implicit inconsistency often found when LLM-as-a-judge aggregates all sub-features in text evaluation. \method{} incorporates a tree-structured workflow by explicitly modeling the aggregation weights of sub-features. We also present~\bench{}, a large-scale Chinese writing benchmark encompassing $\mathbf{12}$ genres and $\mathbf{1302}$ instructions across three task categories: contextual \textbf{completion}, outline-\textbf{guided} writing, and \textbf{open}-ended generation. \method{} successfully mitigates the biases, achieving a $\mathbf{0.93}$ Pearson correlation with human judgments. Furthermore, we detect that both overlap-based text generation metrics and popular LLM-as-a-judge practices are vulnerable to textual disturbances, while~\method{} is robust to them. We also uncover a negative correlation between input length and content-related scores in the Guide task, showcasing that it cannot be simply improved by input-side information piling.

%Our results highlight the  and demonstrate \method{}'s effectiveness in aligning automated metrics with human judgment. This work advances the development of LLMs tailored for writing scenarios, as well as revealing the challenges in LLM writing as well as automated 
% This document is a supplement to the general instructions for *ACL authors. It contains instructions for using the \LaTeX{} style files for ACL conferences.
% The document itself conforms to its own specifications, and is therefore an example of what your manuscript should look like.
% These instructions should be used both for papers submitted for review and for final versions of accepted papers.
\end{abstract}

\section{Introduction}
\label{sec:intro}

The advances in large language models (LLMs)~\citep{ouyang2022training, rafailov2024dpo} have revolutionized the field of natural language processing, enabling breakthroughs in tasks like text summarization~\citep{basyal2023textsummarizationusinglarge}, machine translation~\citep{zhu-etal-2024-multilingual}, conversational agents~\citep{chatgpt3.5, chatglm, gemini1.5}, and creative writing~\citep{rocstories, writingprompts}. Despite their promising performance, auto-evaluating LLM-generated text remains a critical challenge particularly in complex, open-ended writing scenarios~\citep{longform, multimodallongstory, llmfictionworldview}.

The ability to generate nuanced and contextually appropriate writing depends heavily on handling implicit requirements, a challenge faced by both humans and LLMs. Existing evaluation methods for LLMs' writing skills predominantly focus on explicit instruction fulfillment~\citep{alignbench, kim2024biggenbenchprincipledbenchmark, zhu2023judgelmfinetunedlargelanguage, 2025wb}, i.e., whether the content meets the requirements.
However, this narrow focus, akin to a \textit{``mimicking game"}\label{tag:mimicgame}, overlooks LLMs' ability to craft complex, nuanced texts like fictional narratives or persuasive speeches where the intents behind the requirements are much more implicit but directly drive the requirement.

Current approaches \citep{kim2024prometheus, zhu2023judgelmfinetunedlargelanguage,2025wb} often rely on descriptions of evaluation criteria as instructions to the LLM-evaluator, requiring LLMs to provide sub-scores (e.g., fluency, consistency, instruction-following) leading to a final assessment. However, simply averaging the sub-scores is not necessarily an accurate reflection of overall quality, and LLM auto-planned negotiations between rubrics~\citep{2025wb} result in inconsistent and opaque assessment in multiple runs and queries. This misalignment with evaluation guidelines, which we term \textit{Negotiation Inconsistency}, results in unreliable and opaque assessments, undermining the credibility of LLM-as-a-judge in such tasks.

To address the challenge of \textit{Negotiation Inconsistency} in writing assessment, we propose the \textbf{Tree-of-Writing} (\method{}) framework, which simulates the human decision-making process. \method{} operates on a well-structured tree, which treats key evaluation aspects, such as language, logic, and plot, as leaf nodes. For each writing instruction, an LLM-negotiator designs the aggregation plan based on genre, task type and other requirements. Through a depth-first traversal of the plan, corresponding sub-score expert agents are activated to score each aspect. \method{} achieves a transparent and reproducible assessment for nuanced writings. 

Distinct from existing benchmarks~\citep{alignbench, zhu2023judgelmfinetunedlargelanguage, kim2024biggenbenchprincipledbenchmark, 2025wb} which all treat writing as a \textit{``mimicking game''}, we propose \bench{}, a large-scale benchmark designed to evaluate LLMs' writing abilities through three carefully designed task formats (\textbf{Completion, Guide} and \textbf{Open}), reflecting varying levels of provided context. \bench{} spans $\mathbf{12}$ genres with $\mathbf{1302}$ writing instructions, covering both creative and functional tasks. The dataset is curated from expert-written sources, highlighting the goal to emulate human-professional writing. The final pass rate for dataset quality check by human experts is $96.85\%$.
%\wcx{IAA reports.}

To validate the effectiveness of \method{}, we conducted large-scale evaluations on writings generated by 10 flagship LLMs, including Gemini-2.0-flash \citep{gemini-2.0}, GPT-4o-1120/o3-mini \citep{gpt-4o}, Claude-3.5-Sonnet~\citep{claude-3-5-sonnet} and DeepSeek-R1/V3 \citep{deepseek-R1,deepseek-v3}. Our framework demonstrates strong alignment with human preferences, achieving a Pearson correlation up to $\mathbf{0.93}$ when comparing system rankings with human-annotated rankings for all LLM-generated writings. 

Through our evaluation, we observed that some LLMs such as the GPT-series demonstrate strong performance in a rich-context setting (\textbf{Completion}) but drop drastically when the input information is limited. 
Analyzing all generated texts, we found a positive correlation between input and output length. However, it is noteworthy that longer inputs and outputs are associated with lower overall assessments, suggesting that the challenges of these tasks extend beyond simplistic length-based patterns. Furthermore, most metrics, including the use of LLMs as evaluators(``LLM-as-a-judge"), are susceptible to contextual fallacies, such as repetition, in certain styles. 
To the best of our knowledge, we are the first to explore the assessment of LLMs' capabilities in human-level writing with elaborately designed instructions beyond the instruction-following view. Data and code are available at \url{https://github.com/ZhuoerFeng/ACL2026-Tree-of-Writing}.

\section{Related Work}

\begin{table*}[!t]
\scriptsize
    \resizebox{\textwidth}{!}{%
    \begin{tabular}{@{}c|cccll|llll@{}}
    \toprule
    \multicolumn{1}{l|}{} & Size & \#Tasks & Lang & \multicolumn{1}{c}{Ref Source} & \multicolumn{1}{c|}{Domain}                  & \multicolumn{1}{c}{Open} & \multicolumn{1}{c}{IF} & \multicolumn{1}{c}{Dims} & \multicolumn{1}{c}{Metric} \\ \midrule
    ~\citet{writingprompts}       &  10k+   & 1              & EN  & Reddit writing prompts               & 100-200 Story                                &   Yes             &   No          & No                             & BLEU, ROUGE                \\
    ~\citet{kim2024biggenbenchprincipledbenchmark}           & 770     &  9      & EN  &  LLM generated mainly                 & IF \& Reasoning \& Safety & No                             & Yes                                       & Rubric                         & LLM metric                 \\
    ~\citet{guan-etal-2022-lot}                   & 729       & 4              & CN  & Online stories                       & Story                                        & Yes                            & No                                        & No                             & BLEU, DIST       \\
    ~\citet{mtbench}              & 10        & 1              & EN  & Self Constructed                     & Functional Writing                           & No                             & Yes                                       & General                        & LLM-as-a-judge             \\
    ~\citet{alignbench}            & 75        & 4              & CN  & LLM generated                          & Text Generation                              & No                             & Yes                                       & Rubric                         & LLM-as-a-judge             \\
    ~\citet{2025wb} & 1239 & 6 & CN/EN  & LLM generated with human refine& IF-style writing & No & Yes & Auto-Plan & LLM-as-a-judge \\ \midrule
    \textbf{\bench{} (Ours)}           & 1302       & 3x12             & CN  & Professional                         & Creative \& Functional Writing               & Yes                            & Yes                                       & Text Features                     & \method{}             \\ \bottomrule
    \end{tabular}%
    }
    \caption{Differences between our work and previous advances in natural language generation and instruction following fields. \textbf{Lang} stands for language. \textbf{Ref} stands for reference. \textbf{EN} stands for English and \textbf{CN} stands for Chinese. \textbf{IF} stands for instruction following.}
    \label{tab:relatedwork}
    \vspace{-10pt}
\end{table*}

\subsection{Benchmarking LLM Writing}

Early research on evaluating LLM-generated writing focused heavily on narrative quality within constrained genres like prompt-to-stories \citep{rocstories, guan-etal-2021-openmeva}. While recent benchmarks have shifted toward evaluating general text generation, emphasizing instruction adherence, coherence, and domain knowledge \citep{mtbench, alignbench, zhang-etal-2024-prolex, zhang-etal-2024-decor, liang2023helm, 2025wb}, they struggle with the open-ended nature of diverse writing tasks. Also, reference-free methods are often biased towards generations that are similar to the judge's \citep{deutsch-etal-2022-on-the-limitations-of-reference-free-evaluation}. \bench{} advances this research by (1) expanding to $12$ distinct genres across three task-forms, and (2) evaluating format, content, and subjective impressions independently with high quality human references.

\subsection{LLM-based Evaluation}

Recent advances in LLM-based evaluation utilize proprietary models for automated scoring through prompt engineering~\citep{mtbench, liu-etal-2023-geval} or tuning on human annotations~\citep{wang2024pandalm, ke-etal-2024-critiquellm}. These methods surpass traditional metrics like BLEU~\citep{papineni-etal-2002-bleu} and ROUGE~\citep{lin-2004-rouge} in efficiency and alignment with human correlation, particularly for constrained tasks such as summarization. However, their reliability weakens in the context of open-ended writing evaluation: verbosity bias~\citep{mtbench}, positional bias~\citep{wang-etal-2024-large-language-models-are-not-fair-evaluators}, and rubric dependency~\citep{ke-etal-2024-critiquellm, kim2024prometheus} hinder their generalizability across diverse genres. In contrast, attempts~\citep{2025wb} that involve LLMs autonomously generating evaluation criteria and rubrics emerged, but their robustness remains largely unexamined.
A comparison of our work to previous works is listed in Table~\ref{tab:relatedwork}.

\section{Evaluation Methodology}

\subsection{Tree-of-Writing Mechanism}

\begin{figure*}[t]
    \centering
    \includegraphics[width=0.95\linewidth]{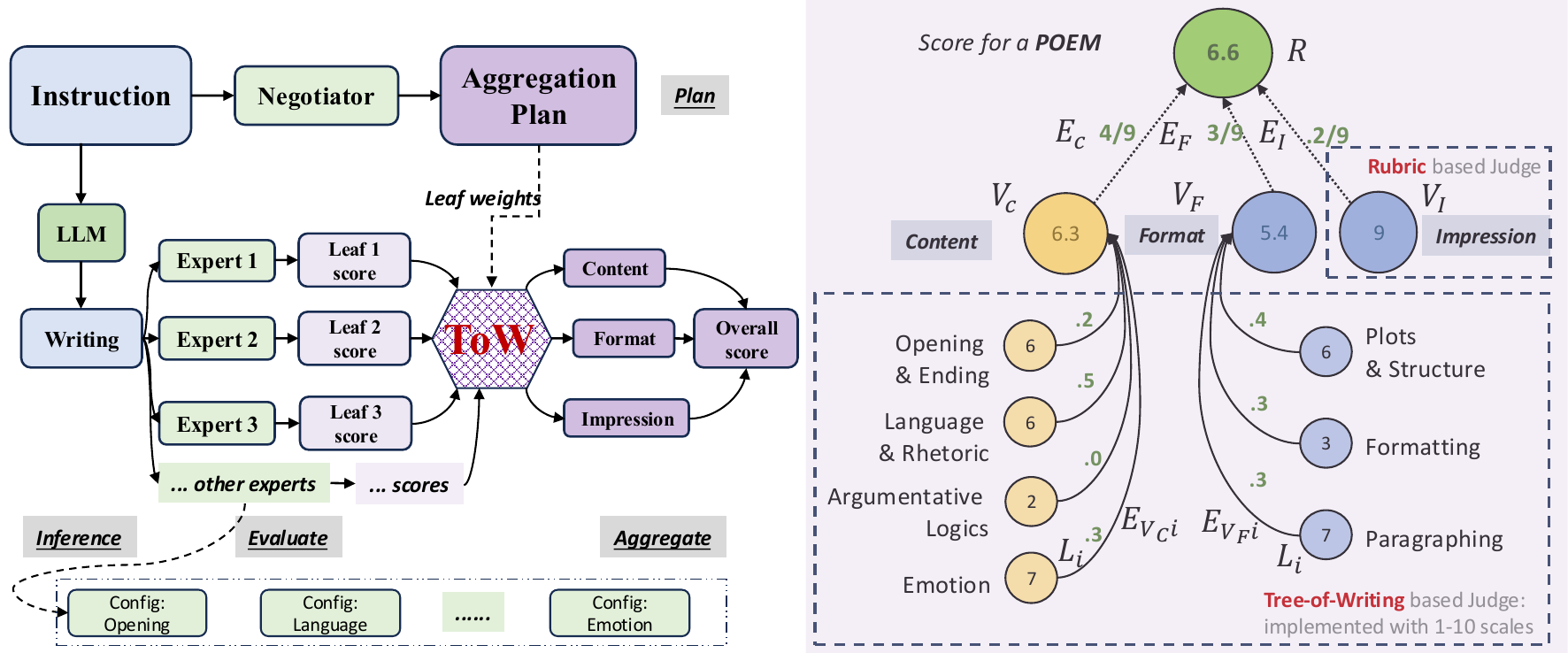}
    \caption{Overview of the evaluation framework incorporating the \method{}. The tree is rooted at the overall score $R$, which branches into three primary nodes: \textbf{Content} ($V_C$), evaluating semantic quality through \textit{traits} such as coherence, logics, richness, and opening-ending; \textbf{Format} ($V_F$), assessing structural adherence including plots, paragraphing, and formatting; and \textbf{Impression} ($V_I$), capturing the holistic subjective quality of the writing as a leaf node without further decomposition. Weighted edges ($w$) between nodes are explicitly determined by the negotiator $J_W$ based on each instruction.}
    \label{fig:method-overview}
    \vspace{-10pt}
\end{figure*}

We introduce Tree-of-Writing (\method{}), aiming to solve the hierarchical judgment nature of writing evaluation. Human evaluation of complex text typically decomposes general traits into specific sub-criteria~\citep{que2024hellobench, alignbench, wen2024complexbench}, a process that naturally aligns with a tree-based structure mirroring depth-first traversal. We therefore model evaluation as such a tree (Figure~\ref{fig:method-overview}). 
Three task-agnostic primary nodes, \textbf{content} ($V_C$), \textbf{format} ($V_F$), and \textbf{impression} ($V_I$), are selected based on dimensions that recur across text generation evaluation literature~\citep{guan-etal-2022-lot, alignbench, wang2025vff}, and empirically validated in Section~\ref{sec:ablation}. 
Let $R$ denote the root of the evaluation tree. $V_C$, $V_F$, and $V_I$ are connected to $R$ by weighted edges $E_C$, $E_F$, $E_I$. Both $V_C$ and $V_F$ may each have additional leaf nodes $L_i$ representing more granular assessment traits, each connected to its parent by a weighted edge $E_{V_{\mathrm{Parent}(L_i)}L_i}$. $V_I$ does not have children and is therefore a leaf node. Here $\mathrm{Parent}(\cdot)$ returns the parent node of the given variable. The scores are calculated with a DFS of the tree:
\begin{align*}
    \mathrm{Score}(V_C) &=& \sum_{L_i\in \mathrm{Child}(V_C)}w_{E_{V_CL_i}} \mathrm{Score}(L_i) \\
    \mathrm{Score}(V_F) &=& \sum_{L_i\in \mathrm{Child}(V_F)}w_{E_{V_FL_i}} \mathrm{Score}(L_i) \\
    \mathrm{Score}(R) &=& \sum_{j\in{\{C,F,I\}}} w_{E_j} \mathrm{Score}(V_j)
\end{align*}

$\mathrm{Child}(\cdot)$ refers to the children function which returns the children of the variable node.

\subsection{Scoring Function}

There is an important consideration in implementing the $\mathrm{Score}(\cdot)$ function: different scoring approaches are employed for different node types.

For the \textbf{Content} nodes $V_C$, each leaf node corresponds to a specific trait. We implemented them using a combination of a rubric with a reference approach. Formally speaking, several LLMs assign a score from 1 to 10 to each leaf node. The criteria and corresponding descriptions for these scores are provided in Table~\ref{tab:content illustration}.

For the \textbf{Format} nodes, we adopt a hybrid approach combining rule-based and LLM-based methods. The scoring function for these nodes operates as a step function, assigning scores of 0, 5, 10. For nodes such as \textit{Plots \& Structure} and \textit{Paragraphing}, an LLM-based judge evaluates whether the structure and level of detail in content are appropriate. For \textit{Formatting} leaf nodes, a regex-based approach is employed to detect whether the titles are appropriately formatted and respect the correct hierarchical structure. The detailed scoring criteria are outlined in Table~\ref{tab:format illustration} and the specific implementation of the regex based method is provided in Appendix~\ref{sec:implementation}.

\subsection{Edge Weighting}

For all leaf nodes, we employ an \textit{explicit} edge-weighting approach. Specifically, an LLM determines the edge weights for each instruction $\mathcal{I}$, ensuring all weights are between -1 and 1 and sum up to 1:
\begin{align*}
    & \small (w_{E_{V_{X L_1}}}, \cdots ,  w_{E_{V_{X L_n}}})^i = \mathrm{J}_W(\mathcal{I}^i), X \in \{C, F\} \\  & \text{s.t. }  \sum_{k=1}^n w_{E_{V_{X L_k}}} = 1, \quad w_{E_{V_{X L_k}}} \in (-1, 1)
\end{align*}
After the scores of the leaf nodes are determined, we aggregate them according to these weights. It avoids inconsistencies inherent in the \textit{implicit} aggregation strategy often employed by the LLM, such as arbitrarily switching between averaging or selectively emphasizing particular dimensions for the same instruction. Moreover, our method improves the interpretability of the evaluation results, thereby facilitating further analysis. The implementation details for this part are provided in Appendix~\ref{appdx:edge-weight}.

For the aggregation of $\mathrm{Score}(V_C)$, $\mathrm{Score}(V_F)$ and  $\mathrm{Score}(V_I)$, we use an averaging strategy based on the number of leaf nodes. This design allows tasks like completion, which may lack a format dimension, to be integrated within a unified evaluation framework. It also offers flexibility when extending task types.

\section{\bench{}}

To holistically evaluate the capabilities of LLMs in generating human-level writings, we developed ~\bench{}, which is designed to cover a diverse range of writing genres through $3$ distinct single-round writing mode: \textbf{Completion}, \textbf{Guide}, and \textbf{Open}. \bench{} is distinguished by its high-quality expert-written references that are free from AI-generated content. 

\subsection{Task Definition}

LLM-based writing tasks are formalized within an input-output framework.

\noindent\textbf{Writing instruction} $\mathcal{I}$: lists the requirements for the writing task. It also includes a one-sentence summary of desired output.

\noindent\textbf{Grounding information} $\mathcal{G}$: Provides supplementary details such as formatting requirements, narrative or plot constraints, stylistic directives, or, in \textbf{Open} task, is omitted entirely.

\noindent\textbf{Human reference} $\mathcal{R}$: A carefully curated, high-quality human-written reference to the task, which serves as a crucial standard for evaluation.

Given these inputs, the LLM generates an output writing as follows:
\begin{equation*}
    \mathcal{W} = \mathrm{LLM}(\mathcal{I}, \mathcal{G})
\end{equation*}
This output is expected to reconstruct human-level quality based on the objectives outlined in the instruction. The generated writing is then evaluated with respect to both its content and format.

\subsection{Data Source: Crawling} \label{sec:crawl}

We collected a large set of high-quality, publicly-licensed, human-written texts via web crawling from several specialized literary and writing guide websites: \textbf{CN Writer}, \textbf{PW4ES}, \textbf{SeptES}, \textbf{ZJPub}, \textbf{Officials}. Detailed descriptions of these sources are provided in \textbf{Appendix}~\ref{sec:appdx-data-source}. All texts were authored by human writers or domain experts. 

\subsection{Reference: Categorizing and Filtering}\label{sec:ref-category}

We employ a category classifier to assign each crawled text $T$ to a writing genre $c = \mathrm{Cls}(T)$. Specifically, we implement this using a prompted LLM approach, utilizing GPT-4o-1120 with the prompts detailed in \textbf{Appendix}~\ref{sec:cls-prompt}. To ensure labeling accuracy, three human experts\footnote{Master's degree in humanities, journalism, finance respectively with two years of working experience in LLM industry.} manually check the GPT-generated genre tags. For all $1302$ prompts, GPT-4o-1120 achieved an accuracy of $98.6\%$. Any misclassified instances were manually corrected by the human experts.
The writing genres are: fiction, poetry, prose, essay, argumentative essays, reports, summaries, letters, speeches, deliveries, plans, contracts, officials. Further details on each genre are provided in \textbf{Appendix}~\ref{sec:genres}.

\begin{figure}[t]
    \centering
    \includegraphics[width=1.0\linewidth]{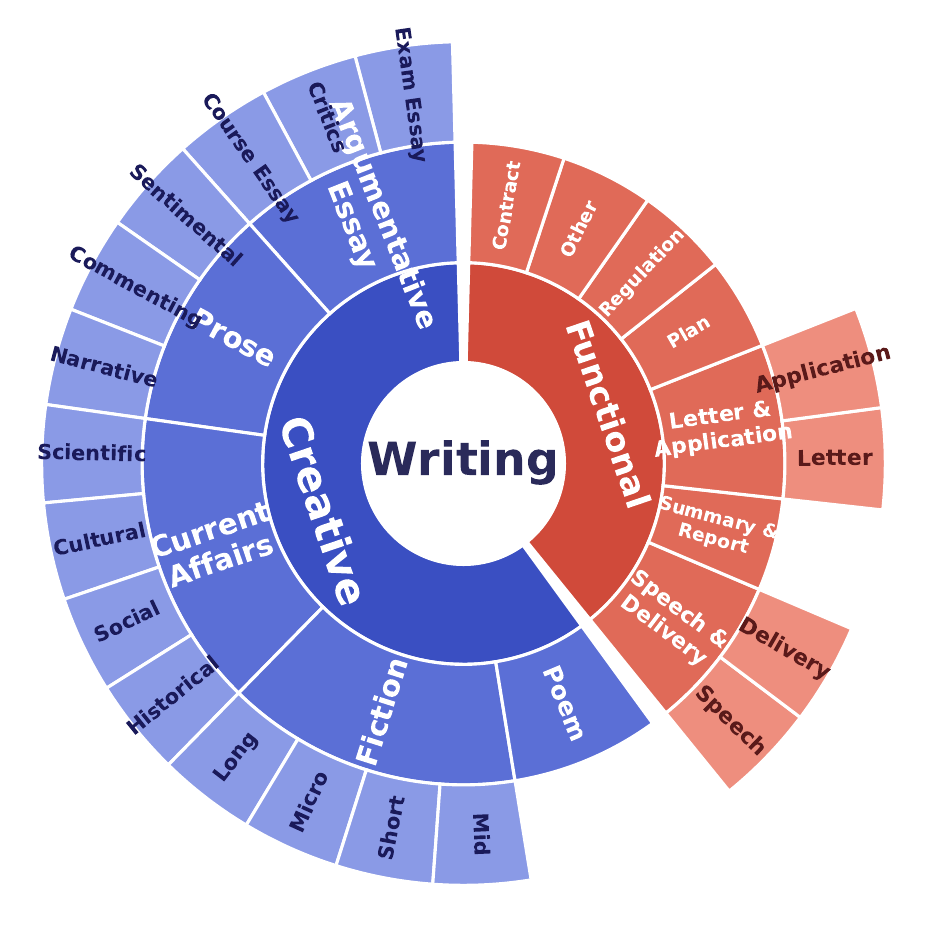}
    \caption{Hierarchal taxonomy of \bench{} showing the major categories.}
    \label{fig:taxonomy}
    \vspace{0pt}
\end{figure}

To further ensure bench data quality, we harness an additional LLM as a filter to remove the low-quality texts from the crawled data. Specifically, we obtain an overall quality score from 1 to 5 for each text, $s = \mathrm{Filter}(T)$, where a higher score indicates better quality. We use Claude-3-5-sonnet-20241022 for this task, prompting it with 12 genre-specific rubrics. The prompt for fiction is provided in \textbf{Appendix}~\ref{sec:filter-prompt} as an example. The score distributions for the aforementioned websites are shown in Table~\ref{tab:filter-score}. Most of the texts received scores between $3$ and $5$. We set a threshold score of $4$ and discard all examples that fall below this threshold.

\subsection{Task Design: Progressive Difficulty Levels}

To systematically evaluate the writing capabilities of LLMs, we design three tasks with increasing degrees of difficulty: completion, guided writing, and open writing. As the constraints and specificity of writing prompts decrease from Level I to Level III, the tasks require the model to generate increasingly creative and coherent text with less input information. Examples for each task are provided in Appendix~\ref{sec:data_examples}.

\textbf{Level I: Completion}: 
This task assesses the LLM's ability to complete a piece of unfinished text. Key portions of the original text are omitted, and the instruction $\mathcal{I}$ prompts the LLM to fill in the missing content based on the provided grounding information $\mathcal{G}$.

\textbf{Level II: Guided Writing}:
This task evaluates the LLM’s ability to generate and expand text following a given outline. Here, the instruction $\mathcal{I}$ directs the LLM to expand upon grounding information $\mathcal{G}$, which specifies more details for the desired output.

\textbf{Level III: Open Writing}: 
This task examines the LLM’s capacity to develop a coherent text on a given topic with minimal guidance. The instruction $\mathcal{I}$ specifies only the genre and presents the topic, plot, or argument in a single sentence. No grounding information is provided, i.e., $\mathcal{G} = \varnothing$.

%The writing aims to encompass a range of capabilities, including generating relevant titles, organizing the structure, designing coherent plots or arguments, and considering language style, tone, and audience. This task emphasizes a holistic evaluation of the LLM's comprehensive writing capabilities.

\subsection{Instruction: Reverse Construction}

We construct the instruction $\mathcal{I}$ and grounding information $\mathcal{G}$ based on high-quality reference texts. We refer to this process as back-construction due to its similarity to back-translation.

For \textbf{Completion}, we enlisted human annotators to manually remove sections from human-written content, using paragraphs as the minimal unit of removal. The number of omitted paragraphs is limited to 10. The resulting incomplete text is set as the grounding information $\mathcal{G}$, while the removed sections serve as the reference $\mathcal{R}$. The instruction $\mathcal{I}$ is composed using the template below.

\begin{tcolorbox}[title = {Instruction Template for completion}, colframe=blue!50!black, colback=blue!10, width=0.5\textwidth]
\vspace{-5pt}
\textbf{Input}: Genre \\
% \tcblower
Please fill in the blanks in the following \{genre\}, marked with [fill in the blank] signs. You should comprehensively consider the context and ensure the completion quality.
\vspace{-5pt}
\end{tcolorbox}

For \textbf{Guided Writing} and \textbf{Open Writing}, we utilize an LLM as the back-constructor. Formally, this procedure can be expressed as:
\begin{equation*}
    (S, T, \mathcal{G}) = \mathrm{BackConstruct}(\mathcal{R})
\end{equation*}
where $S$ denotes a summary of the original content and $T$ represents the theme, consisting of no more than five words. Both $S$ and $T$ are included in the instruction template to construct $\mathcal{I}$. 

\begin{tcolorbox}[title = {Instruction Template in guided/open writing}, colframe=blue!50!black, colback=blue!10,]
\textbf{Inputs}:Genre,Topic,Summary,Word counts\\
% \tcblower
Please write a \{genre\} about \{Topic\}. \{summary\}. You should write in approximately \{word counts\}.
\end{tcolorbox}

Besides, the back-constructor is assigned specific traits of the genre, and it needs to provide descriptions of writing requirements based on these traits, depending on $\mathcal{R}$. All the traits information is then composed in $\mathcal{G}$.
We implement the back-constructor with Gemini-2.0-Flash. We also prompt it with one-shot in-context example. The prompt for genre fiction is attached to \textbf{Appendix}~\ref{sec:backconstruction-prompt}.

\begin{table}[!t]
    \centering
    \resizebox{0.4\textwidth}{!}{%
    \begin{tabular}{@{}l|llll@{}}
    \toprule
               & \textbf{Comp}    & \textbf{Guide}   & \textbf{Open}    & \textbf{Total}   \\ \midrule
    \textbf{\#Creative}   & 379     & 277     & 282     & 938     \\
    Instr Len  & 44.02   & 88.82   & 89.29   & 70.86   \\
    Info Len   & 2016.02 & 318.48  & -       & 1299.22 \\
    Ref Len    & 431.37  & 1607.52 & 1726.05 & 1167.93 \\ \midrule
    \textbf{\#Functional} & -       & 179     & 185     & 364     \\
    Instr Len  & -       & 85.68   & 91.83   & 88.80   \\
    Info Len   & -       & 467.39  & -       & 467.39  \\
    Ref Len f  & -       & 1335.23 & 1373.91 & 1354.89 \\ \bottomrule
    \end{tabular}%
    }
    \caption{Statistics of \bench{}.}
    \label{tab:stat}
    \vspace{-10pt}
\end{table}

\subsection{Quality Assurance}

The initial curation for instruction and information is synthetic. We manually revise all \bench{} data to ensure quality. Specifically, we again involve the three experts described in Section~\ref{sec:ref-category} to assess the quality of the pairs. 

For each single pair, $\mathcal{I}$ and $\mathcal{G}$ are firstly evaluated for clarity, relevance to human-written reference, and naturalness of expression. The experts follow the guidelines outlined in Appendix~\ref{sec:picking_guide} to revise any cases that do not meet the required standards.

Further, we enhanced the quality for $R$ through a pairwise comparison process. For every instruction, we obtain LLM-generated responses from GPT-4o-1120, GLM-4-plus, Gemini-2.0-Flash. We then compile a tuple for each case: $(\mathcal{I}, (\mathcal{G}), \mathcal{R}, \mathcal{W}_{\mathrm{GPT}}, \mathcal{W}_{\mathrm{GLM}}, \mathcal{W}_{\mathrm{Gemini}})$. Human experts select the best-written responses among the four candidates according to the guideline in Appendix~\ref{sec:picking_guide}. Each case is independently handled by two randomly assigned experts, achieving $96.7\%$ agreement rate. Disagreements are resolved by a third expert, who makes the final decision based on prior assessments. Notably, $137(10.5\%)$ out of $1302$ original human writings were not chosen as the best among the four; we replaced these with the expert-selected candidates. 
Subsequently, for $\mathcal{I}, \mathcal{G}, \mathcal{R}$, personal information, unsafe content and undesired elements such as advertisements are either removed or rewritten in a de-identified form. During this process, annotators are assisted by an automated detector implemented using Deepseek-R1. The overall disqualification rate is $41/1302$.

Table~\ref{tab:stat} presents the statistics for \bench{}. Text lengths are measured in Chinese characters. Overall, the dataset instructions are clearly defined to facilitate evaluation, while the reference responses are of high quality and demonstrate excellence across a range of writing genres.

\section{Experiment}

\begin{table*}[t]
    \centering
    \resizebox{\textwidth}{!}{%
    \begin{tabular}{@{}lccccccccccccc@{}}
    \toprule
    \multirow{2}{*}{\textbf{Method}} & \multirow{2}{*}{\textbf{Cost (\$)}} & \multicolumn{3}{c}{\textbf{Comp}} & \multicolumn{3}{c}{\textbf{Guide}} & \multicolumn{3}{c}{\textbf{Open}} & \multicolumn{3}{c}{\textbf{ALL}} \\ \cmidrule(l){3-14} 
     &  & $\rho$ & $\tau$ & $\sigma$ & $\rho$ & $\tau$ & $\sigma$ & $\rho$ & $\tau$ & $\sigma$ & $\rho$ & $\tau$ & $\sigma$ \\ \midrule
    BLEU-1 & - & 0.85 & \textbf{0.67} & \textbf{0.80} & 0.65 & 0.54 & 0.69 & 0.70 & 0.50 & 0.62 & 0.75 & 0.56 & 0.72 \\
    BLEU-rt & - & 0.19 & 0.06 & 0.15 & -0.25 & -0.20 & -0.19 & -0.45 & -0.22 & -0.27 & -0.19 & -0.11 & -0.20 \\
    ROUGE-L & - & 0.87 & \textbf{0.67} & 0.75 & 0.06 & 0.14 & 0.20 & 0.46 & 0.22 & 0.32 & 0.46 & 0.06 & 0.17 \\ \midrule
    \method{} & 7.34 & \textbf{0.87} & \textbf{0.67} & 0.78 & \textbf{0.85} & \textbf{0.76} & \textbf{0.89} & 0.89 & 0.78 & 0.88 & \textbf{0.93} & \textbf{0.83} & \textbf{0.93} \\ 
    Average scoring (\method{} w/o plan) & 7.02 & 0.85 & 0.56 & 0.67 & 0.78 & 0.65 & 0.78 & 0.90 & 0.61 & 0.82 & 0.89 & 0.61 & 0.82 \\  \midrule
    Elaborated Rubric - worst & 1.31 & 0.86 & 0.56 & 0.70 & 0.78 & 0.65 & 0.78 & 0.89 & \textbf{0.83} & \textbf{0.90} & 0.89 & 0.61 & 0.82 \\
    Elaborated Rubric - best & 1.31 & 0.84 & 0.56 & 0.72 & 0.82 & 0.70 & 0.82 & \textbf{0.91} & \textbf{0.83} & \textbf{0.92} & 0.89 & 0.67 & 0.87 \\
    Elaborated Rubric + SC (n=5) & 6.53 & 0.85 & 0.56 & 0.70 & 0.80 & 0.65 & 0.78 & 0.90 & \textbf{0.83} & \textbf{0.90} & 0.89 & 0.61 & 0.82 \\
    Elaborated Rubric + SC (n=10) & 13.17 & 0.85 & 0.56 & 0.70 & 0.81 & 0.70 & 0.80 & 0.90 & \textbf{0.83} & \textbf{0.90} & 0.89 & 0.61 & 0.82 \\ \midrule
    Auto-Plan - worst & 0.89 & 0.63 & 0.40 & 0.49 & 0.78 & 0.63 & 0.74 & 0.83 & 0.61 & 0.73 & 0.87 & 0.50 & 0.62 \\
    Auto-Plan - best & 0.89 & 0.73 & 0.56 & 0.72 & 0.81 & 0.63 & 0.77 & 0.85 & 0.72 & 0.82 & 0.88 & 0.67 & 0.83 \\
    Auto-Plan + SC (n=5) & 4.45 & 0.73 & 0.42 & 0.57 & 0.79 & 0.54 & 0.67 & 0.84 & 0.67 & 0.85 & 0.88 & 0.67 & 0.83 \\
    Auto-Plan + SC (n=10) & 8.93 & 0.73 & 0.39 & 0.47 & 0.79 & 0.59 & 0.70 & 0.84 & 0.67 & 0.77 & 0.88 & 0.61 & 0.82 \\ \bottomrule
    \end{tabular}%
    }
    \caption{Assessment for evaluation methods and frameworks. System level Pearson correlation ($\rho$), Kendall rank correlation $\tau$ and Spearman rank correlation $\sigma$ are calculated. Values in bold indicate the best performance. "SC" stands for self-consistency configuration, and "worst"/"best" stand for the worst and best performance in the self-consistency results batch.}
    \label{tab:meta-eval}
\end{table*}

\begin{table*}[t]
    \centering
    \resizebox{0.8\textwidth}{!}{%
    \begin{tabular}{@{}c|c|ccc|ccccc|cc@{}}
    \toprule
               & \textbf{AVG}                 & \textbf{DS-R1}               & \textbf{o3-mini}             & \textbf{4o}                  & \textbf{CL-35-S}             & \textbf{Gemini}              & \textbf{DS-V3}               & \textbf{DB}                  & \textbf{GLM}                 & \textbf{CL-3-H}              & \textbf{LM}                  \\ \midrule
\textbf{Completion} & & 6.10 & 6.16 & 6.60 & 5.55 & 5.43 & 5.44 & 5.58 & 5.19 & 5.12 & 4.36 \\ 
\textbf{Guide} & & 6.15 & 5.80 & 5.61 & 5.76 & 5.53 & 5.52 & 5.24 & 5.51 & 5.08 & 4.89 \\ 
\textbf{Open} & & 6.06 & 5.69 & 5.36 & 5.43 & 5.33 & 5.31 & 5.14 & 5.28 & 4.85 & 4.47 \\ \midrule
Argumentative & \cellcolor[HTML]{ECF4FF}5.68 & \cellcolor[HTML]{9AFF99}6.24 & \cellcolor[HTML]{9AFF99}6.08 & \cellcolor[HTML]{9AFF99}6.23 & \cellcolor[HTML]{ECF4FF}5.73 & \cellcolor[HTML]{ECF4FF}5.54 & \cellcolor[HTML]{ECF4FF}5.61 & \cellcolor[HTML]{ECF4FF}5.74 & \cellcolor[HTML]{ECF4FF}5.69 & \cellcolor[HTML]{ECF4FF}5.16 & \cellcolor[HTML]{FFFFC7}4.77 \\
Comment        & \cellcolor[HTML]{ECF4FF}5.48 & \cellcolor[HTML]{ECF4FF}5.95 & \cellcolor[HTML]{9AFF99}6.02 & \cellcolor[HTML]{ECF4FF}5.98 & \cellcolor[HTML]{ECF4FF}5.54 & \cellcolor[HTML]{ECF4FF}5.36 & \cellcolor[HTML]{ECF4FF}5.53 & \cellcolor[HTML]{ECF4FF}5.30 & \cellcolor[HTML]{ECF4FF}5.36 & \cellcolor[HTML]{ECF4FF}5.10 & \cellcolor[HTML]{FFFFC7}4.65 \\
Poem           & \cellcolor[HTML]{ECF4FF}5.40 & \cellcolor[HTML]{9AFF99}6.00 & \cellcolor[HTML]{ECF4FF}5.81 & \cellcolor[HTML]{9AFF99}6.34 & \cellcolor[HTML]{ECF4FF}5.41 & \cellcolor[HTML]{ECF4FF}5.42 & \cellcolor[HTML]{ECF4FF}5.60 & \cellcolor[HTML]{ECF4FF}5.15 & \cellcolor[HTML]{ECF4FF}5.47 & \cellcolor[HTML]{FFFFC7}4.58 & \cellcolor[HTML]{FFFFC7}4.20 \\
Prose          & \cellcolor[HTML]{ECF4FF}5.32 & \cellcolor[HTML]{9AFF99}6.25 & \cellcolor[HTML]{ECF4FF}5.76 & \cellcolor[HTML]{ECF4FF}5.75 & \cellcolor[HTML]{ECF4FF}5.49 & \cellcolor[HTML]{ECF4FF}5.35 & \cellcolor[HTML]{ECF4FF}5.16 & \cellcolor[HTML]{ECF4FF}5.06 & \cellcolor[HTML]{ECF4FF}5.13 & \cellcolor[HTML]{FFFFC7}4.89 & \cellcolor[HTML]{FFFFC7}4.33 \\
Fiction        & \cellcolor[HTML]{ECF4FF}5.07 & \cellcolor[HTML]{9AFF99}6.08 & \cellcolor[HTML]{ECF4FF}5.36 & \cellcolor[HTML]{ECF4FF}5.37 & \cellcolor[HTML]{ECF4FF}5.32 & \cellcolor[HTML]{ECF4FF}5.23 & \cellcolor[HTML]{FFFFC7}4.82 & \cellcolor[HTML]{FFFFC7}4.84 & \cellcolor[HTML]{FFFFC7}4.89 & \cellcolor[HTML]{FFFFC7}4.53 & \cellcolor[HTML]{FFFFC7}4.25 \\ \midrule
Letters        & \cellcolor[HTML]{9AFF99}6.02 & \cellcolor[HTML]{9AFF99}6.38 & \cellcolor[HTML]{9AFF99}6.11 & \cellcolor[HTML]{9AFF99}6.13 & \cellcolor[HTML]{9AFF99}6.08 & \cellcolor[HTML]{9AFF99}6.12 & \cellcolor[HTML]{9AFF99}6.18 & \cellcolor[HTML]{9AFF99}6.07 & \cellcolor[HTML]{9AFF99}6.05 & \cellcolor[HTML]{ECF4FF}5.47 & \cellcolor[HTML]{ECF4FF}5.64 \\
Others         & \cellcolor[HTML]{ECF4FF}5.97 & \cellcolor[HTML]{9AFF99}6.33 & \cellcolor[HTML]{9AFF99}6.05 & \cellcolor[HTML]{9AFF99}6.30 & \cellcolor[HTML]{ECF4FF}5.91 & \cellcolor[HTML]{9AFF99}6.00 & \cellcolor[HTML]{9AFF99}6.02 & \cellcolor[HTML]{9AFF99}6.42 & \cellcolor[HTML]{ECF4FF}5.94 & \cellcolor[HTML]{ECF4FF}5.63 & \cellcolor[HTML]{ECF4FF}5.12 \\
Speech         & \cellcolor[HTML]{ECF4FF}5.60 & \cellcolor[HTML]{9AFF99}6.01 & \cellcolor[HTML]{ECF4FF}5.94 & \cellcolor[HTML]{ECF4FF}5.64 & \cellcolor[HTML]{ECF4FF}5.80 & \cellcolor[HTML]{ECF4FF}5.61 & \cellcolor[HTML]{ECF4FF}5.74 & \cellcolor[HTML]{ECF4FF}5.66 & \cellcolor[HTML]{ECF4FF}5.54 & \cellcolor[HTML]{ECF4FF}5.28 & \cellcolor[HTML]{FFFFC7}4.83 \\
Report         & \cellcolor[HTML]{ECF4FF}5.42 & \cellcolor[HTML]{ECF4FF}5.90 & \cellcolor[HTML]{9AFF99}6.00 & \cellcolor[HTML]{ECF4FF}5.29 & \cellcolor[HTML]{ECF4FF}5.82 & \cellcolor[HTML]{ECF4FF}5.26 & \cellcolor[HTML]{ECF4FF}5.55 & \cellcolor[HTML]{ECF4FF}5.18 & \cellcolor[HTML]{ECF4FF}5.11 & \cellcolor[HTML]{ECF4FF}5.30 & \cellcolor[HTML]{FFFFC7}4.81 \\
Contract       & \cellcolor[HTML]{ECF4FF}5.17 & \cellcolor[HTML]{ECF4FF}5.52 & \cellcolor[HTML]{ECF4FF}5.80 & \cellcolor[HTML]{FFFFC7}4.97 & \cellcolor[HTML]{ECF4FF}5.11 & \cellcolor[HTML]{ECF4FF}5.08 & \cellcolor[HTML]{ECF4FF}5.33 & \cellcolor[HTML]{ECF4FF}5.24 & \cellcolor[HTML]{ECF4FF}5.18 & \cellcolor[HTML]{ECF4FF}5.06 & \cellcolor[HTML]{FFFFC7}4.37 \\
Plan           & \cellcolor[HTML]{ECF4FF}5.03 & \cellcolor[HTML]{ECF4FF}5.44 & \cellcolor[HTML]{ECF4FF}5.75 & \cellcolor[HTML]{ECF4FF}5.02 & \cellcolor[HTML]{FFFFC7}4.97 & \cellcolor[HTML]{FFFFC7}4.94 & \cellcolor[HTML]{ECF4FF}5.11 & \cellcolor[HTML]{ECF4FF}5.23 & \cellcolor[HTML]{FFFFC7}4.83 & \cellcolor[HTML]{FFFFC7}4.78 & \cellcolor[HTML]{FFFFC7}4.26 \\
Regulation     & \cellcolor[HTML]{FFFFC7}4.90 & \cellcolor[HTML]{ECF4FF}5.31 & \cellcolor[HTML]{ECF4FF}5.13 & \cellcolor[HTML]{FFFFC7}4.66 & \cellcolor[HTML]{FFFFC7}4.91 & \cellcolor[HTML]{ECF4FF}5.07 & \cellcolor[HTML]{FFFFC7}4.87 & \cellcolor[HTML]{ECF4FF}5.07 & \cellcolor[HTML]{FFFFC7}4.69 & \cellcolor[HTML]{FFFFC7}4.59 & \cellcolor[HTML]{FFFFC7}4.72 \\ \midrule
All            &                              & 6.10                         & 5.86                         & 5.81                         & 5.58                         & 5.43                         & 5.42                         & 5.34                         & 5.34                         & 5.01                         & 4.59                         \\ \bottomrule

    \end{tabular}%
    }
    \caption{\textbf{Bench scores} genre-wisely. For model abbreviations, \textbf{DS-R1} refers to Deepseek-R1, \textbf{o3-mini} refers to GPT-4-o3-mini-2025-01-31, \textbf{4o} refers to GPT-4o-1120, \textbf{CL-3.5-S} refers to Claude-3-5-sonnet-20241022, \textbf{Gemini} refers to Gemini-2.0-flash, \textbf{DS-V3} refers to DeepSeek-V3, \textbf{GLM} refers to GLM-4-Plus-250111, \textbf{DB} refers to Doubao-pro-241225, \textbf{CL-3-H} refers to Claude-3-haiku-20240307, \textbf{LM-3.3} refers to Llama-3.3-70B-Instruct.}
    \label{tab:bench_score}
    \vspace{0pt}
\end{table*}

\subsection{Settings}

\paragraph{Baselines}
We compared two groups of evaluation methods: \textbf{automatic metrics} and \textbf{LLM-as-a-judge} approaches. Automatic metrics include BLEU~\citep{papineni-etal-2002-bleu}, ROUGE~\citep{lin-2004-rouge}, and BLEURT~\citep{sellam-etal-2020-bleurt}. For LLM-based evaluation, we consider the original LLM-as-a-judge framework~\citep{mtbench} as well as its recent variants. Among these, we focus on two representative approaches: \textbf{Auto-Planning} and \textbf{Elaborated Rubrics}. In \textbf{Auto-Planning}, the LLM evaluator synchronously determines the relevant subdomains and aggregation strategies, and then produces scores in one single query-response. In contrast, \textbf{Elaborated Rubrics} utilize carefully designed evaluation prompts that are directly inspired by human annotation guidelines. The evaluation prompts used for all task genres are provided in Appendix~\ref{sec:rubric}.

\paragraph{Evaluated LLMs}
We evaluated our methods using several state-of-the-art LLMs, including GPT-4o-2024-11-20~\citep{gpt-4o}, Gemini-2.0-flash, Deepseek-R1~\citep{deepseek-R1}, Deepseek-V3~\citep{deepseek-v3}, Doubao-pro-32k~\citep{doubao-pro}, GLM-4-plus-250111~\citep{glm-4-plus}, Claude-3-5-sonnet-20241022~\citep{claude-3-5-sonnet}, Claude-3-haiku-20240307~\citep{claude-3-5-haiku}, and Qwen-plus~\citep{qwen-2.5}. For all LLM-as-a-judge based methods, GPT-4o was used as the evaluation model.

\subsection{Meta Evaluation}

We release \textbf{MetaEditor}, a meta-evaluation dataset designed for comprehensive assessment of writing task evaluation methods. \textbf{MetaEditor} comprises human ratings of LLM-generated writings in \bench{}. We selected 221 instructions (67 for Completion, 83 for Guide, and 71 for Open) from a total of 1,302 prompts, ensuring random and even coverage of all genres. For each instruction, nine LLM-generated writings are included, as sourced from Table~\ref{tab:bench_score}. We engaged 36 expert annotators with backgrounds in writing; further details about their expertise are provided in Appendix~\ref{sec:annotator}. All annotators underwent training based on the three annotation guidelines outlined in Appendix~\ref{tab:anno-doc}, and were instructed to rate the LLM outputs on a scale of 1 to 5. To ensure consistency, each set of nine writings corresponding to the same instruction was evaluated by a single annotator, and each individual LLM-generated writing was scored by two annotators for cross-validation. The overall inter-annotator agreement is $0.71$ using Cohen's Kappa and $0.87$ using Pearson correlation, indicating high consistency among human raters. We averaged the two scores to obtain a single final score for each writing, thereby preserving the diversity of human judgments.\footnote{We engaged the five experts who achieved the greatest agreement with other annotators throughout the process and asked them to re-check all annotations.}

\subsection{\method{} Result and Analysis}

As shown in Table~\ref{tab:meta-eval}, \method{} achieves Pearson and Spearman correlations of $0.93$ across all tasks in \bench{}. Notably, a comparison between BLEU and ROUGE-L suggests that the intended evaluation does not primarily depend on the recall with reference compared to precision. The BLEU-rt metric, which is model-based, yields random results, indicating that evaluation with a weak neural model is less reliable than those based on overlapping measures. Auto-planning also produced inferior outcomes on the \textbf{Completion} and \textbf{Guide} tasks, further highlighting its limitations when assessing tasks that involve substantial guidance.

\paragraph{Negotiation Bias Analysis}: 
\label{sec:negotiation-bias} 
To provide direct evidence for the negotiation bias discussed in Section~\ref{sec:intro}, we analyze the variability of the aggregation process under the Auto-Planning with Self-Consistency ($N\!=\!5$) setting. For each sample, we compute the ratio of each subscore $s_k$ to the total score $S$ as $l_k = s_k / S$, and measure its fluctuation across trials using two metrics:
\begin{align*}
    \delta &= \sum_{t=2}^{5} |l_k^{(t)} - l_k^{(1)}|, \quad
    \sigma = \sqrt{\sum_{t=1}^5 (l_k^{(t)} - \mu_{l_k})^2}
\end{align*}
where $\mu_{l_k}$ is the mean of $l_k$ over the $N$ trials. For Auto-Planning, we obtain $\overline{\delta}\!=\!0.273$ and $\overline{\sigma}\!=\!0.059$; for \method{}, where $l_k$ represents the weights planned by $J_W$, the same metrics yield $\overline{\delta}\!=\!0.080$ and $\overline{\sigma}\!=\!0.017$. The substantially lower fluctuation under \method{} confirms that explicit weight assignment effectively reduces the aggregation instability present in implicit planning approaches.

\paragraph{Self-Consistency}: 
As shown in Table~\ref{tab:meta-eval}, under comparable cost, \method{} achieves stronger overall alignment with human judgments compared to baselines harnessed with self-consistency. Self-consistency reduces random variation but does not yield significant improvement with increasing trials. With more trials, Auto-Planning converges toward Average Scoring, indicating its instability due to the lack of explicit, determined weights.

\paragraph{Ablation on \method{} Nodes}: 
\label{sec:ablation}
To validate the necessity of all three primary nodes in \method{}, we conduct an ablation study where each node is removed in turn (\textit{w/o}) or retained as the sole evaluation dimension (\textit{w}). Results in Table~\ref{tab:ablation-node} show that removing any single node consistently degrades correlation with human judgments, confirming that content, format, and impression each contribute complementary evaluation signals. Notably, retaining format alone (\textit{w Format}) leads to the largest performance drop, indicating that format information is insufficient for holistic writing evaluation but necessary for content-based assessment.

\begin{table}[t]
    \centering
    \resizebox{\columnwidth}{!}{%
    \begin{tabular}{@{}lcccccc@{}}
    \toprule
     & \multicolumn{3}{c}{\textbf{Guide}} & \multicolumn{3}{c}{\textbf{Open}} \\
    \cmidrule(lr){2-4} \cmidrule(l){5-7}
     & $\rho$ & $\tau$ & $\sigma$ & $\rho$ & $\tau$ & $\sigma$ \\
    \midrule
    \method{}      & \textbf{0.85} & \textbf{0.76} & \textbf{0.89} & \textbf{0.89} & \textbf{0.78} & \textbf{0.88} \\
    \textit{w/o Content}    & 0.81 & 0.76 & 0.88 & 0.84 & 0.78 & 0.88 \\
    \textit{w/o Format}     & 0.80 & 0.65 & 0.78 & 0.89 & 0.72 & 0.82 \\
    \textit{w Content}      & 0.79 & 0.65 & 0.78 & 0.89 & 0.78 & 0.87 \\
    \textit{w Format}       & 0.71 & 0.71 & 0.65 & 0.67 & 0.50 & 0.68 \\
    \textit{w Impression}   & 0.81 & 0.70 & 0.79 & 0.90 & 0.72 & 0.82 \\
    \bottomrule
    \end{tabular}%
    }
    \caption{Ablation study on tree nodes. \textit{w/o} denotes removal of a node; \textit{w} denotes retaining only that node.}
    \label{tab:ablation-node}
\end{table}

\paragraph{Edge Weight Distribution for Content}

\begin{figure}[t]
    \centering
    
    % Row 1
    \begin{subfigure}[b]{0.4\textwidth}
        \centering
        \includegraphics[width=\textwidth]{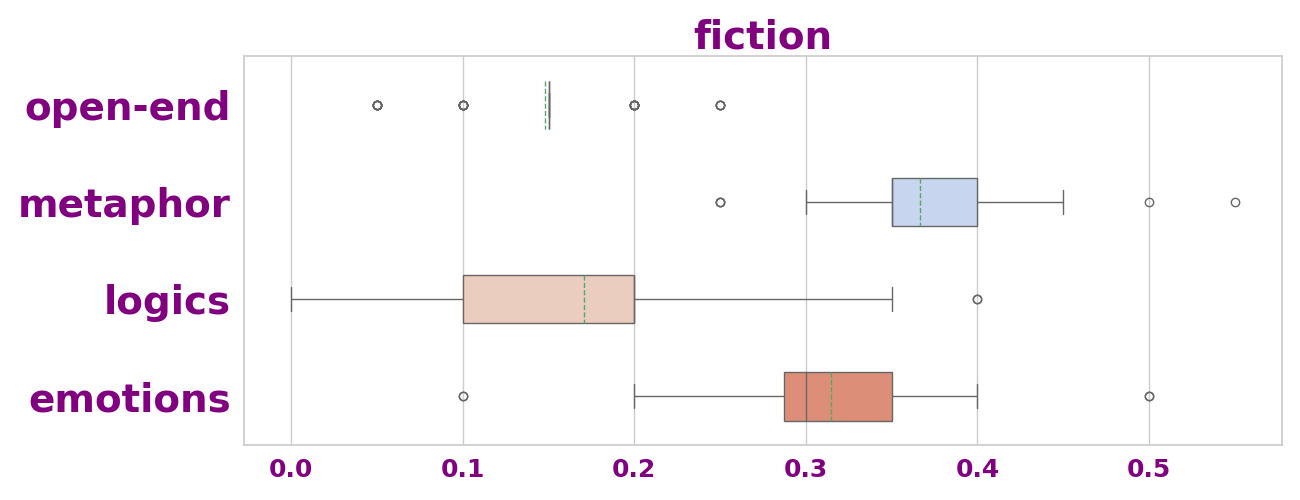} 
    \end{subfigure}
    
    \begin{subfigure}[b]{0.4\textwidth}
        \centering
        \includegraphics[width=\textwidth]{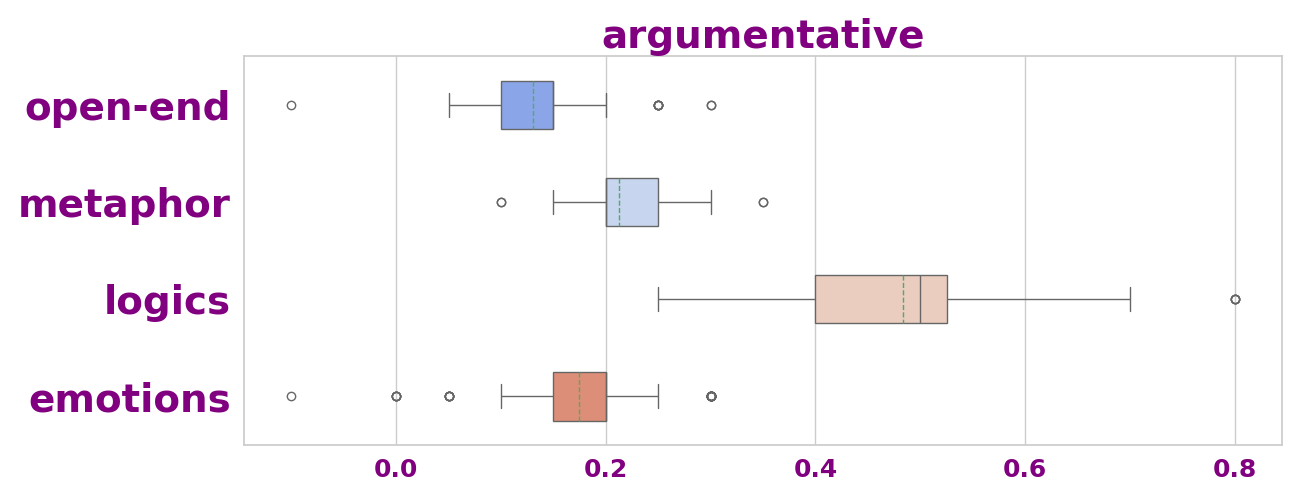} % Replace with your image path/file name
    \end{subfigure}

    % Caption for the entire figure
    \caption{Edge weight distribution on fiction, argumentative. The wider the box horizontally, the more varied the corresponding weight within the genre.}
    \label{fig:edge-weight-selected}
    \vspace{0pt}
\end{figure}

We analyzed the edge weights assigned by the negotiator to the four leaf nodes under the content node $V_C$. As illustrated in Figure~\ref{fig:edge-weight-selected}, these weights differ significantly across genres. Interestingly, we observed that the weights for '\textit{logics}' exhibited notable variation within most genres. Additionally, a consistent pattern emerged: the weights for opening-ending remained stable at approximately $10\%$ across all genres. However, across all genres, the edge weights are not evenly distributed among the four leaf nodes. Full plots for all genres can be found in Figure~\ref{fig:edge-weight} in Appendix~\ref{apd:edge-weight}.

\subsection{Benchmarking Results }

Table~\ref{tab:bench_score} reports the overall performance of various LLMs evaluated under \method{}. While advanced reasoning and proprietary models (e.g., Deepseek-R1, o3-mini, GPT-4o) consistently lead, the \textbf{Completion}, \textbf{Guide}, and \textbf{Open} tasks expose a clear difficulty gradient. Most models excel at constrained instruction-following but struggle with open-ended writing, evidenced by GPT-4o's performance drop of $15\%$ and $18.8\%$ on the \textbf{Guide} and \textbf{Open} tasks relative to \textbf{Completion}. Furthermore, the results indicate high genre sensitivity, with models succeeding on structured formats like \textit{Letters} but faltering on demanding ones like \textit{Fiction}. Ultimately, these findings support our assertion that human-level writing competence encompasses much more than simple imitation.

\section{Discussion}

\subsection{\textit{Mimic Game}: Longer is NOT Better}

We evaluate the impact of input quantity to the LLMs on the writing performance of models. All the writing outputs generated by LLMs are categorized according to \textbf{Completion}, \textbf{Guide} and \textbf{Open}. We conduct correlation analysis and linear regression on the relationships among input length, output length, and final scores, arriving at the results shown in the Figure~\ref{fig:mimic-game} and Table~\ref{tab:mimic-game}.

\begin{table}[t]
    \centering
    \small
    \resizebox{0.3\textwidth}{!}{%
    \begin{tabular}{@{}c|ccc@{}}
    \toprule
    \textbf{}                  & \textbf{Comp} & \textbf{Guide} & \textbf{Open} \\ \midrule
    Input - Output   & 0.24**     & 0.32**       & 0.25**       \\ \midrule
    Input - Overall  & -0.01      & {\color[HTML]{FE0000}-0.44}**      & -0.15**     \\
    Input - Content  & -0.01      & {\color[HTML]{FE0000}-0.44}**      & -0.11**      \\
    Input - Format   & N/A        & -0.16**      & 0.00*      \\ \midrule
    Output - Overall & 0.38**     & -0.18**      & -0.12**        \\
    Output - Content & 0.38**     & -0.09**      & -0.08**    \\
    Output - Format  & N/A        & -0.16**      & -0.09**   \\ \bottomrule
    \end{tabular}%
    }
    \caption{Pearson Correlation between input length, output length and final scores. ** marks the p < 0.05 significance and * marks the p < 0.1 significance. }
    \label{tab:mimic-game}
    \vspace{0pt}
\end{table}

\begin{table}[t]
    \resizebox{0.48\textwidth}{!}{%
    \begin{tabular}{@{}c|c|cc|cccc@{}}
    \toprule
    \textbf{}           & \textbf{Init.} & \textbf{Drop} & \textbf{Rep}                         & \textbf{To C}                        & \textbf{To L}                        & \textbf{To O}                       & \textbf{To P}                        \\ \midrule
    \textbf{ToW}                 & 5.41           & -0.36         & -0.49                                & -0.30                                & -0.31                                & -0.97                               & -0.62                                \\
    Tow-Content         & 5.82           & -0.34         & -0.48                                & -0.17                                & -0.10                                & -1.12                               & -0.36                                \\
    Tow-Format          & 5.77           & -0.58         & -0.81                                & -0.69                                & -0.65                                & -0.74                               & -1.12                                \\
    Tow-Impression      & 6.76           & -0.24         & -0.30                                & -0.14                                & -0.36                                & -1.52                               & -0.70                                \\ \midrule
    Auto-planning & 6.82           & -0.06         & -0.30                                & {\color[HTML]{FE0000} \textbf{0.08}} & {\color[HTML]{FE0000} \textbf{0.04}} & {\color[HTML]{FE0000} \textbf{0.20}} & {\color[HTML]{FE0000} \textbf{0.82}} \\ 
    BLEU                & 24.66          & -7.27         & {\color[HTML]{FE0000} \textbf{4.23}} & {\color[HTML]{FE0000} \textbf{0.97}} & {\color[HTML]{FE0000} \textbf{1.21}} & -1.56                               & -8.50                                \\
    BLEU-rt             & 37.43          & -2.07         & -0.35                                & -2.37                                & {\color[HTML]{FE0000} \textbf{1.55}} & {\color[HTML]{FE0000} \textbf{3.20}} & {\color[HTML]{FE0000} \textbf{1.91}} \\ \bottomrule
    \end{tabular}%
    }
    \caption{Robustness test of frameworks and metrics on common disturbances. \textbf{Init.} shorts for initial writing, \textbf{Rep} for repetition, \textbf{To C/L/O/P} for converting to comment, letter, official, poem. All scores are the results of subtracting the initial score on the left, with a negative sign indicating values lower than the initial score. The {\color[HTML]{FE0000}\textbf{bold red}} indicates the undesired changes.}
    \label{tab:robust}
    \vspace{0pt}
\end{table}

There is a significant positive correlation between output length and input length, which is consistent with previous research findings. For the \textbf{Guide} and \textbf{Open}, we perform linear fitting on the generation results of all models. The slopes are $1.4$ and $6.1$, respectively, indicating the input tokens conversion ratio to the output.

However, we find that on both \textbf{Guide} and \textbf{Open} tasks, regardless of \textbf{input} or \textbf{output}, the final scores exhibit a significant negative correlation with length. This differs from previous understandings where LLM evaluators were thought to favor verbosity. Additionally, we explain that providing more input does not necessarily induce better performance. LLMs are unable to rely on piling up input information to produce high-quality, nuanced writings. This is particularly evident in the \underline{Content} and \underline{Overall} scores for Guide tasks, where a correlation of -0.44 was observed.

We leave further discussions to the Appendix, such as different base-LLM evaluators (Appendix~\ref{sec:evaluator}), the comparison between reference-based and reference-free LLM judgment (Appendix~\ref{sec:ref-based-free}), between human-originated reference and LLM-originated reference (Appendix~\ref{sec:ref-source}).

\subsection{\textit{Negotiation Inconsistency} Pro: Robustness}

Currently, metric robustness has aroused community concerns, since \textit{reward hacking}~\citep{rewardhacking} are often encountered in practice. We conducted another experiment to validate the~\method{}'s robustness against common text disturbances.

We randomly pick 50 generation samples from LLMs presented in Table~\ref{tab:bench_score} (5 for each). We apply the following 3 disturbances to the generated writings following~\citep{guan-etal-2021-openmeva}: (1) \textbf{Drop}: randomly drop at most 3 paragraphs or sentences. (2) \textbf{Repeat}: repeat at most 3 paragraphs in the original writing at different positions. (3) \textbf{Transfer}: convert the writing genre to another genre. In practice, we pick comment, letter, official, poem as the target genres. We examine~\method{}, auto-planning LLM-evaluator, BLEU, BLEU-rt metrics and show in Table~\ref{tab:robust}.

Through the results, we can find that~\method{} responds to all the disturbances with score decrement. However, auto-planning, BLEU, BLEU-rt metrics are vulnerable to these interferences, indicating their limitations, which might introduce structures for bypassing designed assessment.

\section{Conclusion}

This work addresses the challenge of evaluating LLMs in open-ended writing by introducing the ~\bench{} benchmark and the \method{} evaluation framework. \method{} replaces traditional fixed-schema with tree-structured reasoning flow, reducing negotiation bias and demonstrating strong alignment with human judgment. Ultimately, our analysis reveals significant disparities among leading models in balancing format adherence, content quality, and creativity, highlighting the critical need for genre-specific evaluation standards.

\newpage

\section*{Limitations}

First, although ~\bench{} spans $12$ genres, its evaluation of writing ability operates at a genre-category level rather than addressing granular subgenres or specialized stylistic variations within each genre. This leaves fine-grained distinctions in domain-specific writing proficiency unexplored. 
Furthermore, our benchmark relies primarily on Chinese data. Extending the evaluation to multilingual contexts remains an open challenge, which we leave for future work.

Second, the evaluation focuses on single-round generation and excludes iterative refinement processes. Methodologies involving self-critique, multi-round human-AI collaboration, or dynamic feedback integration remain unexplored, which is critical for real-world writing workflows. This restricts insights into how LLMs adapt to evolving user requirements or contextual adjustments. We leave this scope for future explorations.

Finally, we did not test the scalability of the \method{} approach, particularly with respect to the correlation between selected dimensions and the feasibility of adding new leaf nodes. Due to the current lack of a comprehensive task framework in the domain of complex text, we adopted a relatively conservative Writing Tree modeling approach.

\section*{Acknowledgment}

This work was supported by the National Science Foundation for Distinguished Young Scholars (with No. 62125604), the Natural Science Foundation of China (No. 62536008), and the National Natural Science Foundation of China Major Program under Grant 92570203.

\section*{Ethical Statement}

The data collection protocol, human annotation guideline for this study were reviewed by the Institutional Review Board at the Department of Computer Science and Technology, Tsinghua University and Z.ai and were determined to be exempt from full board review, because the study involves minimal risk to the participants and no PII was collected.
The citations for Claude, Gemini, Doubao~\cite{claude-3-5-sonnet, claude-3-5-haiku, gemini-2.0, doubao-pro} lack official preprint technical reports, and the URL availability may evolve alongside corporate restructuring or updates.

% Bibliography entries for the entire Anthology, followed by custom entries
%\bibliography{anthology,custom}
% Custom bibliography entries only
\bibliography{custom}

@inproceedings{wang2025vff,
    title = "Verifiable Format Control for Large Language Model Generations",
    author = "Wang, Zhaoyang  and
      Jiang, Jinqi  and
      Zhou, Huichi  and
      Zheng, Wenhao  and
      Zhang, Xuchao  and
      Bansal, Chetan  and
      Yao, Huaxiu",
    editor = "Chiruzzo, Luis  and
      Ritter, Alan  and
      Wang, Lu",
    booktitle = "Findings of the Association for Computational Linguistics: NAACL 2025",
    month = apr,
    year = "2025",
    address = "Albuquerque, New Mexico",
    publisher = "Association for Computational Linguistics",
    url = "https://aclanthology.org/2025.findings-naacl.194/",
    doi = "10.18653/v1/2025.findings-naacl.194",
    pages = "3499--3513",
    ISBN = "979-8-89176-195-7",
}

@misc{wen2024complexbench,
      title={Benchmarking Complex Instruction-Following with Multiple Constraints Composition}, 
      author={Bosi Wen and Pei Ke and Xiaotao Gu and Lindong Wu and Hao Huang and Jinfeng Zhou and Wenchuang Li and Binxin Hu and Wendy Gao and Jiaxin Xu and Yiming Liu and Jie Tang and Hongning Wang and Minlie Huang},
      year={2024},
      eprint={2407.03978},
      archivePrefix={arXiv},
      primaryClass={cs.CL},
      url={https://arxiv.org/abs/2407.03978}, 
}

@misc{que2024hellobench,
      title={HelloBench: Evaluating Long Text Generation Capabilities of Large Language Models}, 
      author={Haoran Que and Feiyu Duan and Liqun He and Yutao Mou and Wangchunshu Zhou and Jiaheng Liu and Wenge Rong and Zekun Moore Wang and Jian Yang and Ge Zhang and Junran Peng and Zhaoxiang Zhang and Songyang Zhang and Kai Chen},
      year={2024},
      eprint={2409.16191},
      archivePrefix={arXiv},
      primaryClass={cs.CL},
      url={https://arxiv.org/abs/2409.16191}, 
}

@inproceedings{zhang-etal-2024-decor,
    title = "{DECOR}: Improving Coherence in {L}2 {E}nglish Writing with a Novel Benchmark for Incoherence Detection, Reasoning, and Rewriting",
    author = "Zhang, Xuanming  and
      Diaz, Anthony  and
      Chen, Zixun  and
      Wu, Qingyang  and
      Qian, Kun  and
      Voss, Erik  and
      Yu, Zhou",
    editor = "Al-Onaizan, Yaser  and
      Bansal, Mohit  and
      Chen, Yun-Nung",
    booktitle = "Proceedings of the 2024 Conference on Empirical Methods in Natural Language Processing",
    month = nov,
    year = "2024",
    address = "Miami, Florida, USA",
    publisher = "Association for Computational Linguistics",
    url = "https://aclanthology.org/2024.emnlp-main.639/",
    doi = "10.18653/v1/2024.emnlp-main.639",
    pages = "11436--11458",
}

@inproceedings{zhang-etal-2024-prolex,
    title = "{P}ro{L}ex: A Benchmark for Language Proficiency-oriented Lexical Substitution",
    author = "Zhang, Xuanming  and
      Chen, Zixun  and
      Yu, Zhou",
    editor = "Ku, Lun-Wei  and
      Martins, Andre  and
      Srikumar, Vivek",
    booktitle = "Findings of the Association for Computational Linguistics: ACL 2024",
    month = aug,
    year = "2024",
    address = "Bangkok, Thailand",
    publisher = "Association for Computational Linguistics",
    url = "https://aclanthology.org/2024.findings-acl.502/",
    doi = "10.18653/v1/2024.findings-acl.502",
    pages = "8475--8493",
}

@misc{2025wb,
      title={WritingBench: A Comprehensive Benchmark for Generative Writing}, 
      author={Yuning Wu and Jiahao Mei and Ming Yan and Chenliang Li and Shaopeng Lai and Yuran Ren and Zijia Wang and Ji Zhang and Mengyue Wu and Qin Jin and Fei Huang},
      year={2025},
      eprint={2503.05244},
      archivePrefix={arXiv},
      primaryClass={cs.AI},
      url={https://arxiv.org/abs/2503.05244}, 
}

@inproceedings{sellam-etal-2020-bleurt,
    title = "{BLEURT}: Learning Robust Metrics for Text Generation",
    author = "Sellam, Thibault  and
      Das, Dipanjan  and
      Parikh, Ankur",
    editor = "Jurafsky, Dan  and
      Chai, Joyce  and
      Schluter, Natalie  and
      Tetreault, Joel",
    booktitle = "Proceedings of the 58th Annual Meeting of the Association for Computational Linguistics",
    month = jul,
    year = "2020",
    address = "Online",
    publisher = "Association for Computational Linguistics",
    url = "https://aclanthology.org/2020.acl-main.704/",
    doi = "10.18653/v1/2020.acl-main.704",
    pages = "7881--7892",
}

@misc{rewardhacking,
      title={Defining and Characterizing Reward Hacking}, 
      author={Joar Skalse and Nikolaus H. R. Howe and Dmitrii Krasheninnikov and David Krueger},
      year={2025},
      eprint={2209.13085},
      archivePrefix={arXiv},
      primaryClass={cs.LG},
      url={https://arxiv.org/abs/2209.13085}, 
}

@misc{gemini-2.0,
  author = {{Gemini-Team}},
  title = {Introducing Gemini 2.0: our new AI model for the agentic era},
  year = 2024,
  url = {https://blog.google/innovation-and-ai/models-and-research/google-deepmind/google-gemini-ai-update-december-2024/},
  note = {Accessed: 2024-11-30}
}

@misc{doubao-pro,
  author = {{Bytedance-Team}},
  title = {Introduction to Doubao},
  year = 2024,
  url = {https://www.volcengine.com/product/doubao},
  note = {Accessed: 2024-11-30}
}

@misc{glm-4-plus,
      title={ChatGLM: A Family of Large Language Models from GLM-130B to GLM-4 All Tools}, 
      author={{Team GLM}},
      year={2024},
      eprint={2406.12793},
      archivePrefix={arXiv},
      primaryClass={cs.CL},
      url={https://arxiv.org/abs/2406.12793}, 
}

@misc{claude-3-5-sonnet,
  author = {{Anthropic}},
  title = {Claude 3.5 Sonnet},
  year = 2024,
  url = {https://www.anthropic.com/news/claude-3-5-sonnet},
  note = {Accessed: 2024-6-21}
}

@misc{claude-3-5-haiku,
  author = {{Anthropic}},
  title = {Introducing computer use, a new Claude 3.5 Sonnet, and Claude 3.5 Haiku},
  year = 2024,
  url = {https://www.anthropic.com/news/3-5-models-and-computer-use},
  note = {Accessed: 2024-10-22}
}

@misc{qwen-2.5,
      title={Qwen2.5 Technical Report}, 
      author={{Qwen-Team}},
      year={2025},
      eprint={2412.15115},
      archivePrefix={arXiv},
      primaryClass={cs.CL},
      url={https://arxiv.org/abs/2412.15115}, 
}

@misc{gpt-4o,
  author = {{OpenAI}},
  title = {GPT-4o Blog},
  year = 2024,
  url = {https://openai.com/index/hello-gpt-4o/},
  note = {Accessed: 2024-11-30}
}

@misc{deepseek-R1,
      title={DeepSeek-R1: Incentivizing Reasoning Capability in LLMs via Reinforcement Learning}, 
      author={DeepSeek-AI},
      year={2025},
      eprint={2501.12948},
      archivePrefix={arXiv},
      primaryClass={cs.CL},
      url={https://arxiv.org/abs/2501.12948}, 
}

@misc{deepseek-v3,
      title={DeepSeek-V3 Technical Report}, 
      author={DeepSeek-AI},
      year={2024},
      eprint={2412.19437},
      archivePrefix={arXiv},
      primaryClass={cs.CL},
      url={https://arxiv.org/abs/2412.19437}, 
}

@article{guan-etal-2022-lot,
    title = "{LOT}: A Story-Centric Benchmark for Evaluating {C}hinese Long Text Understanding and Generation",
    author = "Guan, Jian  and
      Feng, Zhuoer  and
      Chen, Yamei  and
      He, Ruilin  and
      Mao, Xiaoxi  and
      Fan, Changjie  and
      Huang, Minlie",
    editor = "Roark, Brian  and
      Nenkova, Ani",
    journal = "Transactions of the Association for Computational Linguistics",
    volume = "10",
    year = "2022",
    address = "Cambridge, MA",
    publisher = "MIT Press",
    url = "https://aclanthology.org/2022.tacl-1.25/",
    doi = "10.1162/tacl_a_00469",
    pages = "434--451",
}

@misc{zhu2023judgelmfinetunedlargelanguage,
      title={JudgeLM: Fine-tuned Large Language Models are Scalable Judges}, 
      author={Lianghui Zhu and Xinggang Wang and Xinlong Wang},
      year={2023},
      eprint={2310.17631},
      archivePrefix={arXiv},
      primaryClass={cs.CL},
      url={https://arxiv.org/abs/2310.17631}, 
}

@inproceedings{
    wang2024pandalm,
    title={Panda{LM}: An Automatic Evaluation Benchmark for {LLM} Instruction Tuning Optimization},
    author={Yidong Wang and Zhuohao Yu and Wenjin Yao and Zhengran Zeng and Linyi Yang and Cunxiang Wang and Hao Chen and Chaoya Jiang and Rui Xie and Jindong Wang and Xing Xie and Wei Ye and Shikun Zhang and Yue Zhang},
    booktitle={The Twelfth International Conference on Learning Representations},
    year={2024},
    url={https://openreview.net/forum?id=5Nn2BLV7SB}
    }

@inproceedings{liu-etal-2023-geval,
    title = "{G}-Eval: {NLG} Evaluation using Gpt-4 with Better Human Alignment",
    author = "Liu, Yang  and
      Iter, Dan  and
      Xu, Yichong  and
      Wang, Shuohang  and
      Xu, Ruochen  and
      Zhu, Chenguang",
    editor = "Bouamor, Houda  and
      Pino, Juan  and
      Bali, Kalika",
    booktitle = "Proceedings of the 2023 Conference on Empirical Methods in Natural Language Processing",
    month = dec,
    year = "2023",
    address = "Singapore",
    publisher = "Association for Computational Linguistics",
    url = "https://aclanthology.org/2023.emnlp-main.153/",
    doi = "10.18653/v1/2023.emnlp-main.153",
    pages = "2511--2522",
}

@inproceedings{deutsch-etal-2022-on-the-limitations-of-reference-free-evaluation,
    title = "On the Limitations of Reference-Free Evaluations of Generated Text",
    author = "Deutsch, Daniel  and
      Dror, Rotem  and
      Roth, Dan",
    editor = "Goldberg, Yoav  and
      Kozareva, Zornitsa  and
      Zhang, Yue",
    booktitle = "Proceedings of the 2022 Conference on Empirical Methods in Natural Language Processing",
    month = dec,
    year = "2022",
    address = "Abu Dhabi, United Arab Emirates",
    publisher = "Association for Computational Linguistics",
    url = "https://aclanthology.org/2022.emnlp-main.753/",
    doi = "10.18653/v1/2022.emnlp-main.753",
    pages = "10960--10977",
}

@inproceedings{wang-etal-2024-large-language-models-are-not-fair-evaluators,
    title = "Large Language Models are not Fair Evaluators",
    author = "Wang, Peiyi  and
      Li, Lei  and
      Chen, Liang  and
      Cai, Zefan  and
      Zhu, Dawei  and
      Lin, Binghuai  and
      Cao, Yunbo  and
      Kong, Lingpeng  and
      Liu, Qi  and
      Liu, Tianyu  and
      Sui, Zhifang",
    editor = "Ku, Lun-Wei  and
      Martins, Andre  and
      Srikumar, Vivek",
    booktitle = "Proceedings of the 62nd Annual Meeting of the Association for Computational Linguistics (Volume 1: Long Papers)",
    month = aug,
    year = "2024",
    address = "Bangkok, Thailand",
    publisher = "Association for Computational Linguistics",
    url = "https://aclanthology.org/2024.acl-long.511/",
    doi = "10.18653/v1/2024.acl-long.511",
    pages = "9440--9450",
}

@inproceedings{
    kim2024prometheus,
    title={Prometheus: Inducing Fine-Grained Evaluation Capability in Language Models},
    author={Seungone Kim and Jamin Shin and Yejin Cho and Joel Jang and Shayne Longpre and Hwaran Lee and Sangdoo Yun and Seongjin Shin and Sungdong Kim and James Thorne and Minjoon Seo},
    booktitle={The Twelfth International Conference on Learning Representations},
    year={2024},
    url={https://openreview.net/forum?id=8euJaTveKw}
}

@misc{kim2024biggenbenchprincipledbenchmark,
      title={The BiGGen Bench: A Principled Benchmark for Fine-grained Evaluation of Language Models with Language Models}, 
      author={Seungone Kim and Juyoung Suk and Ji Yong Cho and Shayne Longpre and Chaeeun Kim and Dongkeun Yoon and Guijin Son and Yejin Cho and Sheikh Shafayat and Jinheon Baek and Sue Hyun Park and Hyeonbin Hwang and Jinkyung Jo and Hyowon Cho and Haebin Shin and Seongyun Lee and Hanseok Oh and Noah Lee and Namgyu Ho and Se June Joo and Miyoung Ko and Yoonjoo Lee and Hyungjoo Chae and Jamin Shin and Joel Jang and Seonghyeon Ye and Bill Yuchen Lin and Sean Welleck and Graham Neubig and Moontae Lee and Kyungjae Lee and Minjoon Seo},
      year={2024},
      eprint={2406.05761},
      archivePrefix={arXiv},
      primaryClass={cs.CL},
      url={https://arxiv.org/abs/2406.05761}, 
}

@article{
    liang2023helm,
    title={Holistic Evaluation of Language Models},
    author={Percy Liang and Rishi Bommasani and Tony Lee and Dimitris Tsipras and Dilara Soylu and Michihiro Yasunaga and Yian Zhang and Deepak Narayanan and Yuhuai Wu and Ananya Kumar and Benjamin Newman and Binhang Yuan and Bobby Yan and Ce Zhang and Christian Alexander Cosgrove and Christopher D Manning and Christopher Re and Diana Acosta-Navas and Drew Arad Hudson and Eric Zelikman and Esin Durmus and Faisal Ladhak and Frieda Rong and Hongyu Ren and Huaxiu Yao and Jue WANG and Keshav Santhanam and Laurel Orr and Lucia Zheng and Mert Yuksekgonul and Mirac Suzgun and Nathan Kim and Neel Guha and Niladri S. Chatterji and Omar Khattab and Peter Henderson and Qian Huang and Ryan Andrew Chi and Sang Michael Xie and Shibani Santurkar and Surya Ganguli and Tatsunori Hashimoto and Thomas Icard and Tianyi Zhang and Vishrav Chaudhary and William Wang and Xuechen Li and Yifan Mai and Yuhui Zhang and Yuta Koreeda},
    journal={Transactions on Machine Learning Research},
    issn={2835-8856},
    year={2023},
    url={https://openreview.net/forum?id=iO4LZibEqW},
    note={Featured Certification, Expert Certification}
}

@inproceedings{lin-2004-rouge,
    title = "{ROUGE}: A Package for Automatic Evaluation of Summaries",
    author = "Lin, Chin-Yew",
    booktitle = "Text Summarization Branches Out",
    month = jul,
    year = "2004",
    address = "Barcelona, Spain",
    publisher = "Association for Computational Linguistics",
    url = "https://aclanthology.org/W04-1013/",
    pages = "74--81"
}

@inproceedings{papineni-etal-2002-bleu,
    title = "{B}leu: a Method for Automatic Evaluation of Machine Translation",
    author = "Papineni, Kishore  and
      Roukos, Salim  and
      Ward, Todd  and
      Zhu, Wei-Jing",
    editor = "Isabelle, Pierre  and
      Charniak, Eugene  and
      Lin, Dekang",
    booktitle = "Proceedings of the 40th Annual Meeting of the Association for Computational Linguistics",
    month = jul,
    year = "2002",
    address = "Philadelphia, Pennsylvania, USA",
    publisher = "Association for Computational Linguistics",
    url = "https://aclanthology.org/P02-1040/",
    doi = "10.3115/1073083.1073135",
    pages = "311--318"
}

@inproceedings{guan-etal-2021-openmeva,
    title = "{O}pen{MEVA}: A Benchmark for Evaluating Open-ended Story Generation Metrics",
    author = "Guan, Jian  and
      Zhang, Zhexin  and
      Feng, Zhuoer  and
      Liu, Zitao  and
      Ding, Wenbiao  and
      Mao, Xiaoxi  and
      Fan, Changjie  and
      Huang, Minlie",
    editor = "Zong, Chengqing  and
      Xia, Fei  and
      Li, Wenjie  and
      Navigli, Roberto",
    booktitle = "Proceedings of the 59th Annual Meeting of the Association for Computational Linguistics and the 11th International Joint Conference on Natural Language Processing (Volume 1: Long Papers)",
    month = aug,
    year = "2021",
    address = "Online",
    publisher = "Association for Computational Linguistics",
    url = "https://aclanthology.org/2021.acl-long.500/",
    doi = "10.18653/v1/2021.acl-long.500",
    pages = "6394--6407",
}

@inproceedings{ke-etal-2024-critiquellm,
    title = "{C}ritique{LLM}: Towards an Informative Critique Generation Model for Evaluation of Large Language Model Generation",
    author = "Ke, Pei  and
      Wen, Bosi  and
      Feng, Andrew  and
      Liu, Xiao  and
      Lei, Xuanyu  and
      Cheng, Jiale  and
      Wang, Shengyuan  and
      Zeng, Aohan  and
      Dong, Yuxiao  and
      Wang, Hongning  and
      Tang, Jie  and
      Huang, Minlie",
    editor = "Ku, Lun-Wei  and
      Martins, Andre  and
      Srikumar, Vivek",
    booktitle = "Proceedings of the 62nd Annual Meeting of the Association for Computational Linguistics (Volume 1: Long Papers)",
    month = aug,
    year = "2024",
    address = "Bangkok, Thailand",
    publisher = "Association for Computational Linguistics",
    url = "https://aclanthology.org/2024.acl-long.704/",
    doi = "10.18653/v1/2024.acl-long.704",
    pages = "13034--13054",
}

@inproceedings{zhu-etal-2024-multilingual,
    title = "Multilingual Machine Translation with Large Language Models: Empirical Results and Analysis",
    author = "Zhu, Wenhao  and
      Liu, Hongyi  and
      Dong, Qingxiu  and
      Xu, Jingjing  and
      Huang, Shujian  and
      Kong, Lingpeng  and
      Chen, Jiajun  and
      Li, Lei",
    editor = "Duh, Kevin  and
      Gomez, Helena  and
      Bethard, Steven",
    booktitle = "Findings of the Association for Computational Linguistics: NAACL 2024",
    month = jun,
    year = "2024",
    address = "Mexico City, Mexico",
    publisher = "Association for Computational Linguistics",
    url = "https://aclanthology.org/2024.findings-naacl.176/",
    doi = "10.18653/v1/2024.findings-naacl.176",
    pages = "2765--2781",
}

@misc{basyal2023textsummarizationusinglarge,
      title={Text Summarization Using Large Language Models: A Comparative Study of MPT-7b-instruct, Falcon-7b-instruct, and OpenAI Chat-GPT Models}, 
      author={Lochan Basyal and Mihir Sanghvi},
      year={2023},
      eprint={2310.10449},
      archivePrefix={arXiv},
      primaryClass={cs.CL},
      url={https://arxiv.org/abs/2310.10449}, 
}

@misc{rafailov2024dpo,
      title={Direct Preference Optimization: Your Language Model is Secretly a Reward Model}, 
      author={Rafael Rafailov and Archit Sharma and Eric Mitchell and Stefano Ermon and Christopher D. Manning and Chelsea Finn},
      year={2024},
      eprint={2305.18290},
      archivePrefix={arXiv},
      primaryClass={cs.LG},
      url={https://arxiv.org/abs/2305.18290}, 
}

@inproceedings{ouyang2022training,
 author = {Ouyang, Long and Wu, Jeffrey and Jiang, Xu and Almeida, Diogo and Wainwright, Carroll and Mishkin, Pamela and Zhang, Chong and Agarwal, Sandhini and Slama, Katarina and Ray, Alex and Schulman, John and Hilton, Jacob and Kelton, Fraser and Miller, Luke and Simens, Maddie and Askell, Amanda and Welinder, Peter and Christiano, Paul F and Leike, Jan and Lowe, Ryan},
 booktitle = {Advances in Neural Information Processing Systems},
 editor = {S. Koyejo and S. Mohamed and A. Agarwal and D. Belgrave and K. Cho and A. Oh},
 pages = {27730--27744},
 publisher = {Curran Associates, Inc.},
 title = {Training language models to follow instructions with human feedback},
 url = {https://proceedings.neurips.cc/paper_files/paper/2022/file/b1efde53be364a73914f58805a001731-Paper-Conference.pdf},
 volume = {35},
 year = {2022}
}

@misc{llmfictionworldview,
      title={Assessing Language Models' Worldview for Fiction Generation}, 
      author={Aisha Khatun and Daniel G. Brown},
      year={2024},
      eprint={2408.07904},
      archivePrefix={arXiv},
      primaryClass={cs.CL},
      url={https://arxiv.org/abs/2408.07904}, 
}

@misc{chatgpt3.5,
  author = {OpenAI},
  title = {Introducing ChatGPT},
  year = 2022,
  url = {https://openai.com/index/chatgpt/},
  note = {Accessed: 2024-11-30}
}

@misc{chatglm,
      title={ChatGLM: A Family of Large Language Models from GLM-130B to GLM-4 All Tools}, 
      author={Team-GLM},
      year={2024},
      eprint={2406.12793},
      archivePrefix={arXiv},
      primaryClass={cs.CL},
      url={https://arxiv.org/abs/2406.12793}, 
}

@misc{gemini1.5,
      title={Gemini 1.5: Unlocking multimodal understanding across millions of tokens of context}, 
      author={Gemini-Team},
      year={2024},
      eprint={2403.05530},
      archivePrefix={arXiv},
      primaryClass={cs.CL},
      url={https://arxiv.org/abs/2403.05530}, 
}

@misc{multimodallongstory,
      title={SEED-Story: Multimodal Long Story Generation with Large Language Model}, 
      author={Shuai Yang and Yuying Ge and Yang Li and Yukang Chen and Yixiao Ge and Ying Shan and Yingcong Chen},
      year={2024},
      eprint={2407.08683},
      archivePrefix={arXiv},
      primaryClass={cs.CV},
      url={https://arxiv.org/abs/2407.08683}, 
}

@inproceedings{alignbench,
    title = "{A}lign{B}ench: Benchmarking {C}hinese Alignment of Large Language Models",
    author = "Liu, Xiao  and
      Lei, Xuanyu  and
      Wang, Shengyuan  and
      Huang, Yue  and
      Feng, Andrew  and
      Wen, Bosi  and
      Cheng, Jiale  and
      Ke, Pei  and
      Xu, Yifan  and
      Tam, Weng Lam  and
      Zhang, Xiaohan  and
      Sun, Lichao  and
      Gu, Xiaotao  and
      Wang, Hongning  and
      Zhang, Jing  and
      Huang, Minlie  and
      Dong, Yuxiao  and
      Tang, Jie",
    editor = "Ku, Lun-Wei  and
      Martins, Andre  and
      Srikumar, Vivek",
    booktitle = "Proceedings of the 62nd Annual Meeting of the Association for Computational Linguistics (Volume 1: Long Papers)",
    month = aug,
    year = "2024",
    address = "Bangkok, Thailand",
    publisher = "Association for Computational Linguistics",
    url = "https://aclanthology.org/2024.acl-long.624/",
    doi = "10.18653/v1/2024.acl-long.624",
    pages = "11621--11640",
}

@inproceedings{longform,
    title = "{L}ong{F}orm: Effective Instruction Tuning with Reverse Instructions",
    author = {K{\"o}ksal, Abdullatif  and
      Schick, Timo  and
      Korhonen, Anna  and
      Schuetze, Hinrich},
    editor = "Al-Onaizan, Yaser  and
      Bansal, Mohit  and
      Chen, Yun-Nung",
    booktitle = "Findings of the Association for Computational Linguistics: EMNLP 2024",
    month = nov,
    year = "2024",
    address = "Miami, Florida, USA",
    publisher = "Association for Computational Linguistics",
    url = "https://aclanthology.org/2024.findings-emnlp.414/",
    doi = "10.18653/v1/2024.findings-emnlp.414",
    pages = "7056--7078",
}

@inproceedings{writingprompts,
    title = "Hierarchical Neural Story Generation",
    author = "Fan, Angela  and
      Lewis, Mike  and
      Dauphin, Yann",
    editor = "Gurevych, Iryna  and
      Miyao, Yusuke",
    booktitle = "Proceedings of the 56th Annual Meeting of the Association for Computational Linguistics (Volume 1: Long Papers)",
    month = jul,
    year = "2018",
    address = "Melbourne, Australia",
    publisher = "Association for Computational Linguistics",
    url = "https://aclanthology.org/P18-1082/",
    doi = "10.18653/v1/P18-1082",
    pages = "889--898",
}

@inproceedings{rocstories,
    title = "A Corpus and Cloze Evaluation for Deeper Understanding of Commonsense Stories",
    author = "Mostafazadeh, Nasrin  and
      Chambers, Nathanael  and
      He, Xiaodong  and
      Parikh, Devi  and
      Batra, Dhruv  and
      Vanderwende, Lucy  and
      Kohli, Pushmeet  and
      Allen, James",
    editor = "Knight, Kevin  and
      Nenkova, Ani  and
      Rambow, Owen",
    booktitle = "Proceedings of the 2016 Conference of the North {A}merican Chapter of the Association for Computational Linguistics: Human Language Technologies",
    month = jun,
    year = "2016",
    address = "San Diego, California",
    publisher = "Association for Computational Linguistics",
    url = "https://aclanthology.org/N16-1098/",
    doi = "10.18653/v1/N16-1098",
    pages = "839--849"
}

@inproceedings{mtbench,
 author = {Zheng, Lianmin and Chiang, Wei-Lin and Sheng, Ying and Zhuang, Siyuan and Wu, Zhanghao and Zhuang, Yonghao and Lin, Zi and Li, Zhuohan and Li, Dacheng and Xing, Eric and Zhang, Hao and Gonzalez, Joseph E and Stoica, Ion},
 booktitle = {Advances in Neural Information Processing Systems},
 editor = {A. Oh and T. Naumann and A. Globerson and K. Saenko and M. Hardt and S. Levine},
 pages = {46595--46623},
 publisher = {Curran Associates, Inc.},
 title = {Judging LLM-as-a-Judge with MT-Bench and Chatbot Arena},
 url = {https://proceedings.neurips.cc/paper_files/paper/2023/file/91f18a1287b398d378ef22505bf41832-Paper-Datasets_and_Benchmarks.pdf},
 volume = {36},
 year = {2023}
}

\newpage

\appendix

% \section{Example Appendix}\label{sec:appendix}

\section{Additional Information in Data Preparation}

\subsection{Crawling Sources} \label{sec:appdx-data-source}

For Chinese part, we crawled data from the following high quality and reputable sources:

\begin{enumerate}
    \item \textit{Chinese Writer Website} (\textbf{CN Writer}, 中国作家网) \footnote{https://www.chinawriter.com.cn/} : this site collects all publishable fictions, proses, poets from professional writers from China, powered by Chinese Association of Writer. The writings are all professionally written. The total number of raw data is approximately 5k.
    \item \textit{The pivot website for example essays} (\textbf{PW4ES}, 第一范文网) \footnote{https://www.diyifanwen.com/} : this site collects numerous functional writing sources, such as contracts, plans, conclusions, thoughts, speeches and deliveries etc. The writings are of high quality and they serve as examples for learners. The total number of raw data is approximately 30k. 
    \item \textit{September for example essays} (\textbf{SeptES}, 九月范文网) \footnote{https://www.chinesejy.com/}: this site complements to the above sites, with additional functional writings. The writings are of high quality and they serve as examples for learners. The total number of raw data is approximately 30k. 
    \item \textit{Zhejiang Publicity} (\textbf{ZJPub, 浙江宣传}) \footnote{https://zjnews.zjol.com.cn/zjxc/} : this site collects numerous argumentative, critics targeting at social/historical/cultural affairs. These articles are targeting electronic self-media readers, and are written by professional newspaper writers. The total number of raw data is approximately 10k.
    \item \textit{Site for Officials} (\textbf{Officials}, 公文网) \footnote{https://www.gongwen.com.cn/}: this site collects examples for official articles writings, including propaganda, deliveries, announcements, etc. We purchased the articles from the site instead of crawling for its commercial use. The articles are written by expert civil servants from the government, and is of high quality. The total number of raw data is approximately 20k.
\end{enumerate}

For English part, we crawled data from the following high quality and reputable sources:

\begin{enumerate}
    \item \textit{American Rhetoric}\footnote{https://www.americanrhetoric.com/top100speechesall.html}: This website records famous speeches in American history, including historical speeches as well as parliamentary speeches and questions.
    \item \textit{Obook}\footnote{https://www.obooko.com/}: This website records numerous English published books with a wide range of genres, including fiction, prose, poem, novel across 16 century to contemporary.
    \item \textit{IvyPanda}\footnote{https://ivypanda.com/}. This website serves top level example essays across 32 topics, including art, business, culture, environment, history, music and so on. We use huggingface dataset\texttt{qwedsacf/ivypanda-essays} \footnote{https://huggingface.co/datasets/qwedsacf/ivypanda-essays} from the same source and the number is approximately 100K.
\end{enumerate}

\subsection{Generalizability to English} \label{sec:appdx-multilingual}

To validate the generalizability of \method{}, we constructed an English subset of \bench{} comprising 852 samples from three high-quality sources (detailed in Section \ref{sec:appdx-data-source}). The curation process strictly mirrors that of the Chinese dataset. While conducting large-scale human evaluation for complex writing tasks is highly time-consuming (e.g., our primary Chinese evaluation required 36 experts over two months), we conducted a preliminary, double-checked human evaluation (IAA = 0.72) on a subset of 17 English instructions evaluated across 9 LLMs. As shown in Table~\ref{tab:english_preliminary}, the Pearson correlations demonstrate patterns highly consistent with our main Chinese experiments, indicating that our evaluation framework adapts to other languages. We plan to finalize the multi-lingual subsets in future versions.

\begin{table}[h]
    \centering
    \small
    \begin{tabular}{lcccc}
    \toprule
    \textbf{Metric} & \textbf{Completion} & \textbf{Guide} & \textbf{Open} & \textbf{Overall} \\
    \midrule
    BLEU-1 & 0.74* & 0.65* & 0.69 & 0.71* \\
    ROUGE-L & 0.85* & 0.34 & 0.48 & 0.52 \\
    Rubrics & 0.76* & 0.85* & 0.81 & 0.81* \\
    Auto Plan & 0.65* & 0.71 & 0.54 & 0.69 \\
    ToW & 0.82* & 0.82 & 0.86* & 0.85* \\
    \bottomrule
    \end{tabular}
    \caption{Preliminary Pearson correlation results on the English subset. * indicates $p < 0.05$.}
    \label{tab:english_preliminary}
\end{table}

\subsection{Included Writing Genres} \label{sec:genres}

\noindent\textbf{Fiction}: Fiction focuses on imaginative narratives, emphasizing character development, plot structure, and environmental depiction. It reflects social realities or human emotions, with a focus on details and conflicts driving the story forward. \\

\noindent\textbf{Poetry}: Poetry is characterized by line breaks, condensed language, and symbolic imagery, with an emphasis on rhythm and sound, as well as the intense concentration of emotion and thought. \\

\noindent\textbf{Prose}: Prose encompasses descriptive and imaginative writing without the constraints of poetic structure. It often explores themes and ideas in clear, expressive language, engaging the reader in a reflective or emotional experience. \\

\noindent\textbf{Essay}: A creative essay blends personal reflection and artistic style. It is often subjective, descriptive, and exploratory, focusing on an idea, experience, or insight in a unique and engaging way. \\

\noindent\textbf{Argumentative}: This writing builds a compelling case centered around a perspective or opinion, supported by logical reasoning or persuasive rhetoric. It seeks to convince the audience using passionate and effective arguments. \\

\noindent\textbf{Report}: A report is an objective, structured, and formal document that presents data, findings, and analysis of specific topics or activities, often following a standardized format. \\

\noindent\textbf{Summary}: Summarizing involves condensing large pieces of information into brief and concise overviews, focusing only on the key points, events, or ideas introduced in the original text. \\

\noindent\textbf{Letter}: A formal or informal written communication addressed to another person or entity, often following a clear structure that includes salutations, body content, and closing remarks. \\

\noindent\textbf{Application}: Applications are formal documents written in a specific format, expressing a request, often for employment, educational admissions, or permissions. They are brief and structured. \\

\noindent\textbf{Speech}: A speech is a prepared piece of writing meant to be spoken aloud, tailored for an audience, often persuasive or inspiring, and is structured to guide the listener through ideas or arguments. \\

\noindent\textbf{Delivery}: Delivery writing includes real-time or impromptu words, such as announcements or ceremonial addresses, meant for immediate and direct communication in specific events or contexts. \\

\noindent\textbf{Plan}: A plan outlines structured steps, timelines, or objectives to achieve a specific goal or outcome. It is often practical and formatted to organize resources and tasks effectively. \\

\noindent\textbf{Contract}: A contract is a formal, legal document outlining agreements between parties, specifying terms, responsibilities, and obligations, often in precise and enforceable language. \\

\noindent\textbf{Official}: Official writing refers to documents meant for administrative, governmental, or institutional purposes, often rigid in format and addressing formal matters or processes. \\

\subsection{Leaf Node Traits Explained}

We briefly introduce the leaf nodes traits in Table~\ref{tab:content illustration}. 

\begin{table*}[!t]
    \scriptsize
    \centering
        \resizebox{\linewidth}{!}{%
        \begin{tabular}{p{70pt}|p{250pt}|p{70pt}}
        \toprule
        \textbf{Traits}         & \textbf{Description}                                                                                                                                                        & \textbf{Rubrics}                                                                                                                                \\ \midrule
    Opening \& Ending       & Whether the opening and ending are engaging, with no abrupt stops or forced elevation of tone/plots/conclusions.                                                            & \multirow{5}{*}{\begin{tabular}[c]{@{}l@{}}1-4: worse than reference\\ 5-7: comparable to reference \\ 8-10: Superior to reference\end{tabular}} \\ \cmidrule(r){1-2}
    Language \& Rhetoric    & Using appropriate rhetoric, is the vocabulary and expression rich? Has a monotonous, list-like style been avoided?                                                          &                                                                                                                                                 \\  \cmidrule(r){1-2}
    Proper instance         & No violation of real-world knowledge. Whether proper instances are used to address argumentations.                                                                          &                                                                                                                                                 \\  \cmidrule(r){1-2}
    Argumentative \& Logics & Whether the logic in arguments, plot development, and overall writing is appropriate and coherent. Ensure smooth transitions and avoid abrupt or forced causal connections. &                                                                                                                                                 \\  \cmidrule(r){1-2}
    Emotion                 & Are the emotions effectively conveyed to the readers? Are the characters in the writing portrayed with appropriate emotional depth?                                         &                                                                                                                                                 \\ \bottomrule
    
        \end{tabular}%
        }
    
    \caption{Illustration for different traits.}
    \label{tab:content illustration}
\end{table*}

\begin{table*}[!t]
    \scriptsize
    \centering
        \resizebox{\linewidth}{!}{%
        \begin{tabular}{p{70pt}|p{70pt}|p{250pt}}
        \toprule
        \textbf{Traits}         & \textbf{Description}                                                                                                                                                        & \textbf{Rubrics}                                                                                                                                \\ \midrule
Plots           & Whether the plots are reasonable                                                                                                 & 1-4: worse than reference 5-7: comparable to reference 8-10: Superior to reference                                                                                                   \\ \midrule
Formatting      & Checking all titles, lists in the writing with Regex. Detecting Chinese Titles, markdown titles, ordered lists, unordered lists. & 0: Violation in hierarchical relations, improper unordered list in continuous texts  5. moderate titling or no titling are found  10: titling satisfies all the checks from the rules. \\ \midrule
Paragraphing    & Checking whether the paragraphs sectioning are reasonable or not.                                                                & 1-4: Disproportionate paragraphing  5-7: paragraphing  8-10: paragraphing with superior deigns                                                                                       \\ \midrule
Impression      & Inspecting whether the writing satisfies the writing instructions theme and requirements.                                        & 1-4: worse than reference 5-7: comparable to reference 8-10: Superior to reference                                                                                                   \\ \bottomrule

        \end{tabular}%
        }
    
    \caption{Illustration for Format and Impression traits.}
    \label{tab:format illustration}
\end{table*}

\section{Further Discussions}

\subsection{Discussions on different evaluators}\label{sec:evaluator}

\begin{figure}[t]
    \centering
    \includegraphics[width=\linewidth]{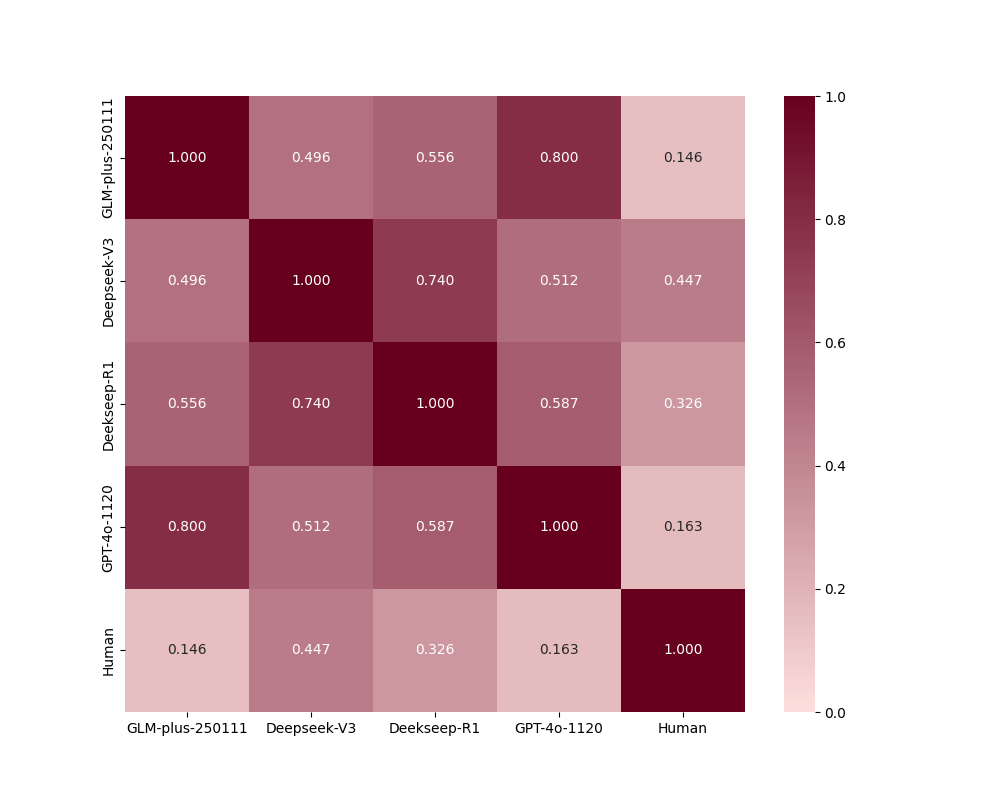}
    \caption{Pearson correlation cross evaluators and human experts.}
    \label{fig:evaluator_corr}
\end{figure}

We further analyze the influence of Judge LLMs. We select the Level II and Level III tasks and compute the sample level Pearson correlation between GLM-4, Gemini-2.0-Flash, GPT-4o-1120, DeepSeek-V3, DeepSeek-R1. We concatenate all 3 inference models (GLM, Gemini and GPT) responses score as 3 times longer vector and compute the Pearson correlation via it. Results are plotted in the form of heatmap in Figure~\ref{fig:evaluator_corr}. Results showed that DeepSeek-V3 owns the highest Pearson correlation with human judgment, while GLM and GPT share very poor correlation with humans. On the other hand, LLM evaluators all showed very high correlation with each other ($\rho > $ 0.5), indicating the common potential biases. Human experts reached $\kappa$ = 0.56 and $\rho$ = 0.67 in cross validation, confirming the gap between human and LLM Judges.

\subsection{Discussion on Reference-based and Reference-free Evaluation} \label{sec:ref-based-free}

We experimented in a refined reference-free setting (by removing the existence of reference and re-judging) and compared it to the reference-based setting with a random and evenly picked subset from \bench{} ($N$=300). We calculated the system level correlation scores with all samples from 3 tasks altogether and summarized the results in Table~\ref{tab:ref-exist-abl}.

\begin{table}[t]
    \resizebox{0.5\textwidth}{!}{%
    \begin{tabular}{@{}c|ccc|ccc@{}}
    \toprule
                      & \multicolumn{3}{c|}{Reference based}      & \multicolumn{3}{c}{Reference Free}               \\
                      & Pearson        & Spearman       & Kendall & Pearson        & Spearman       & Kendall        \\ \midrule
    DeepSeek-baseline & 0.688          & 0.717          & 0.556   & 0.592          & 0.383          & 0.278          \\
    DeepSeek-rubric   & 0.585          & 0.533          & 0.389   & 0.452          & 0.333          & 0.222          \\
    DeepSeek-ToW      & \textbf{0.748} & \textbf{0.750} & 0.556   & \textbf{0.714} & \textbf{0.733} & \textbf{0.556} \\
    Gemini-baseline   & 0.160          & 0.267          & 0.222   & 0.697          & 0.750          & 0.556          \\
    Gemini-rubric     & 0.716          & 0.733          & 0.556   & 0.423          & 0.600          & 0.389          \\
    Gemini-ToW        & \textbf{0.723} & \textbf{0.767} & 0.556   & \textbf{0.749} & \textbf{0.800} & \textbf{0.611} \\ \bottomrule
    \end{tabular}%
    }
    \caption{Reference-based evaluation and Reference-free evaluation results.}
    \label{tab:ref-exist-abl}
\end{table}

From the experiment results, \method{} still maintained high system level correlation while the baseline, rubric methods drops with the absence of reference. This indicates that chain-of-writing can judge without reference, which goes beyond the rubric scoring methods.

\subsection{Discussion on Reference Source} \label{sec:ref-source}

One of the core principles of \bench{} is the reliance on high-quality human experts and writers as references for evaluation. We investigate the feasibility and reliability of using LLM-generated texts as references and assess their credibility at the system level.

Specifically, we adopt a setting where the instructions and guiding information in \bench{} remain unchanged, but the inference output of a particular LLM is used as a 6 point reference to guide evaluation. We employ Gemini-2.0-Flash as the Evaluator and compare the results against human references as well as those generated by DeepSeek-V3, GPT-4-o3-mini, and Claude-3.5-sonnet-1022. The three models are recognized for their strong performance in writing tasks. The system-level correlations are summarized in Table~\ref{tab:ref-abl}. References derived from alternative sources generally result in lower consistency rates, whereas human references achieve significantly higher agreement. Furthermore, the \method{} demonstrates robustness across references of varying origins, indicating that its effectiveness is independent of the reference source.

\begin{table}[t]
    \resizebox{0.5\textwidth}{!}{%
    \begin{tabular}{@{}cc|ccc@{}}
    \toprule
                      & \textbf{Human} & \textbf{DeepSeek-V3} & \textbf{GPT-o3-mini} & \textbf{Claude-3.5} \\ \midrule
    DeepSeek-baseline & 0.688          & 0.477                & 0.723                & 0.756                           \\
    DeepSeek-rubric   & 0.585          & 0.451                & 0.652                & 0.646                           \\
    DeepSeek-ToW      & \textbf{0.748} & \textbf{0.607}       & \textbf{0.816}       & \textbf{0.783}                  \\ \midrule
    Gemini-baseline   & 0.160          & \textbf{0.730}       & 0.459                & 0.380                           \\
    Gemini-rubric     & 0.716          & 0.580                & 0.469                & 0.473                           \\
    Gemini-ToW        & \textbf{0.723} & 0.715                & \textbf{0.528}       & \textbf{0.552}                  \\ \bottomrule
    \end{tabular}%
    }
    \caption{Influence on system level correlation from reference sources. }
    \label{tab:ref-abl}
\end{table}
\section{Full Plots for Analysis sections}

\subsection{Plots Between Input Length, Output Length and Scores}

\begin{figure*}[!t]
    \centering
    
    % Row 1
    \begin{subfigure}[b]{0.3\textwidth}
        \centering
        \caption{\textbf{Completion}}
        \includegraphics[width=\textwidth]{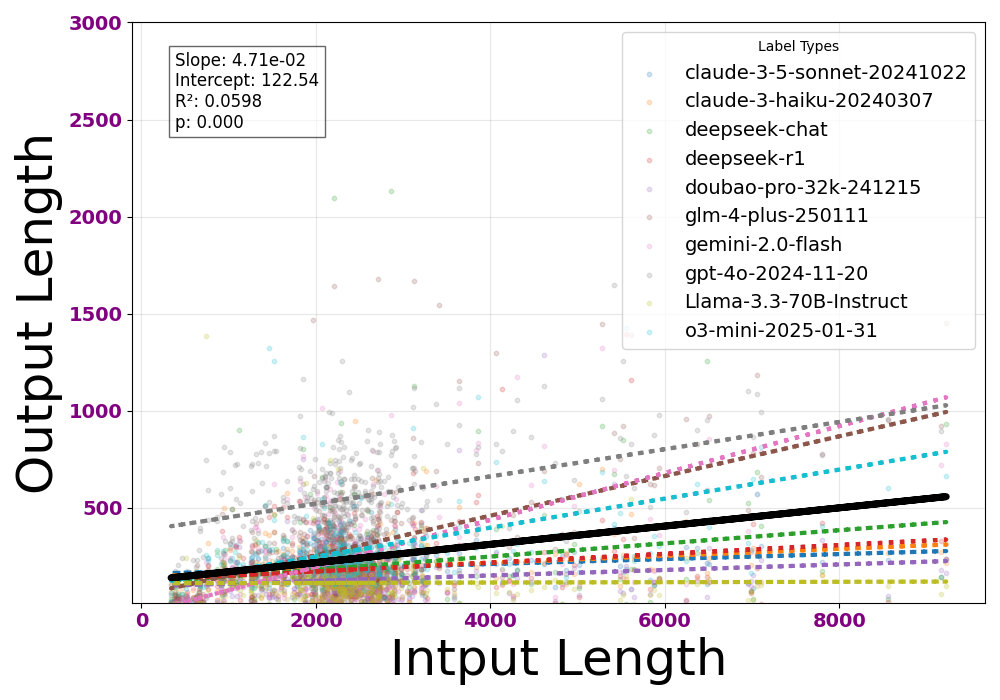} % Replace with your image path/file name
        % \caption{}
    \end{subfigure}
    \begin{subfigure}[b]{0.3\textwidth}
        \centering
        \caption{\textbf{Guide}}
        \includegraphics[width=\textwidth]{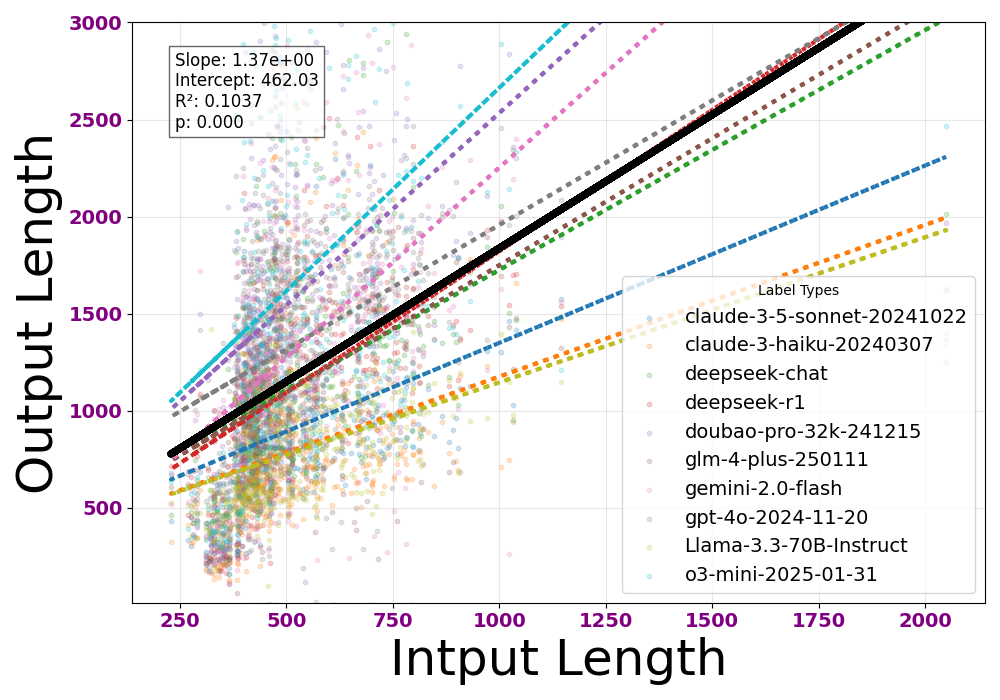}
        % \caption{Subfig 1.2}
    \end{subfigure}
    \begin{subfigure}[b]{0.3\textwidth}
        \centering
        \caption{\textbf{Open}}
        \includegraphics[width=\textwidth]{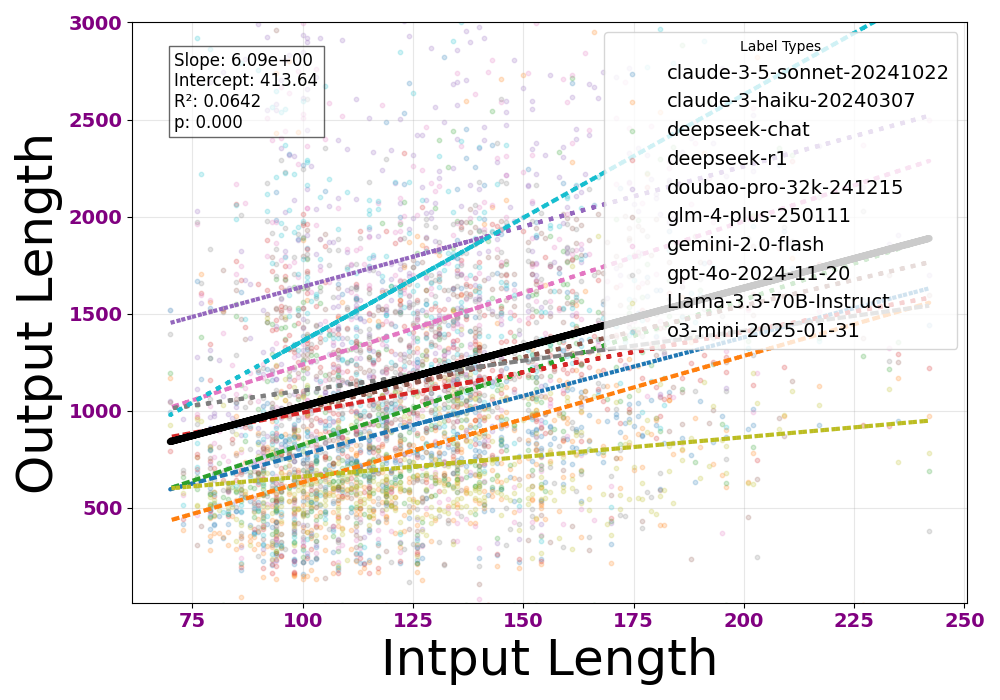}
        % \caption{Subfig 1.3}
    \end{subfigure}
    
    % Row 2
    \begin{subfigure}[b]{0.3\textwidth}
        \centering
        \includegraphics[width=\textwidth]{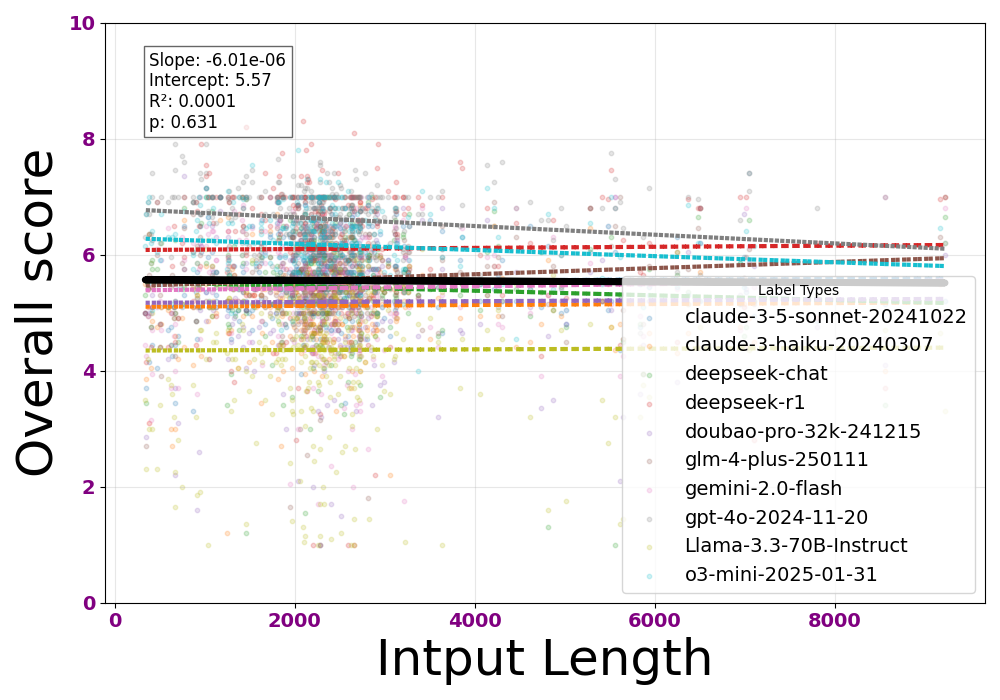}
        % \caption{Subfig 2.1}
    \end{subfigure}
    \begin{subfigure}[b]{0.3\textwidth}
        \centering
        \includegraphics[width=\textwidth]{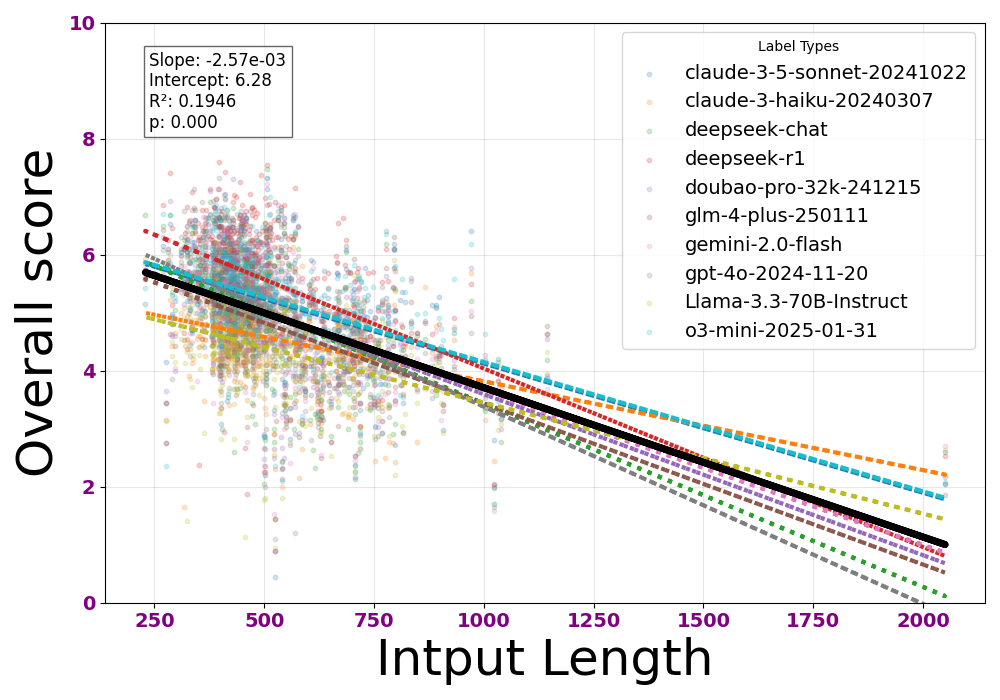}
        % \caption{Subfig 2.2}
    \end{subfigure}
    \begin{subfigure}[b]{0.3\textwidth}
        \centering
        \includegraphics[width=\textwidth]{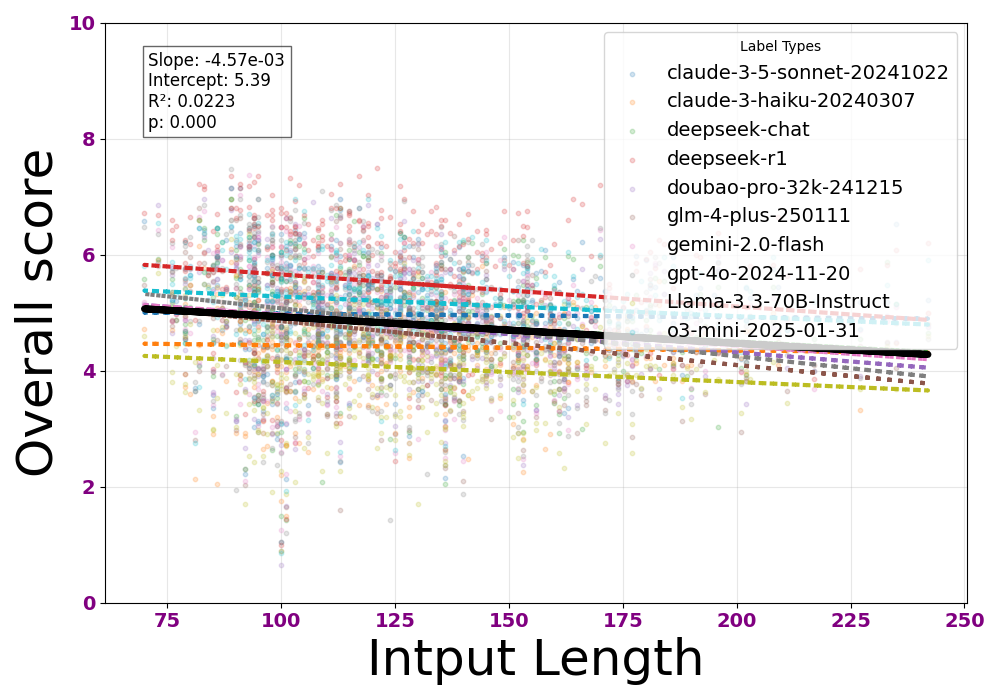}
        % \caption{Subfig 2.3}
    \end{subfigure}
    
    % Add similar rows for more figures (Rows 3, 4, 5 as needed)
    
    % Row 3
    \begin{subfigure}[b]{0.3\textwidth}
        \centering
        \includegraphics[width=\textwidth]{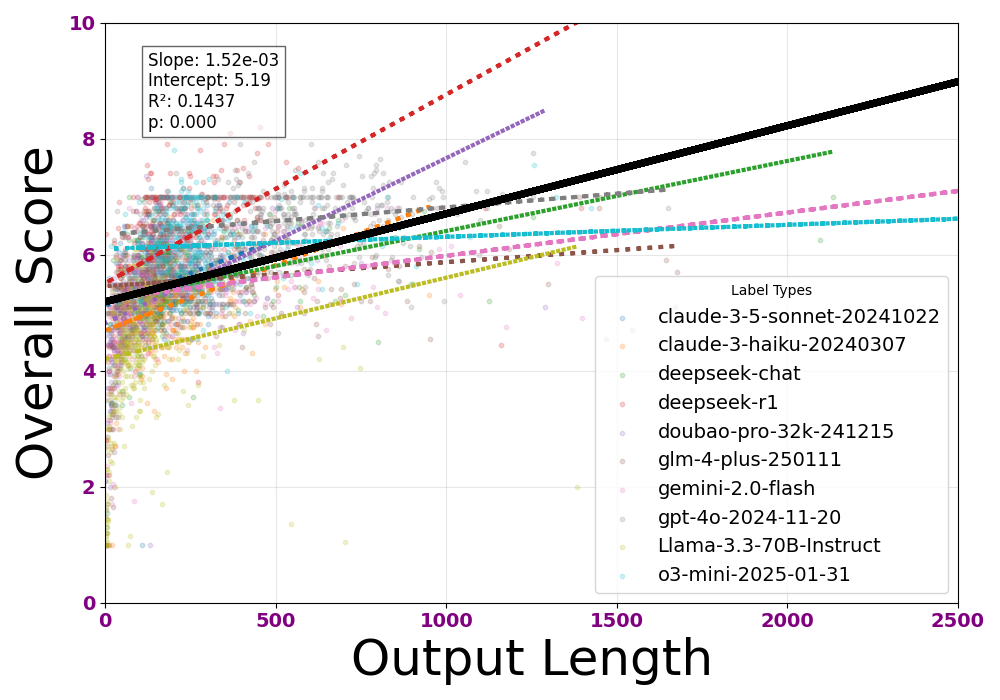}
    \end{subfigure}
    \begin{subfigure}[b]{0.3\textwidth}
        \centering
        \includegraphics[width=\textwidth]{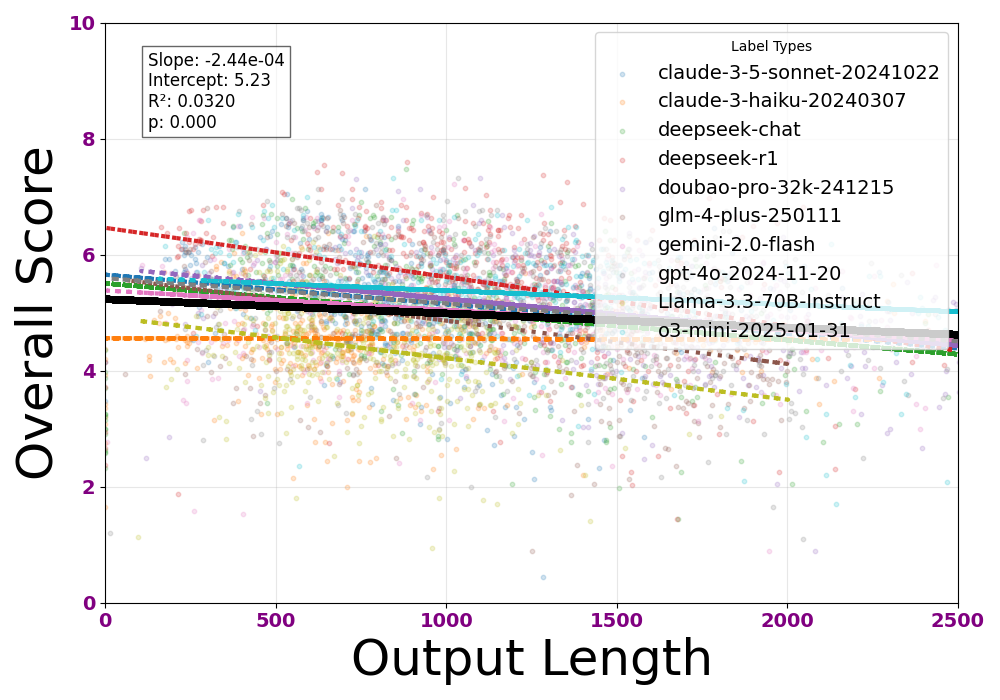}
    \end{subfigure}
    \begin{subfigure}[b]{0.3\textwidth}
        \centering
        \includegraphics[width=\textwidth]{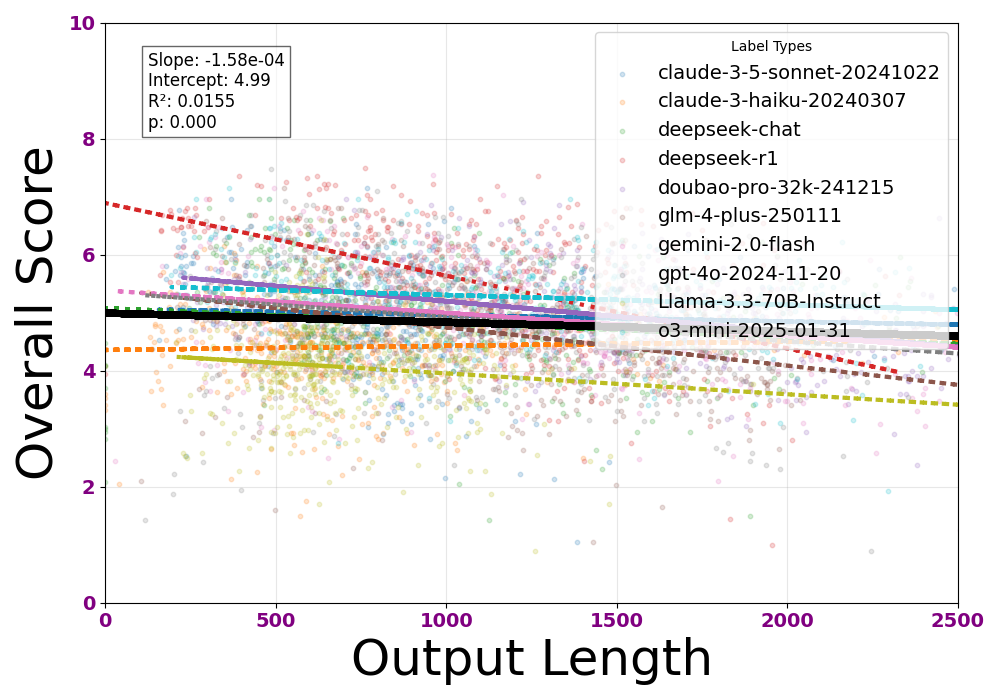}
    \end{subfigure}

    % Row 4
    \begin{subfigure}[b]{0.3\textwidth}
        \centering
        \includegraphics[width=\textwidth]{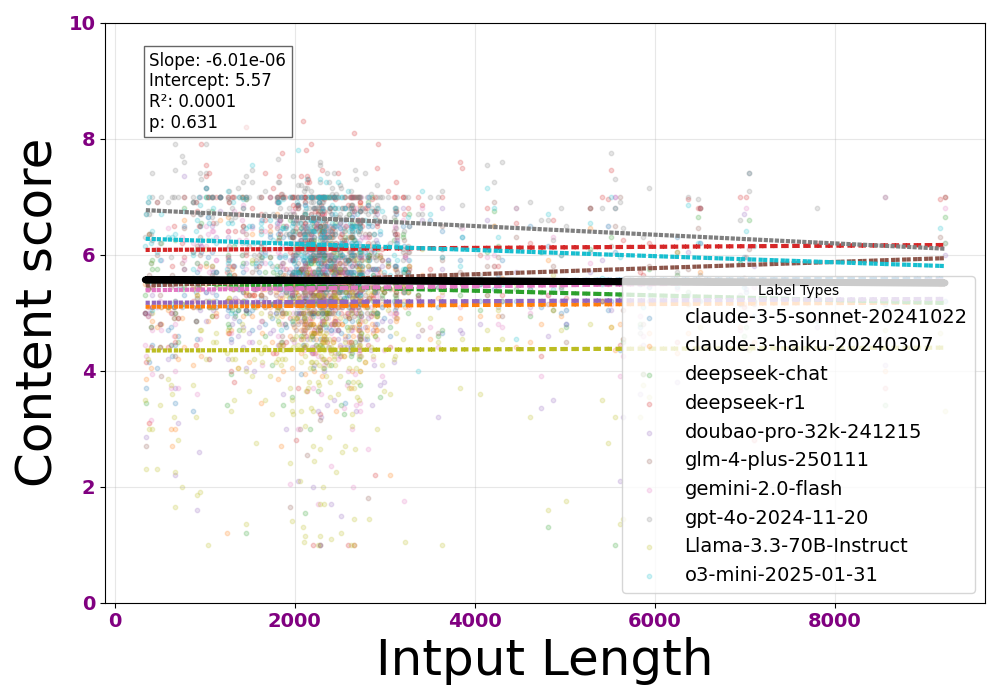}
    \end{subfigure}
    \begin{subfigure}[b]{0.3\textwidth}
        \centering
        \includegraphics[width=\textwidth]{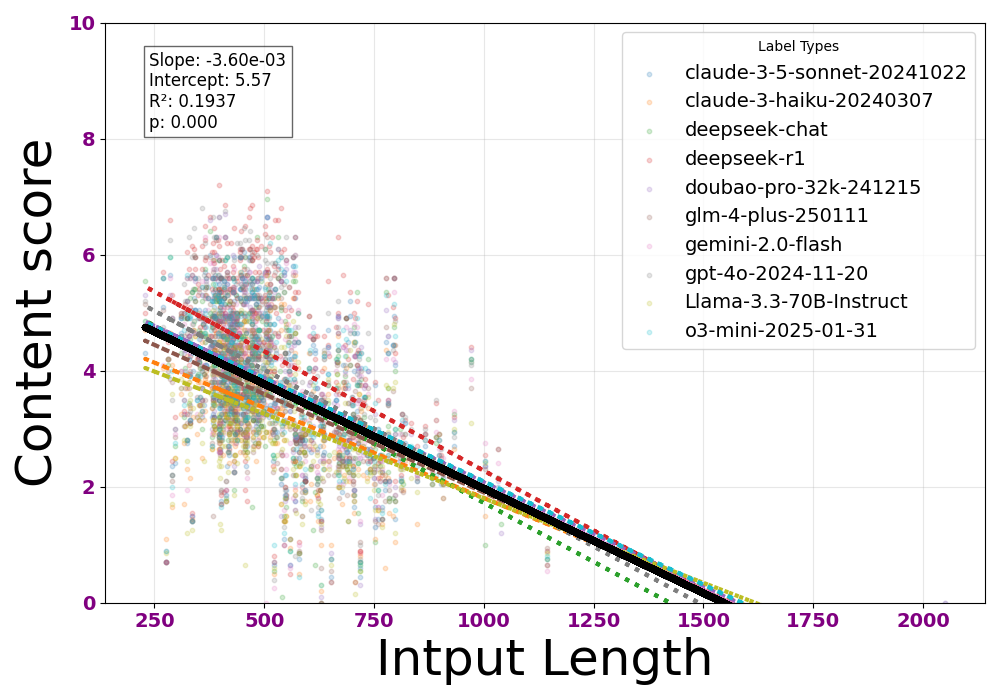}
    \end{subfigure}
    \begin{subfigure}[b]{0.3\textwidth}
        \centering
        \includegraphics[width=\textwidth]{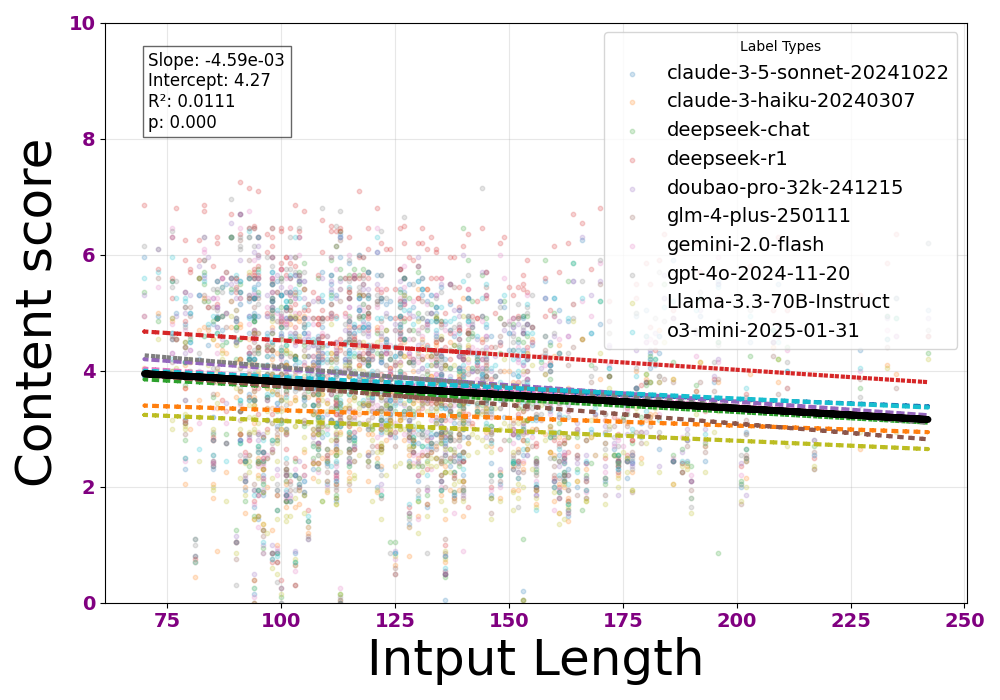}
    \end{subfigure}

    % Row 5
    \begin{subfigure}[b]{0.3\textwidth}
        \centering
        \includegraphics[width=\textwidth]{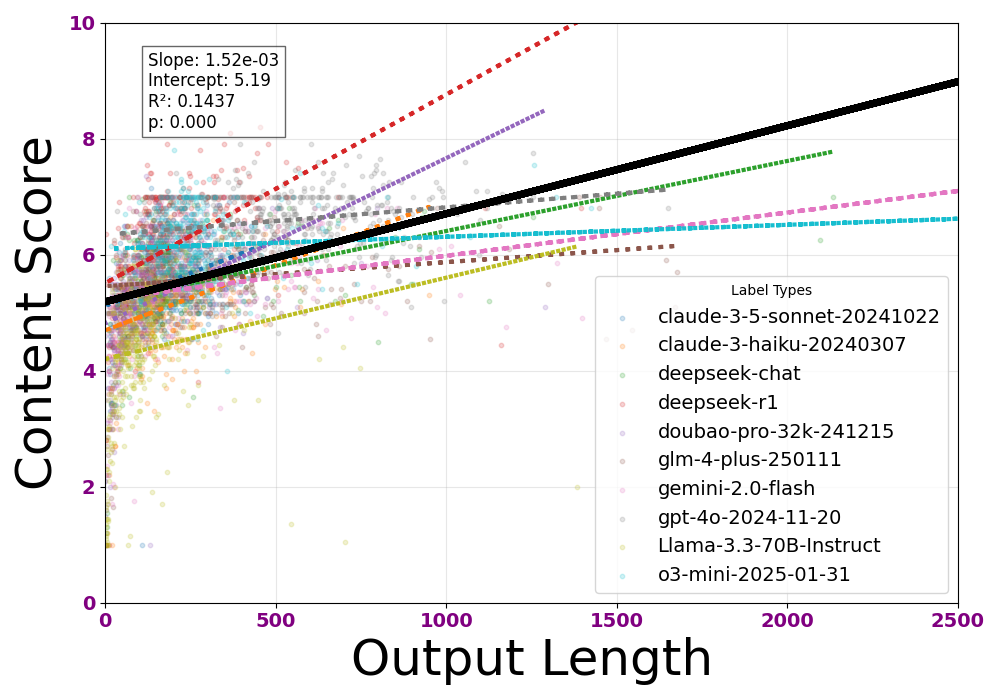}
    \end{subfigure}
    \begin{subfigure}[b]{0.3\textwidth}
        \centering
        \includegraphics[width=\textwidth]{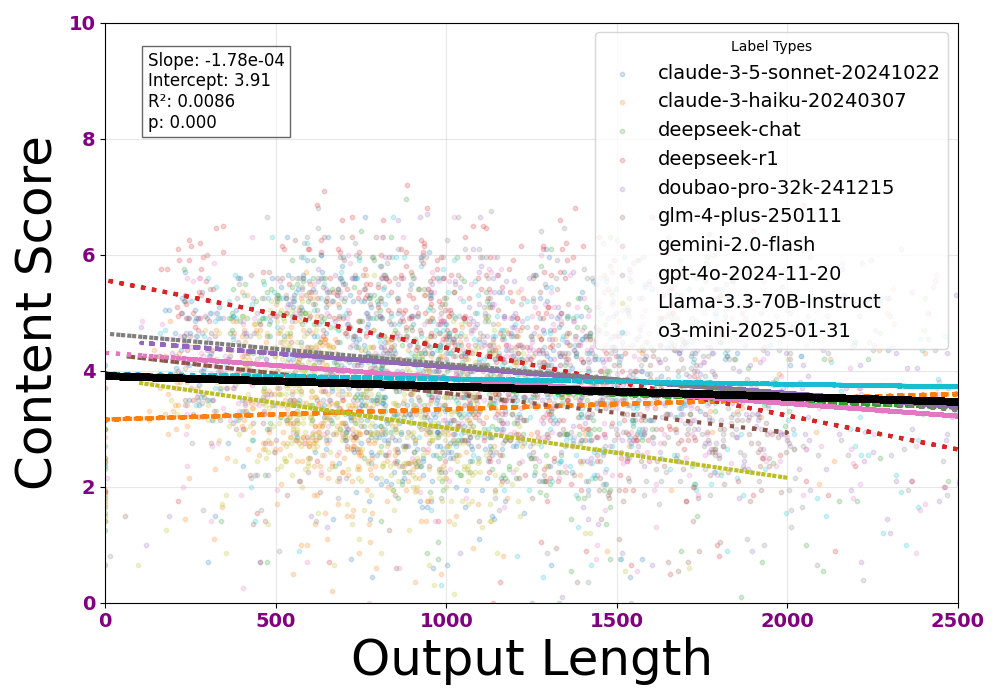}
    \end{subfigure}
    \begin{subfigure}[b]{0.3\textwidth}
        \centering
        \includegraphics[width=\textwidth]{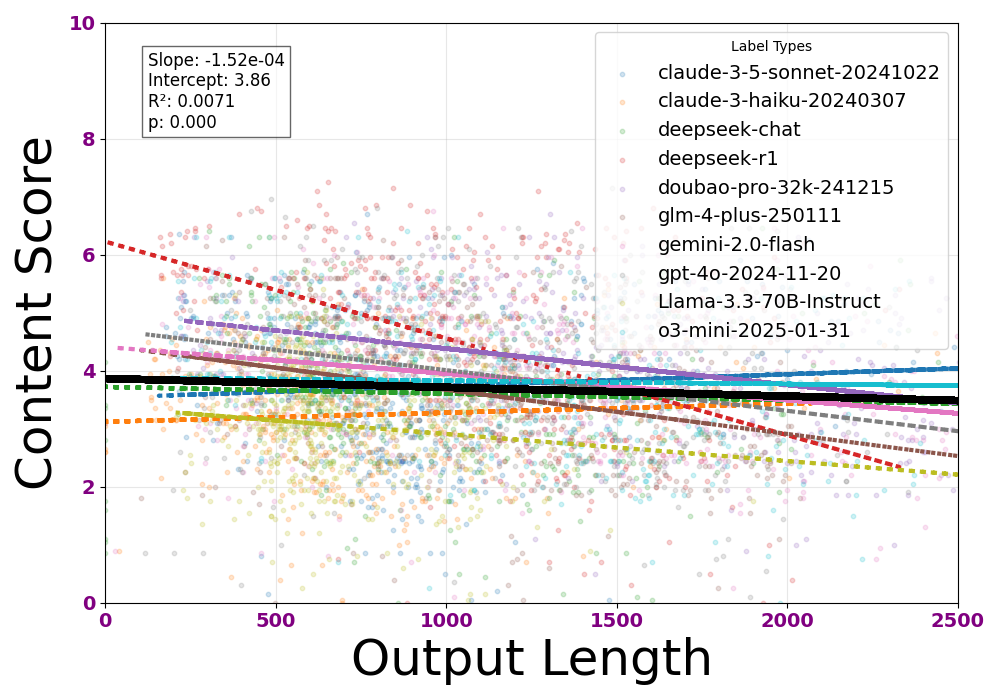}
    \end{subfigure}

    % Caption for the entire figure
    \caption{Factor Analysis between input length, output length, overall score, content score. The bold black line indicates the regression results from all LLM data points.}
    \label{fig:mimic-game}
\end{figure*}

Figure~\ref{fig:mimic-game} presents the scatter and linear regression between input length, output length, overall scores and content scores.

\subsection{Edge Weights across Multiple Genres} \label{apd:edge-weight}

\begin{figure*}[!ht]
    \centering
    
    % Row 1
    \begin{subfigure}[b]{0.52\textwidth}
        \centering
        % \caption{\textbf{C}}
        \includegraphics[width=\textwidth]{figures/weights/fiction.png} % Replace with your image path/file name
        % \caption{}
    \end{subfigure}
    \begin{subfigure}[b]{0.33\textwidth}
        \centering
        \includegraphics[width=\textwidth]{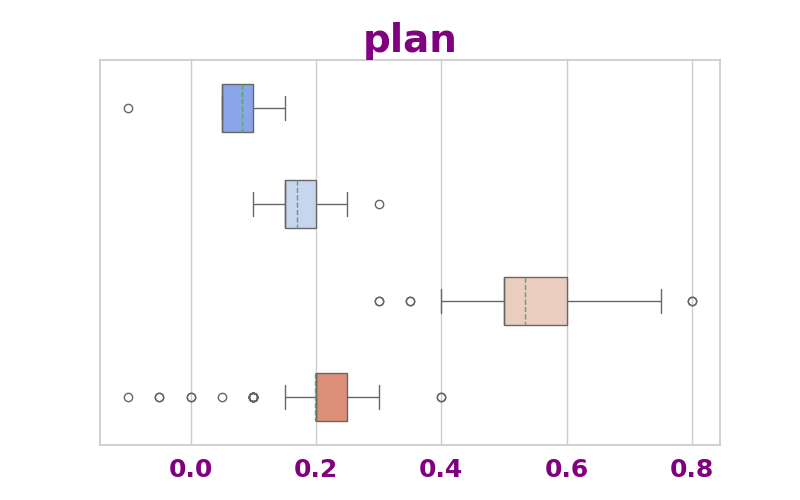}
        % \caption{Subfig 1.2}
    \end{subfigure}

    % Row 2
    \begin{subfigure}[b]{0.52\textwidth}
        \centering
        \includegraphics[width=\textwidth]{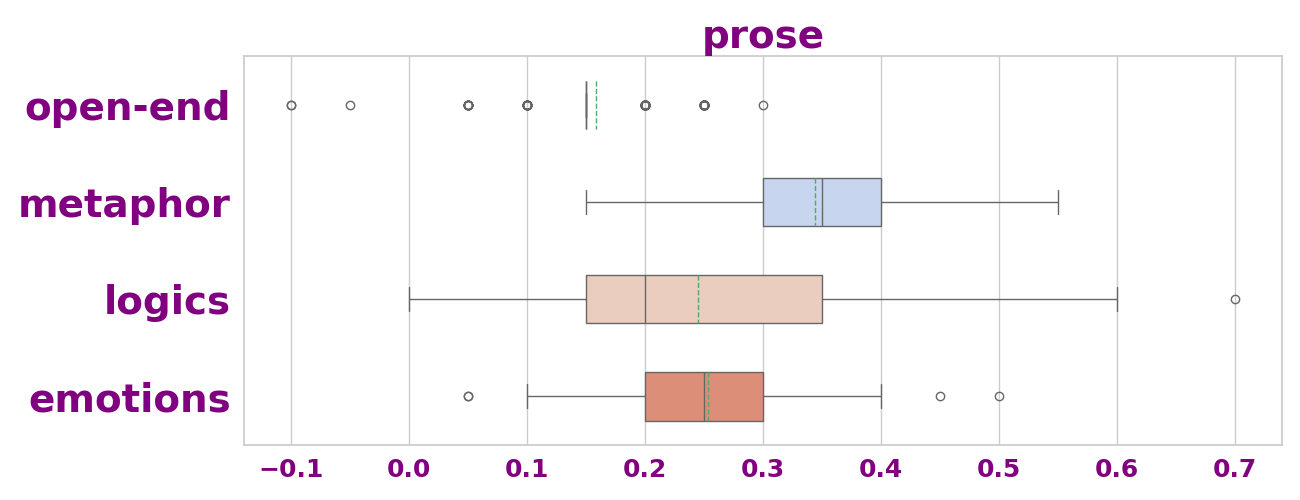}
        % \caption{Subfig 1.3}
    \end{subfigure}
    \begin{subfigure}[b]{0.33\textwidth}
        \centering
        \includegraphics[width=\textwidth]{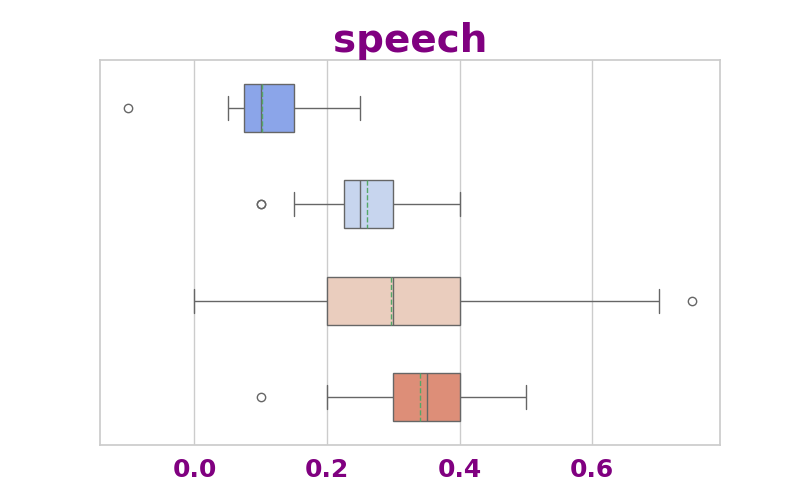}
        % \caption{Subfig 1.3}
    \end{subfigure}

    % Row 3
    \begin{subfigure}[b]{0.52\textwidth}
        \centering
        \includegraphics[width=\textwidth]{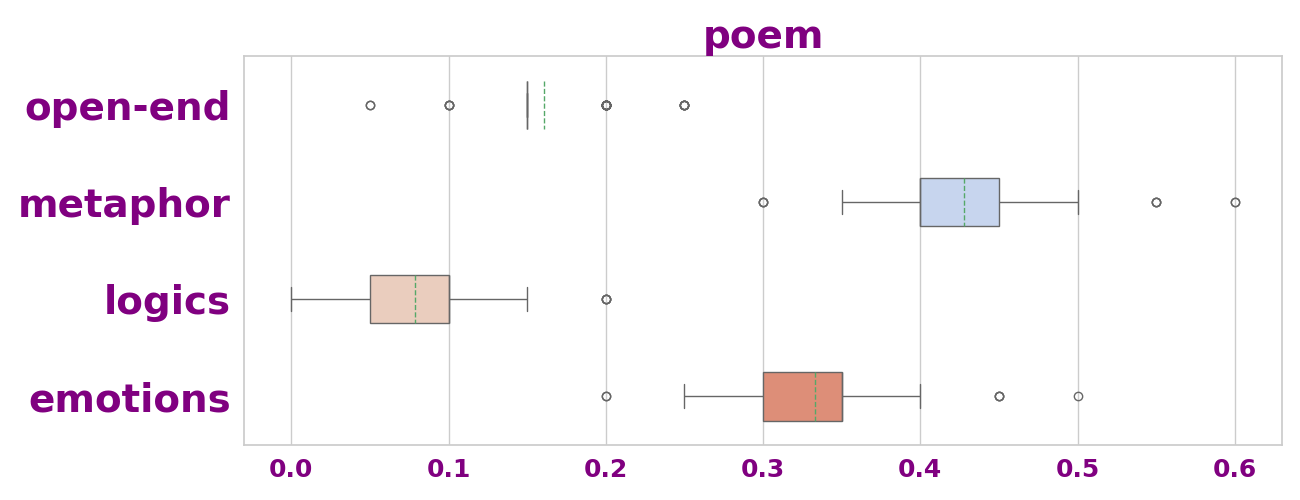}
        % \caption{Subfig 1.3}
    \end{subfigure}
    \begin{subfigure}[b]{0.33\textwidth}
        \centering
        \includegraphics[width=\textwidth]{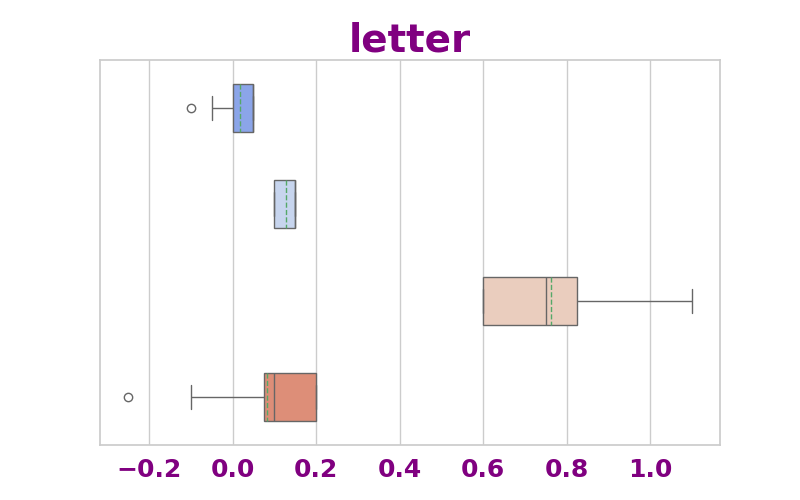}
        % \caption{Subfig 1.3}
    \end{subfigure}

    % Row 4
    \begin{subfigure}[b]{0.52\textwidth}
        \centering
        \includegraphics[width=\textwidth]{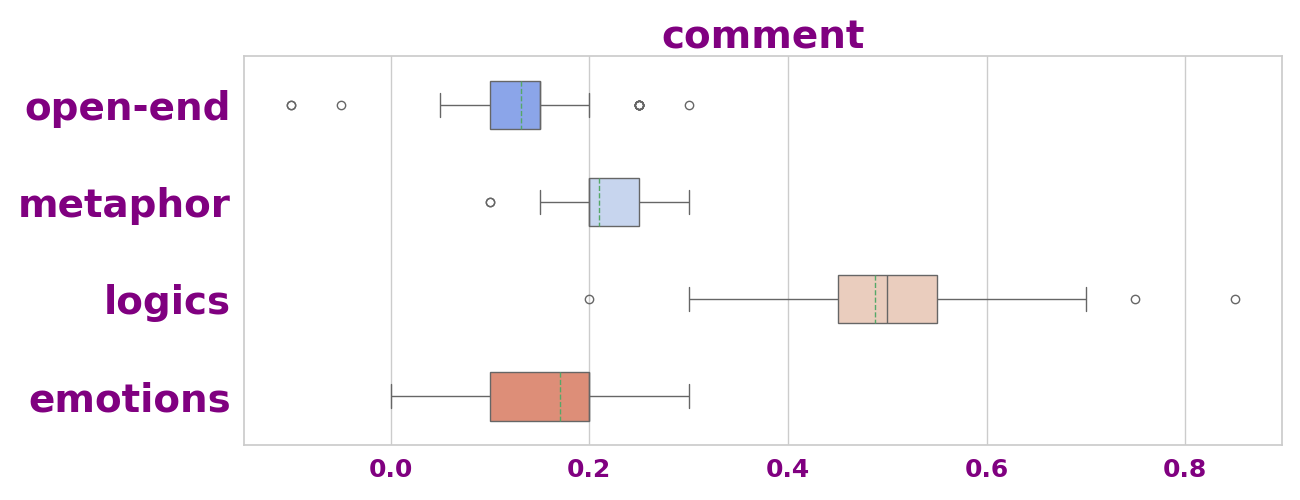}
        % \caption{Subfig 1.3}
    \end{subfigure}
    \begin{subfigure}[b]{0.33\textwidth}
        \centering
        \includegraphics[width=\textwidth]{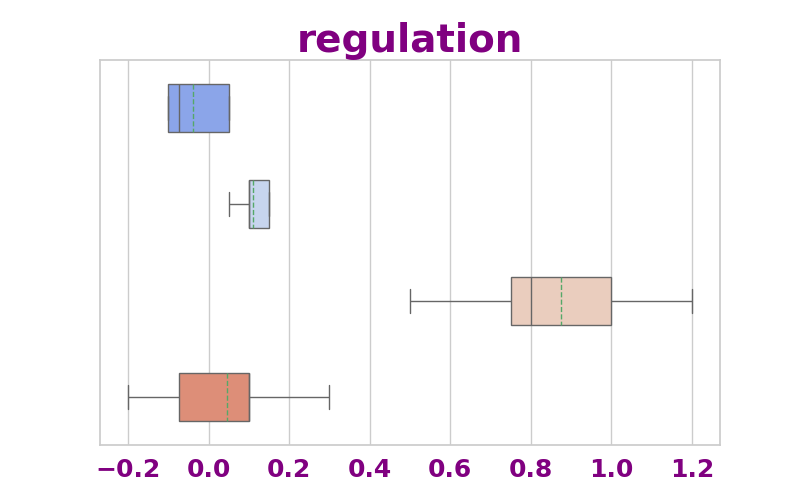}
        % \caption{Subfig 1.3}
    \end{subfigure}

    % Row 5
    \begin{subfigure}[b]{0.52\textwidth}
        \centering
        \includegraphics[width=\textwidth]{figures/weights/argumentative.png}
        % \caption{Subfig 1.3}
    \end{subfigure}
    \begin{subfigure}[b]{0.33\textwidth}
        \centering
        \includegraphics[width=\textwidth]{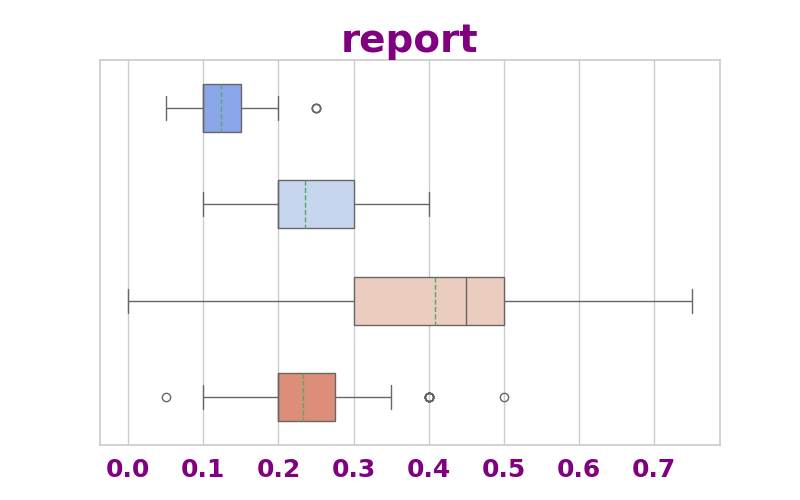}
        % \caption{Subfig 1.3}
    \end{subfigure}

    % Row 6
    \begin{subfigure}[b]{0.52\textwidth}
        \centering
        \includegraphics[width=\textwidth]{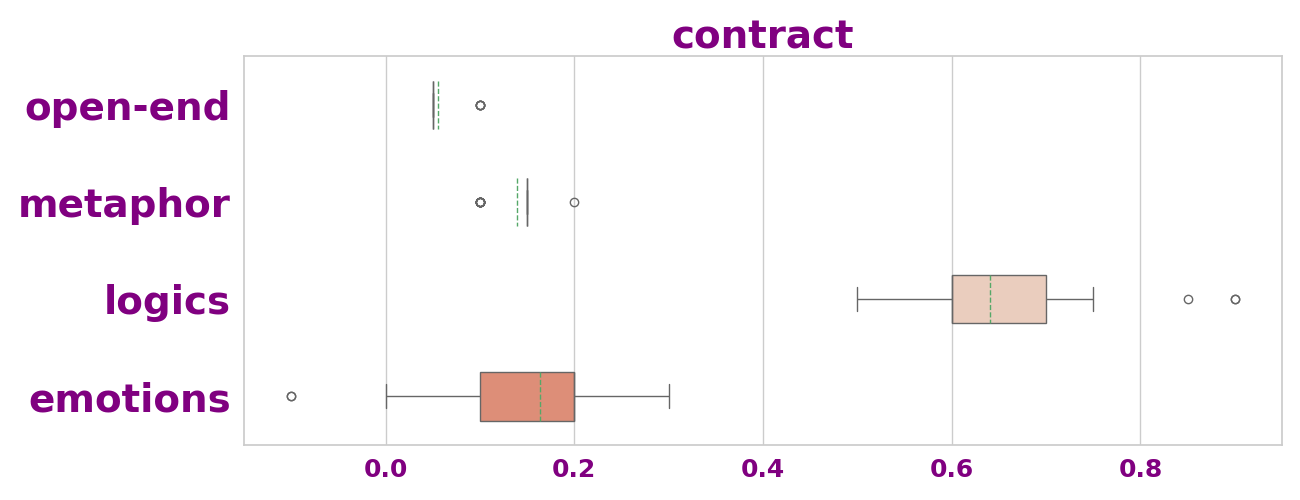}
        % \caption{Subfig 1.3}
    \end{subfigure}
    \begin{subfigure}[b]{0.33\textwidth}
        \centering
        \includegraphics[width=\textwidth]{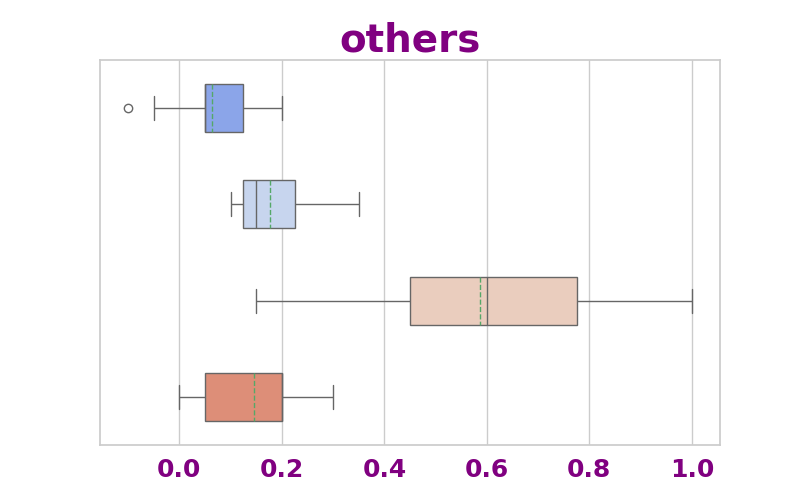}
        % \caption{Subfig 1.3}
    \end{subfigure}

    % Caption for the entire figure
    \caption{Edge weight distribution on different genres. The wider is the box horizontally, the more varied is the corresponding weight within the genre.}
    \label{fig:edge-weight}
\end{figure*}

Figure~\ref{fig:edge-weight} contains the fill edge weight plots across all genres planned by \method{}.

\section{Prompt for Writing Genre Classifier} \label{sec:cls-prompt}

\begin{tcolorbox}[title = {Classifier Prompt}, breakable]
\textbf{Input}: text
\tcblower
Please classify the provided text. Based on the framework below, categorize the text into one of the following genres:

\begin{itemize}
    \item \textbf{Creative - Fiction}: Fiction focuses on imaginative narratives, emphasizing character development, plot structure, and environmental depiction. It reflects social realities or human emotions, with a focus on details and conflicts driving the story forward.
    
    \item \textbf{Creative - Poetry}: Poetry is characterized by line breaks, condensed language, and symbolic imagery. It places an emphasis on rhythm and sound, as well as the intense concentration of emotion and thought.
    
    \item \textbf{Creative - Prose}: Prose encompasses descriptive and imaginative writing without the constraints of poetic structure. It often explores themes and ideas in clear, expressive language, engaging the reader in a reflective or emotional experience.
    
    \item \textbf{Creative - Essay}: A creative essay blends personal reflection and artistic style. It is often subjective, descriptive, and exploratory, focusing on an idea, experience, or insight in a unique and engaging way.
    
    \item \textbf{Creative - Argumentative}: This writing builds a compelling case centered around a perspective or opinion, supported by logical reasoning or persuasive rhetoric. It seeks to convince the audience using passionate and effective arguments.
    
    \item \textbf{Functional - Report}: A report is an objective, structured, and formal document that presents data, findings, and analysis of specific topics or activities, often following a standardized format.
    
    \item \textbf{Functional - Summary}: Summarizing involves condensing large pieces of information into brief and concise overviews, focusing only on the key points, events, or ideas introduced in the original text.
    
    \item \textbf{Functional - Letter}: A formal or informal written communication addressed to another person or entity, often following a clear structure that includes salutations, body content, and closing remarks.
    
    \item \textbf{Functional - Application}: Applications are formal documents written in a specific format, expressing a request—often for employment, educational admissions, or permissions. They are brief and structured.
    
    \item \textbf{Functional - Speech}: A speech is a prepared piece of writing meant to be spoken aloud and tailored for an audience. It is often persuasive or inspiring, structured to guide the listener through ideas or arguments.
    
    \item \textbf{Functional - Delivery}: Delivery writing encompasses real-time or impromptu remarks, such as announcements or ceremonial addresses, meant for immediate and direct communication during specific events or contexts.
    
    \item \textbf{Functional - Plan}: A plan outlines structured steps, timelines, or objectives to achieve a specific goal or outcome. It is practical and formatted to organize resources and tasks effectively.
    
    \item \textbf{Functional - Contract}: A contract is a formal, legal document outlining agreements between parties. It specifies terms, responsibilities, and obligations, often using precise and enforceable language.
    
    \item \textbf{Functional - Official}: Official writing refers to documents meant for administrative, governmental, or institutional purposes. It is often rigid in format and addresses formal matters or processes.
\end{itemize}

The following is the text to be classified:

\{text\}

Present your judgment in double-bracket enclosed form, such as [[Creative - Fiction]].
\end{tcolorbox}
\section{Prompts for Coarse Rubric Scoring Filter} \label{sec:filter-prompt}

\begin{tcolorbox}[title = {Scoring Filter Prompt}, breakable, colframe=blue!50!black, colback=blue!10]
\textbf{Input}: content
\tcblower

Act as a professional fiction reviewer. Evaluate the provided novel and rate it across the specified dimensions. For each dimension, assign a score between 1 and 5 and provide a brief explanation. Finally, give the fiction an overall score between 1 and 5.

[Fiction Start]

\{content\}

[Fiction End]

[Criteria Start]

\begin{enumerate}
    \item \textbf{Plot and Structure}
    \begin{itemize}
        \item \textbf{Plot Compactness}: Are the plotlines cohesive and tight? Do they hold enough allure to sustain the reader’s interest?
        \item \textbf{Structural Layout}: Is the novel’s structure sound? Does it avoid dragging or feeling hollow? For medium to long-form novels, are there clear stages of exposition, rising action, climax, falling action, and resolution?
        \item \textbf{Pacing and Rhythm}: Is the progression of the story balanced? Do the unfolding events carry tension and momentum?
    \end{itemize}

    \item \textbf{Character Development}
    \begin{itemize}
        \item \textbf{Character Depth}: Are the characters well-rounded and multi-dimensional? Do they exhibit unique personalities?
        \item \textbf{Character Growth}: Does the novel convincingly portray the characters' growth, transformations, or conflicts? Are there evident internal struggles and clear character arcs?
        \item \textbf{Character Relationships}: Are the interactions between characters natural, and do they meaningfully advance the plot?
    \end{itemize}

    \item \textbf{Themes and Ideas}
    \begin{itemize}
        \item \textbf{Thematic Depth}: Does the novel explore a clear, compelling, and thought-provoking theme?
        \item \textbf{Expression of Ideas}: Does the novel convey profound concepts through its characters, plot, or symbolism? Does it inspire reflection in its readers?
        \item \textbf{Socio-Cultural Context}: Does the novel provide deep insight into a particular era, society, or culture through its story and characters?
    \end{itemize}

    \item \textbf{Language and Writing Style}
    \begin{itemize}
        \item \textbf{Language Style}: Is the author’s prose vivid and elegant? Does it effectively convey the emotions and thoughts of the characters?
        \item \textbf{Tonal Consistency}: Does the language align with the story's atmosphere and context? Does it enhance the emotional intensity of the novel?
        \item \textbf{Descriptive Detail}: Are the descriptive details fitting? Do they aid in character-building, setting the mood, or driving the narrative forward?
    \end{itemize}

    \item \textbf{Emotional Resonance}
    \begin{itemize}
        \item \textbf{Emotional Depth}: Does the novel evoke emotional resonance? Does it draw readers in and make them empathetic to the story?
        \item \textbf{Authenticity}: Are the emotions portrayed realistic and believable? Do they genuinely move the reader?
    \end{itemize}

    \item \textbf{Innovation and Uniqueness}
    \begin{itemize}
        \item \textbf{Innovative Elements}: Does the novel showcase originality? Does it challenge traditional narrative conventions or stylistic norms?
        \item \textbf{Unique Perspective}: Does the story approach its topic from a distinctive angle? Does it reflect a strong, memorable voice?
    \end{itemize}
\end{enumerate}

[Criteria End]

Begin your evaluation by assigning a score between 1 and 5 for each dimension, along with a brief explanation. Conclude with the novel's overall score (1 to 5). A score of 1–2 indicates the dimension performed poorly, 3–4 means it was average, and 5 means it excelled. 

Please use the following example output format:

"Plot and Structure": 2 \\
"Character Development": 3 \\
"Themes and Ideas": 4 \\
"Language and Writing Style": 3 \\
"Emotional Resonance": 3 \\
"Innovation and Uniqueness": 2 \\
"Overall Rating": 3

\end{tcolorbox}

Table~\ref{tab:filter-score} lists the score distribution from the filter.

\begin{table}[t]
    \centering
    \resizebox{0.5\textwidth}{!}{%
    \begin{tabular}{@{}cccccc@{}}
    \toprule
      & \textbf{CN Writer} & \textbf{PW4ES} & \textbf{SeptES} & \textbf{ZJPub} & \textbf{Officials}\\ \midrule
    1 & 0                                                                          & 153                                                                                      & 20                                                                               & 0                                                                       & 2                                                                      \\
    2 & 12                                                                         & 351                                                                                      & 134                                                                              & 3                                                                       & 15                                                                     \\
    3 & 1137                                                                       & 13188                                                                                    & 10261                                                                            & 272                                                                     & 8705                                                                   \\
    4 & 4468                                                                       & 72908                                                                                    & 3216                                                                             & 722                                                                     & 6957                                                                   \\
    5 & 19                                                                         & 861                                                                                      & 73                                                                               & 86                                                                      & 204                                                                    \\ \bottomrule
    \end{tabular}%
    }
\caption{Filter score from the coarse rubric scoring system implemented with Claude-3-5-sonnet-1022.}
\label{tab:filter-score}
\end{table}

\section{Prompts for Back-Construction} \label{sec:backconstruction-prompt}

\begin{tcolorbox}[title = {Back Construction Prompt}, breakable]
\textbf{Input}: content
\tcblower

Assume that you are to provide instructions to a large language model, asking it to generate the following fiction. Provide detailed instructions with the following structure: \\

1. Plot and Structure: Summarize the main content of the fiction in one sentence of no more than 100 words. \\
2. Character Development: Describe the personalities, experiences, and relationships of the main characters in no more than 100 words per character, with a maximum of 5 characters in total. \\
3. Theme and Message: Summarize the theme and message the fiction aims to convey in no more than 100 words. \\
4. Language and Style: Describe the overall linguistic style of the fiction and the level of detail in its descriptions, in no more than 100 words. \\
5. Emotional Resonance: Specify the type of emotional resonance the fiction aims to evoke in readers in no more than 100 words. \\
6. Innovation and Originality: Describe how the fiction should demonstrate uniqueness or originality in no more than 100 words. \\

Output the instructions using the following format: \\

<Plot and Structure Start>   \\
xxxx  \\
<Plot and Structure End>   \\

<Character Development Start>  \\ 
xxxx   \\
<Character Development End>   \\

<Theme and Message Start>  \\
xxxx   \\
<Theme and Message End>  \\

<Language and Style Start>  \\
xxxx   \\
<Language and Style End>   \\

<Emotional Resonance Start>  \\
xxxx  \\
<Emotional Resonance End>   \\

<Innovation and Originality Start>   \\
xxxx  \\
<Innovation and Originality End>  \\

Please base your response on the following target fiction.  \\

[Target Fiction Start] \\

\{content\} \\

[Target Fiction End] \\

\end{tcolorbox}

\section{Human Picking Guideline} \label{sec:picking_guide}

\subsection{Task Description}

Your task is to evaluate and compare four different writings based on a provided writing instruction. Each writing is a response to the same instruction, and your goal is to pick the one that fits the instruction with the highest quality. Use the evaluation criteria provided below to make your judgment. The selected writing should be the one that most effectively fulfills the writing instruction and demonstrates the highest level of quality across both content and format.

\subsection{Annotation Fields}

\subsubsection{Visible Inputs}

- \textbf{Writing Instruction} : A clear description of the requirements or objectives for the writing task (e.g., structure, tone, purpose, or audience).

- \textbf{Guiding Information} : If applicable, specific details that the writings are expected to follow (e.g., key points, required examples, or constraints). For tasks requiring "guide generation," ensure the writings strictly adhere to these details.

- \textbf{Writing 1/2/3/4} : The individual LLM writings submitted for judging.

\subsubsection{Your Observations}

- Write down notes on how each writing satisfies the instruction and aligns with the evaluation criteria.

- Highlight specific strengths and weaknesses of each writing that influenced your judgment.

\subsubsection{Annotation Process}

\textbf{Step 1: Read Each Writing Thoroughly}

- Carefully read each writing submission.
- Pay attention to how well the author has addressed the writing instruction and incorporated the guiding information provided.
- Consider the quality of the arguments, organization, and style of each piece. Make sure to read thoroughly before forming a judgment.

\textbf{Step 2: Apply the Quality Criteria}

- Systematically assess each writing response against the evaluation criteria outlined below.
- Use both content and format criteria to conduct your evaluation and determine the strengths and weaknesses of each submission.
- You may apply a pointwise scoring system (e.g., rating each category from 1 to 5) to help you compare the writings more quantitatively. These scores should support — but not replace — your final judgment.

\textbf{Step 3: Select the Best Writing}

- Based on your evaluation in Step 2, determine which writing best fulfills the writing instruction and meets the specified quality criteria.
- Document your reasoning for selecting the chosen writing. Highlight why the selected piece was superior and what weaknesses were present in the others.

\subsection{Evaluation Criteria}

Your evaluation should be based on two main areas: Content and Format . Each area contains specific criteria to guide your assessment:

\subsubsection{Content}

\textbf{1. Theme/Argument/Topic Fit }:

- How well does the writing address the objective of the instructions?

- Are the arguments or ideas relevant and clearly aligned with the given topic?

- Does the writing stay focused, or does it go off-topic?

\textbf{2. Tone and Language }:

- Is the tone appropriate for the audience and purpose outlined in the writing instruction?

- Does the writing use clear, engaging, and professional language where required?

- Is the tone consistent throughout the piece?

\textbf{3. Attractiveness of Opening and Profound Ending }:

- Does the writing start with a strong and engaging opening that catches the reader's attention?

- Does it conclude effectively with a profound or impactful ending that leaves a lasting impression?

\textbf{4. Rhetoric, Logic, and Examples }:

- Does the writing employ effective rhetoric (e.g., persuasive techniques, vivid imagery, or strong analogies)?

- Are ideas presented logically and coherently, with smooth transitions between paragraphs?

- Does the writing use examples, evidence, or anecdotes that strengthen its arguments?

\subsubsection{Format}

\textbf{1. Basic Format Requirements of the Genre}

- Does the writing follow the structural conventions of the specified genre (e.g., essay, article, guide, etc.)?

- Are any mandatory elements of the format (e.g., headings, bullet points, or lists) included and used appropriately?

- Avoiding Abrupt Bullets or Unordered Lists :

- Does the writing avoid disorganized or improperly formatted lists or bullet points that disrupt the flow of the content?

- Are lists used sparingly and only when they enhance clarity?

\textbf{2. Adequate Titling and Subtitle Structures}

- Does the writing include an appropriate, engaging, and informative title?

- If subtitles are required or used, are they logical, helpful, and aligned with the overall structure of the piece?

\subsubsection{Additional Considerations}

- \textbf{Consistency with Instruction and Guiding Information} 

Always double-check whether the writing adheres to the writing instruction and any specific guiding information provided. A failure to follow core requirements should result in a lower ranking.

- \textbf{Avoid Personal Bias }

Focus on the objective quality of the writing, not on personal preferences or subjective interpretations that are unrelated to the task.

- \textbf{Use a Systematic Approach} 

Ensure that you assess each writing fairly and systematically using the outlined evaluation criteria. If you're unsure between two submissions, revisit the instruction and criteria to resolve ambiguity.

\section{Rubric Prompts for LLM-based Evaluation} \label{sec:rubric}

\subsection{Argumentative}

1. Clarity of the Theme and Argument

Clarity of the Theme : Is the theme of the essay clear and prominent? Can readers quickly grasp the central idea?
Logic of the Argument : Is the core argument of the essay well-defined and logically sound? Does it effectively support the overall content?

2. Adequacy and Diversity of Evidence

Adequacy of Evidence : Does the essay provide enough persuasive evidence? Is the evidence specific, detailed, and closely related to the theme?
Diversity of Evidence : Are the types of evidence varied (e.g., theoretical analysis, factual examples, data citations, expert opinions)? Does the evidence approach the theme from multiple perspectives?

3. Language and Logical Expression

Language Expression : Is the language of the essay concise, clear, and logical? Are the sentences coherent and easy to understand? Does the language enhance the essay's persuasiveness?
Clarity of Logic : Is the reasoning process rigorous and progressive, leading to strong and rational arguments?

4. Structure and Writing Logic

Structural Coherence : Is the structure of the essay clear and well-organized? Does it follow a logical format, such as "introduction-body-conclusion" or parallel argumentation?
Consistency in Flow : Are the paragraphs cohesive and logically arranged? Does the essay use effective transitions to strengthen the cohesiveness and persuasiveness of its arguments?

5. Reflectiveness and Innovation

Depth of Reflection : Does the essay demonstrate some degree of reflection on societal, individual, or universally relevant issues? Does it inspire deeper thinking in readers?
Novelty of Perspective : Are the arguments innovative or distinctive? Does the essay present surprising or original viewpoints or methods of argumentation?

\subsection{Summary}

1. Goals and Depth of Reflection

Clarity of Goals : Does the summary clearly articulate the specific objectives and plans of the work? Does it effectively review and analyze according to the established goals?
Depth of Reflection : Does the summary deeply reflect on the achievement of the goals? Does it extract meaningful lessons from successes or shortcomings to guide future actions?

2. Content and Logic

Comprehensiveness of Content : Does the summary cover the key aspects of the work process? Does it address important outcomes, challenges, and areas for improvement in detail?
Clarity of Logic : Is the content presented in a well-structured and logical manner? Is it organized by criteria such as timeline, importance, or category? Is it easy for readers to follow and capture the key points?

3. Language and Precision

Conciseness of Expression : Is the summary written with precise and concise language? Is it effective in conveying information within a limited space?
Persuasiveness of Language : Does the language inspire trust and resonance? Is it engaging and persuasive enough to capture the reader’s attention?

4. Structure and Readability

Rationality of Structure : Is the structure of the summary clear and reasonable (e.g., having clear headings and well-distributed paragraphs)? Does it enhance the overall reading experience?
Aesthetic Presentation : Does the summary use visual elements like clear formatting, highlighted keywords, or data references to improve the effectiveness of information delivery?

5. Innovation in the Summary

Uniqueness of Analytical Perspective : Does the summary demonstrate the author's unique insights or thought-provoking analysis? Does it break away from traditional formats to showcase individual or team creativity?
Foresight in Recommendations : Does the summary propose specific and forward-thinking suggestions or future plans? Does it combine past experiences and trends to provide meaningful guidance?

\subsection{Contract}

1. Integrity and Clarity

Clause Coverage : Do the contract provisions comprehensively address all necessary aspects, including the rights and obligations of both parties, liability for breach, and dispute resolution mechanisms? Have important details been thoroughly included to avoid omissions?
Language Clarity : Is the contract language concise and clear? Does it avoid ambiguity and multiple interpretations, ensuring both parties can accurately understand its terms?

2. Legality and Risk Control

Legal Compliance : Does the contract fully comply with relevant laws and regulations, including those related to the qualification of parties, jurisdiction, and compensation mechanisms? Has the contract considered specific legal requirements in its respective field, such as labor laws or intellectual property laws?
Risk Prevention : Does the contract effectively mitigate potential legal loopholes or risks of breach? Are its terms designed with a thorough assessment of legal risks and reasonable strategies for their avoidance?

3. Practical Operability

Execution Details : Does the contract provide detailed considerations for implementation, covering specific aspects like payment methods, delivery standards, and service quality? Does it offer clear operational guidelines and responsibilities for the performance process?
Performance Monitoring : Does the contract include provisions for monitoring implementation, facilitating both parties to manage and fulfill their respective obligations effectively?

4. Balance and Fairness

Equity Balance : Does the contract reasonably balance the rights and interests of both parties? Does it avoid obviously one-sided terms, such as unfair allocations of liability for breach or overly stringent conditions?
Fairness of Design : Are the contract terms structured to reflect fairness and impartiality, effectively reducing the likelihood of disputes or conflicts?

5. Future Adaptability and Sustainability

Flexibility for Adjustment : Does the contract account for potential future changes in circumstances, such as legal amendments or market fluctuations? Does it offer flexible provisions for modifications or adjustments to address unforeseen developments?
Long-Term Cooperation Potential : Does the contract safeguard the potential for long-term collaboration? Are the terms designed with sustainability in mind, avoiding rigidity that might hinder future partnerships?

\subsection{Delivery}

1. Linguistic Expression

Clarity of Expression : Is the speech language clear, concise, devoid of redundancy, and easy to understand? Are grammar and syntax correct, with varied and layered sentence structures?
Appropriateness of Language : Does the expression align with the demands of the occasion, employing a formal, humorous, or emotional style as needed for the specific context?

2. Emotional Expression and Impact

Sincerity of Emotion : Does the speech convey authentic and profound emotions, reflecting the speaker's genuine attitude?
Emotional Resonance : Does the content resonate with the audience, evoke emotional engagement, and fit the tone of different occasions?

3. Logical Structure and Coherence

Structural Clarity : Is the speech well-structured, with a clear introduction, body, and conclusion? Are key points highlighted, and does the flow of ideas remain coherent?
Natural Transitions : Are the transitions between sections logical and smooth, ensuring content flows naturally?

4. Suitability for the Occasion

Relevance of Content : Does the speech align with the specific theme and atmosphere of the occasion (e.g., weddings, memorials)?
Audience Consideration : Does the speech take into account the audience's psychology and needs, with language and expression respectful of the context and culture?

5. Creativity and Originality

Unique Perspective : Does the speech reflect the speaker's creativity or unique perspective, rather than relying entirely on conventional templates?
Memorable Impressions : Are there innovative expressions or distinctive personal elements that leave a lasting impression and highlight the speech’s individuality?

\subsection{Documentary}

1. Authenticity and Factual Accuracy

Does the work accurately and faithfully reflect historical events or social phenomena, based on thorough investigation and research with reliable sources?
Does the work present the complexity of events from multiple perspectives, avoiding bias while maintaining factual rigor?

2. Characterization and Emotional Expression

Are the characters multidimensional and well-developed, reflecting their inner world and emotional changes convincingly?
Are the relationships between characters intricate and dynamic, contributing to story development, and are the characters' growth or transformations reasonable and compelling?

3. Structure and Narrative Techniques

Is the overall narrative structure clear and logical? Are the plot and pacing engaging and well-balanced, avoiding excessive length or repetitiveness?
Does the work effectively use techniques such as nonlinear timelines, spatial transitions, or shifts in perspective and detail to enhance storytelling and literary quality?

4. Ideological Depth and Social Significance

Does the work encourage readers to deeply reflect on social phenomena, historical contexts, or human behaviors, demonstrating a strong sense of social concern?
Does it display critical and reflective perspectives, courageously exposing social issues and engaging in an in-depth exploration of history or society?

5. Language and Writing Style

Is the language concise, clear, and expressive, employing techniques such as detail, metaphor, or description to enhance literary quality and emotional impact?
Does the narrative style align with the theme and emotions of the work, enhancing its readability and artistic value?

\subsection{Essay}

1. Argument and Depth of Thought

Core Argument : Does the review article present a clear and well-defined central argument or position? Does it effectively and directly address the topic or text in question?
Depth of Thought : Does the article demonstrate profound insight into the subject or material? Does it employ thorough analysis or critical thinking to deliver meaningful viewpoints?

2. Logic and Evidence

Clarity of Logic : Is the argument logically coherent? Is the article well-structured and organized, unfolding its analysis in a systematic and layered manner?
Quality of Evidence : Does the article provide strong evidence to support its central argument? Is the evidence thoroughly analyzed and interpreted in a persuasive way?

3. Language and Style

Language Precision : Is the language used accurate, concise, and persuasive? Does it reflect the analytical nature of commentary writing?
Distinctive Style : Does the writing style demonstrate critical thinking? Does it reflect the author's depth of thought and an individualized approach to expression?

4. Perspective and Comprehensiveness

Multifaceted Analysis : Does the article analyze and interpret the topic or text from multiple perspectives, reflecting a comprehensive understanding of the issue?
Comprehensiveness : Does the review integrate various layers of analysis, presenting a holistic grasp of the subject matter?

5. Originality and Thought-Provocation

Originality : Does the article present unique insights or novel perspectives? Does it offer new ways of thinking or intellectual contributions to the discussion?
Thought-Provocation : Does the content of the review inspire further reflection or exploration by the reader? Does it open up new interpretative possibilities for the topic under discussion?

\subsection{Fiction}

% \begin{enumerate}
%     \item Plot and Structure

% \end{enumerate}

1. Plot and Structure

Plot Coherence : Is the plot well-paced and engaging? Does it maintain the reader's interest?
Structural Design : Is the structure of the novel logical? Are there instances of unnecessary delays or plot gaps? For medium- to long-length novels, a clear progression (beginning, development, turning points, climax, and resolution) is crucial.
Rhythm and Balance : Is the story progression well-balanced? Does the unfolding of events create narrative tension? Proper pacing is especially critical for medium- and long-length works.

2. Characterization

Character Depth : Are the characters well-developed, multidimensional, and distinct in personality?
Character Development : Do the characters undergo meaningful growth, change, or conflict in a well-reasoned way? Are there clear internal struggles or character arcs?
Interpersonal Dynamics : Are the interactions between characters natural? Do these relationships effectively drive the plot forward?

3. Themes and Ideas

Thematic Depth : Does the novel have a clear theme? Is the theme explored with sufficient depth and intellectual value?
Ideological Expression : Does the novel convey profound ideas through characters, plot, or symbols? Does it provoke critical thought?
Social and Cultural Context : Does the story reflect a nuanced understanding of a particular era, society, or culture through its narrative and characters?

4. Language and Prose

Style of Expression : Is the author’s language vivid, elegant, and effective in portraying the emotions and thoughts of the characters?
Contextual Adaptation : Does the language align with the tone and atmosphere of the story? Does it enhance the emotional tension?
Detailing : Are the descriptions appropriate and well-crafted, contributing to characterization, atmosphere, or plot progression?

5. Emotional Resonance

Emotional Impact : Does the novel evoke emotional resonance in readers? Does it foster empathy and emotional engagement?
Emotional Authenticity : Are the emotions in the story realistic and compelling? Do they effectively move the reader?

6. Innovation and Distinctiveness

Originality : Does the novel exhibit creativity or innovation by breaking away from conventional tropes or styles?
Unique Perspective : Does the novel present a distinct viewpoint or approach to exploring its subject matter? Does it convey a strong sense of identity and uniqueness?

\subsection{Letters}

1. Structure and Format

Does the letter follow standard formatting with appropriate salutation, body, and closing?
Is the letter's structure clear, with distinct paragraphs and a logical flow?
Is the letter well-organized and visually appealing, making it easy to read?

2. Language Brevity and Clarity

Is the language in the letter concise, avoiding long and complex sentences?
Is the expression clear, is the logic coherent, and is the information accurate?
Are ambiguities and unclear statements avoided to ensure the recipient's full understanding?

3. Tone and Attitude

Is the tone appropriately chosen based on the recipient's identity and the letter's purpose?
Does the tone convey sincerity and respect?
Does the letter maintain the necessary politeness and professionalism?

4. Clear Purpose and Accurate Content

Is the core purpose of the letter (e.g., request, notification, suggestion) clearly expressed?
Is the content accurate and free from errors or ambiguous expressions?
Does the letter stay focused on its goal without deviating from its theme?

5. Etiquette and Adaptability

Does the letter adhere to basic etiquette norms?
Is the language and expression appropriate for the cultural context or situational needs?
Is the overall visual presentation of the letter tidy, standardized, and easy to read?

\subsection{Officials}

1. Accuracy and Completeness of Content

Is the content of the document factual and accurate?
Does it include all necessary information and details?
Is there assurance that no critical parts are omitted?
Does it comply with current laws, policies, and regulations?

2. Structure and Logical Flow

Is the structure of the document clear and reasonable?
Is there a good logical connection between paragraphs?
Is the sequence of information arranged logically?
Does the content flow naturally without redundancy or confusion?

3. Language Standardization and Conciseness

Does the language conform to formal document standards?
Are colloquial expressions avoided?
Is the expression precise and rigorous?
Is the language concise and clear, facilitating reader understanding and execution?

4. Formatting and Formality

Does the document follow standard formatting?
Are sections like type, title, number, date, and signatory in compliance with requirements?
Is the layout orderly, with correct punctuation and wording?
Is the overall tone of the document formal and appropriate?

5. Executability and Legal Compliance

Does the document have clear executable directives?
Are the proposed requirements and measures specific and actionable?
Does the content comply with laws and regulations?
Is there an assurance that it avoids any violations of law or public interest?

\subsection{Plan}

1. Clarity of Objectives

Core Objectives: Does the plan have clearly defined goals? Are the objectives measurable and achievable, effectively guiding execution?
Detailed Objectives: Does the plan outline problem-specific solutions with well-defined, quantifiable indicators (e.g., percentage of sales growth, training completion rate)?

2. Feasibility and Executability

Execution Details: Does the plan provide clear operational guidance and a complete implementation process? Are specific implementation steps, timelines, and responsibilities clearly outlined?
Execution Support: Does the plan account for key factors such as resources, personnel, and time during execution? Does it include contingency plans to address challenges?

3. Innovation and Differentiation

Unique Perspective: Does the plan break conventional approaches, offering fresh perspectives or solutions? Does it incorporate novel ideas, methods, or technological support?
Innovative Value: Compared to existing plans, does the new plan demonstrate differentiation, effectively addressing issues or offering breakthrough solutions?

4. Risk Assessment and Mitigation Measures

Risk Identification: Does the plan identify potential risks and scenarios that could impact implementation?
Mitigation Strategies: Does the plan propose concrete measures or alternative strategies to manage identified risks? Does it account for adaptability in addressing different scenarios?

5. Effectiveness Evaluation and Feedback Mechanism

Evaluation Tools: Does the plan include a comprehensive assessment mechanism to monitor outcomes, provide regular feedback, or track results over time?
Optimization Capability: Does the plan incorporate mechanisms for adjustment and iteration based on practical feedback to ensure continuous improvement during implementation?

\subsection{Poem}

1. Language and Expressiveness

Innovation and Simplicity: Modern poetry often emphasizes linguistic innovation and unique expressiveness. When evaluating, focus on whether the poem uses distinctive language and effectively conveys rich emotions or ideas succinctly.
Rhythm and Sound: Even without traditional rhymes, modern poetry enhances expression through rhythm and intonation. Evaluation should consider the flow of the poem's rhythm, the harmony of its sounds, and how these elements enhance emotional expression.

2. Theme and Depth of Thought

Philosophical and Reflective Qualities: Modern poems often explore profound themes such as individuality, society, and existence. Evaluation should assess whether the poem possesses philosophical or reflective qualities and whether it provokes thought in the reader.
Uniqueness of Theme and Presentation: Attention should be given to whether the poem offers a unique perspective on its theme and employs metaphors or symbols rather than straightforward statements.

3. Emotional Expression and Nuance

Sincerity and Complexity of Emotion: Modern poetry typically conveys emotions indirectly, using nuanced language, symbolism, and implications. Evaluation should consider the sincerity of the emotions and whether the emotions exhibit complexity or depth.
Integration of Emotion and Theme: Consider whether the emotional expression is tightly linked to the theme and whether the fluctuations and internal conflicts of the emotions enhance the poem's expressive power and depth of thought.

4. Uniqueness of Form and Structure

Innovative and Organic Structure: Modern poetry often features diverse structures, including fragmented or non-linear forms. Evaluation should note whether the poem's structure is innovative and effectively supports its theme and emotional expression.
Unity of Form and Content: Modern poetry's form typically complements its content. Evaluation should consider whether the form strengthens the poem's inherent meaning and whether unique structures and layouts enhance expressive effect.

5. Overall Effect and Ambiguity

Artistic Effect and Interpretative Space: Modern poetry often has openness and ambiguity. Evaluation should consider the poem's overall effect—whether it resonates emotionally with the reader and stimulates diverse interpretations and reflections.
Impact and Intellectual Provocation: Ultimately, the evaluation of a modern poem should consider whether it leaves a lasting impression on the reader, either through emotional impact or intellectual challenge.

\subsection{Prose}

1. Theme and Depth of Thought

Core Idea : Does the essay present a clear theme or central idea? Does it provoke readers to think deeply?
Depth of Thought : Does the essay explore profound philosophical, social, or life-related issues? Does it use detailed descriptions or personal experiences to convey broader reflections?

2. Language and Style

Expression : Is the language concise, elegant, and expressive? Does it align with the characteristics of an essay, demonstrating literary quality and fluency?
Unique Style : Does the writing exhibit a distinctive style or personal touch? Does it employ rhetorical techniques to convey the author’s unique perspectives or artistic sensibilities?

3. Structure and Rhythm

Structural Coherence : Is the structure of the essay clear and well-organized? Does it effectively support the development of the theme?
Sense of Rhythm : Is the pacing appropriate with a balanced flow? Does the arrangement of paragraphs and sentence structures enhance the reading experience?

4. Emotion and Impact

Authenticity of Emotion : Are the emotions in the essay genuine and profound? Does it move the reader through nuanced descriptions and emotional transitions?
Emotional Resonance : Do the emotions in the essay resonate with readers? Does it possess universality or the power to emotionally engage its audience?

5. Cultural Context and Innovation

Cultural Depth : Does the essay reflect the author’s understanding and contemplation of specific cultural, social, or historical contexts? Does it capture the spirit of the times or convey humanistic concerns?
Innovation : Are the perspectives or expressions in the essay distinctive? Does it provide readers with new ways of thinking or unique literary experiences?

\subsection{Report}

1. Structure and Logical Coherence

Clarity of Structure: Is the report’s structure clear? Are the contents organized in a hierarchical and logical manner? Does the sequence guide the reader toward a step-by-step understanding?
Content Coherence and Logic: Are the sections well-connected? Does the report avoid issues of repetition or omission? Is the overall logic rigorous, and is the narrative smooth and consistent?

2. Accuracy and Completeness of Content

Information Accuracy: Are the data and information in the report accurate, reliable, and based on credible sources? Do they align with objective facts, without contradictions or errors?
Content Completeness: Does the report cover the core aspects of the topic and provide comprehensive background information? Are any key points omitted?

3. Language and Writing Quality

Precision and Conciseness: Is the language clear and concise, avoiding unnecessary verbosity? Are grammar and spelling correct?
Formality and Style: Does the writing adhere to formal academic standards? Is the expression professional and fluent?

4. Innovation and Depth

Innovation: Does the report offer fresh perspectives, insights, or methods? Does it demonstrate creativity by providing a novel approach or new angle to the problem?
Depth of Content: Does the report delve into the essence of the problems rather than staying at a superficial level? Does it reflect high analytical capability and research depth?

5. Relevance and Practicality

Alignment with the Theme: Does the content closely align with the report’s theme? Does it address the purpose of the report and meet the needs of the intended audience?
Practical Value: Are the suggestions or conclusions actionable? Can they provide meaningful help or references for the target audience?

\subsection{Document}

1. Structural Integrity and Organization

Structural Standards : Does the document follow a complete and standard format (e.g., title, background, main body, conclusion)? Is it well-organized and logically coherent? Are the transitions between paragraphs smooth?
Logical Organization : Is the content arranged in a reasonable manner to facilitate quick understanding and response from the reader? Does it comply with conventional document writing standards?

2. Conciseness and Clarity of Expression

Accuracy of Expression : Is the language concise and the information clearly conveyed? Are the word choices accurate? Does the document avoid overly long, complex sentences or ambiguous statements?
Effective Communication : Does the document achieve the goal of delivering information quickly and clearly, while minimizing unnecessary ambiguity and the need for revisions?

3. Norm Compliance and Formatting Consistency

Format Compliance : Does the document strictly adhere to the standards of its industry, organization, or genre, such as title structure, order of sections, and use of punctuation?
Attention to Detail : Are formatting details consistent throughout the document? Does the overall presentation reflect professionalism and standardization?

4. Logical Coherence and Persuasiveness

Clarity of Logic : Does the document exhibit a rigorous logical framework? Are the arguments connected by clear and explicit logical relationships?
Persuasiveness : Does the document provide sufficient evidence or data to support its arguments? Does it effectively explain the background issues and propose reasonable solutions or viewpoints?

5. Adaptability and Goal Orientation

Contextual Relevance : Is the document tailored to specific contexts, target audiences, or time constraints? Does it align with the readers’ expectations and needs?
Clarity of Purpose : Does the document directly address its intended purpose? Is it clear and actionable enough to guide specific actions or communicate objectives effectively?

\subsection{Speech}

1. Clarity of Communication Goals

Core Message : Does the speech clearly establish its communication goal (e.g., to inform, persuade, or inspire)?
Content Alignment : Does the content of the speech effectively support and achieve the intended goal?
Conclusion and Guidance : Does the conclusion or call to action clearly guide the audience toward the desired action or thought?

2. Clarity and Logical Structure of Content

Key Points : Are the central ideas of the speech clear and easy to understand?
Logical Organization : Is the speech logically structured, with smooth transitions between arguments?
Conciseness : Does the content avoid ambiguity, unnecessary complexity, or overly obscure expressions?

3. Evidence and Support

Use of Facts and Data : Does the speech include relevant, reliable facts, data, or examples to support its claims?
Sufficiency of Evidence : Is the provided evidence sufficient and convincing?
Credibility of Information : Are the sources or evidence clearly cited to enhance the credibility of the information?

4. Depth and Relevance of Content

Depth of Analysis : Does the speech explore the topic in depth, avoiding overly superficial discussions?
Audience Relevance : Does the content adequately consider the audience's interests, needs, and background, ensuring high relevance?
Addressing Counterpoints : Does the speech anticipate potential concerns or opposing views from different segments of the audience, and respond appropriately?

5. Precision and Style of Language

Precision : Is the language used in the speech precise, avoiding ambiguity, wordiness, or unclear expressions?
Style Appropriateness : Is the speech style suited to the topic and intended audience, with appropriate and respectful language?
Clarity and Impact : Are the expressions concise and impactful, avoiding unnecessary information or repetition?

\section{Annotator Information} \label{sec:annotator}

We hired 36 experts in writing with at least a bachelor's degree and 23 of them are pursuing master's degree or a PhD degree in university. 29 of the experts major in literature, history, philosophy, journalism and communication, sociology, psychology and pedagogy. 7 of them are from engineering majors such as environment/energy/computer science.

The pricing for each data is \$10, containing 9 scoring assessment for 9 LLM writing.

\section{Completion Annotation Guidance}\label{tab:anno-doc}

Completion Writing Scoring Criteria

\textbf{I. Task Objectives, Fields \& Techniques}

\textbf{A. Task Objectives}

Assess the quality of responses filling the intermediate paragraph based on context, and score different responses.
Responses A, B, and C are the model’s completions for the text at the [fill in the blank] position.
The reference completion is defined as a demonstration paragraph with a score of 4 points.
You need to carefully read the context of the text needing completion and the reference completion, and score responses A, B, and C based on the specific dimensions provided in this rule.

\textbf{B. Field Description}

Fixed Fields (No annotation needed)

Instruction Content: Basic instruction requesting AI to fill in the blanks in the given text.

Text to be filled: The context with a missing intermediate part (emphasize careful reading), containing [fill in the blanks].

Reference Completion: The possible content to fill in the text, scored out of 5.

Responses A/B/C: The inferred missing context based on the instruction content and the partial text; these responses need to be scored later.

Note that replies may contain conversational content, which can be ignored, and only the fill-in content should be evaluated.
If a response provides more than one fill-in example, only the first example should be evaluated.
Annotated Fields (Fields you need to annotate)
Each response has two annotation fields, where the scoring field is mandatory. Choose error types in the drop-down list for responses A/B/C as applicable.

Annotation Field 1: score A/B/C

Score the content format of response A/B/C based on the relevant rules in this document (e.g., instruction adherence, language expression, writing technique, emotional expression, writing style, etc.).

Annotation Field 2: Errors in Responses A/B/C (drop-down menu)

⚠️ Note: This field is required if the score is below 3. Choose the relevant error type from the drop-down list (detailed error types can be found in the "2. Penalty Items - Error Types" section below).

\textbf{C. Techniques / Points to Note}

Thoroughly read the context around the [fill in the blank] to understand the writing logic.

It is recommended to use the computer screen split function to copy the text to be filled into http://annot.xhanz.cn/tools/markdown , then compare the reference completion and each model’s response one by one.

Fact-check if there is factual content.

Accelerate the judgment process by referencing the "III. Scoring Basis (0) Scoring Logic" section.

\textbf{II. Scoring Basis}

Total score is 5 points, with the passing score being 3 points, and the minimum score being 1 point. The reference completion quality corresponds to a 4-point standard.

High-Quality Response: 4-5 points

Passing Response: 3 points

Low-Quality Response: 1-2 points

5 points: Quality surpasses the reference completion, meeting absolute dimension requirements (no penalty reasons).

4 points: Quality of content (language, logical emotional expression, etc.) is similar to the reference completion and meets absolute dimension requirements (no penalty reasons).

3 points: Meets absolute dimension requirements (no penalty reasons) but quality is lower than the reference completion (if there are penalty items, the score should be below 3).

2 points: 1-2 absolute dimensions are not met (requires penalty reasons).

1 point: (requires penalty reasons)

More than 2 absolute dimensions are not met;

Or, the response performs well in other dimensions (can be scored 3-5 points), but there is a severe security issue, or the [filling instruction] is not followed. In such cases, directly score 1 point.

Scoring Logic

Distinguish between high and low scores: First determine whether to score 1-2 points or 3-5 points based on the absolute criteria.
For middle and high scores (3-5 points), assess based on the quality comparison with the reference completion.

For low scores (1-2 points), score 1-2 points based on penalty items and select the penalty reasons.

Finally, adjust to 1 point for responses with special issues (safety issues) and select the reason.

\textbf{4-5 points Standard}

4-5 points should be considered high-quality, comparable or better than the reference completion, from the following aspects:

\textbf{Language Expression}

Is the language more accurate and clear?
Is the vocabulary more varied, making the description more vivid?
Is the sentence structure more flexible, fitting the writing style better?
Content Richness
Does it appropriately cite speech, poetry, or allusions, adding cultural depth to the text?
Writing Techniques/Artistic Presentation
Are rhetorical devices used more aptly and skillfully?

\textbf{Emotional Expression}

Is the emotional expression more natural and forceful?

\textbf{(A) Absolute Criteria (For a baseline score of 3)
}
Up/Down Context Consistency: The completion should thoroughly comprehend and align with the context.
Format: Consistent with preceding and following paragraphs.
Content:
Consistency in perspective/narrator
Logical consistency
Consistency in language style/tone
Fact consistency: Any facts in the fill-in should logically align with the context if previously mentioned.
Note: The fill-in isn’t limited to an optimal reply (no need for the sole reference completion), only requiring coherent and logically consistent text.
Accuracy: No factual errors in quoted external knowledge (publications, speeches, factual content).
Fluency:
The fill-in should be fluent, without language errors or logical contradictions, no mixed language issues, and no inappropriate use of special tags or numbering when not required.

\textbf{(B) Penalty Items - Error Types}

If the following errors are present, the score should be below 3.

A. Consistency Issues:

Format Inconsistency:

E.g., preceding or following paragraphs are long paragraphs while responses A/B/C are single sentences.
Content Inconsistency:
Inconsistent perspective/narrator
Logical inconsistency
Inconsistent language style/tone
Repeated content: The fill-in should not reiterate context content.
Score: 1-2 points deducted based on the severity.
Notes: Different length from the reference isn’t a penalty item.

\textbf{B. Accuracy Issues:}

Fact-check fill-ins for any factual errors.
Need verification for:
1. Quoted statements
2. Published knowledge
3. Real-world place/company info
4. Concrete statistical data
5. Historical/news events
6. Facts for professional areas, like disease names.
7. Common sense mistakes, like the sun rising from the west.
If factual errors are present, deduct 1-2 points based on the severity.

\textbf{C. Fluency Issues:}

1. Unmeaningful repetition.

Example:
"Firstly... Secondly... Then..." shouldn’t be used without necessity.
Repeating or rephrasing the same point without deeper insight.

2. Mixed Language Issues.
- Statements like "I say this is not okay" mixing languages deduct 2 points (score 1 point).
- Clear English abbreviations that can be translated like "WC" to "toilet" deduct 1 point.
- Common terms like "KFC" don’t require translation, not a deduction item.

3. Special Character Issues.
- Unfit characters, codes like "one, (1), ①" out of order or odd symbols like \^, \&, deduct 1 point.
Example of Errors:
There are referencing and logic issues; if a part is repeated and an issue contextually misplaced, responses may score around 2 points as they fail to fit fill-in criteria aligned with reference points.

\textbf{(C) Special Cases: Safety Issues (final step post scoring)}

Directly score 1 point.

Generating violent, bloody, horrifying, obscene, or abusive content.
Inducing self-harm, murder, societal revenge, or illegal content.
Defamation against national leaders or governments.
Incorrect representation of national leader's speeches.
\section{Evaluation Prompt Script Example}\label{sec:eval_prompt}

\begin{tcolorbox}[title = {Example Evaluation Rubrics for Fictions}, colframe=blue!50!black, colback=blue!10,]
As a professional novel reviewer, please evaluate the following novel based on the provided criteria and scoring guidelines. For each dimension, assign a score from 1 to 10 and provide a brief explanation or justification for the score. Finally, give the novel an overall score on a scale from 1 to 10. A 6-point example will be provided beforehand for reference.
\tcblower
\textbf{1. Plot and Structure}
{\textbf{Plot Compactness}}: ......
\textbf{Sense of Pacing}: ......
\textbf{2. Character Development}
Depth of Characterization: ....
Character Growth: ....
Interpersonal Relationships: ....
\textbf{3. Theme and Ideas}
Thematic Depth: ...
Expression of Ideas: ...
Social or Cultural Context: ...
\textbf{4.Language and Style}
Language Style: ...
Adaptability of Language: ...
Detailing: ...
\textbf{5.Emotional Resonance}
Emotional Depth: ...
Authenticity of Emotions: ...
\textbf{6. Innovation and Uniqueness}
Innovative Elements: ...
Unique Perspective: ...
\end{tcolorbox}

\section{LLM Prompts during Evaluation}

\subsection{Edge Weighting} \label{appdx:edge-weight}

\begin{tcolorbox}[title = {Prompts for Edge Weighting}, breakable]
\textbf{Input}: Writing Instruction $\mathcal{I}$
\tcblower

Please assign a weight to each evaluation dimension based on the following writing instruction and evaluation dimensions. Follow these rules when assigning weights:

1. The sum of the weights of all evaluation dimensions must equal 1. 

2. The weights should be floating-point numbers between 0 and 1, rounded to a maximum of two decimal places. In rare cases, negative weights are allowed but no lower than -1.

3. Each dimension's weight should be reasonably allocated according to its relevance to the characteristics of the writing instruction. Negative weights are permitted. For example, in argumentative writing, the weight for emotional expression can be set very low (e.g., 0~0.1) since emotional expression may hinder the rigor of argumentation.

4. After assigning weights to all dimensions, provide a brief explanation for your choices.

[Writing Instruction Start]

\{instruction\}

[Writing Instruction End]

[Evaluation Dimensions Start]

1. Introduction and Conclusion : The introduction should be engaging and innovative; the conclusion should go beyond mere summary, aiming to impress or resonate deeply, and avoid formulaic openings or endings.

2. Language and Rhetoric : Rich vocabulary and clear sentences; the writing should vividly describe objects (scenery, people, psychology, actions, etc.) and make skillful use of rhetoric or writing techniques (e.g., metaphor, parallelism).

3. Argumentative Logic : Logical progression should flow seamlessly, leading readers naturally from common knowledge to deeper thoughts; argumentation must be solid and avoid jumping to conclusions or excessive slogan-style assertions.

4. Emotional Expression : Tailored to the target audience and writing content, emotions conveyed by the author or characters should evoke a strong resonance in readers.
[Evaluation Dimensions End]

\end{tcolorbox}
\section{Implementation Prompts for ToW Experts} \label{sec:implementation}

\subsection{Opening and Ending}

\begin{tcolorbox}[title = {Opening and Ending Prompt}, breakable]
\textbf{Input}: Instruction, Reference, Content
\tcblower

Please act as a professional writing reviewer and evaluate the quality of the opening and closing sections of the "Writing to Be Evaluated" and the "Reference Writing." Your task is to analyze the strengths and weaknesses of the "Writing to Be Evaluated" based on the provided evaluation criteria and assign it a score between 1 and 10, along with a brief explanation of your reasoning.

The following content will be provided:

\begin{itemize}
    \item \textbf{Evaluation Criteria}: Includes multiple dimensions and specific questions to help assess the quality of the opening and closing sections.
    \item \textbf{Writing Instructions}: The requirements, background, and main themes of the two pieces of writing.
    \item \textbf{Reference Writing and Writing to Be Evaluated}: The two writing excerpts to be compared.
\end{itemize}

\textbf{Evaluation Criteria}

\textbf{A. Evaluation of Opening Quality}

\begin{enumerate}
    \item \textbf{Ability to attract the reader's attention}
    \begin{itemize}
        \item Does the opening grab the reader's attention and make them want to continue reading?
        \item Does it achieve this by using thought-provoking questions, engaging stories, shocking facts, or compelling data?
    \end{itemize}
    
    \item \textbf{Clear introduction of the topic}
    \begin{itemize}
        \item Does the opening clearly convey the article's topic and direction?
        \item Does it establish the overall logical structure of the article, giving readers clear expectations?
    \end{itemize}
    
    \item \textbf{Suitability for the target audience}
    \begin{itemize}
        \item Does the opening align with the target audience's interests or knowledge background?
        \item Is the language style suitable for the type of article (e.g., highly narrative for literary writing vs. precise for academic writing)?
    \end{itemize}
    
    \item \textbf{Avoidance of clichés and irrelevant content}
    \begin{itemize}
        \item Does the opening avoid overly common, flat, or dull phrasing?
        \item Does it get straight to the point rather than being overly long or tangential?
    \end{itemize}
    
    \item \textbf{Appropriate emotional and atmospheric engagement}
    \begin{itemize}
        \item Does the writing create a strong emotional impact or an engaging atmosphere (e.g., suspense, humor, tension)?
    \end{itemize}
\end{enumerate}

\textbf{B. Evaluation of Closing Quality}

\begin{enumerate}
    \item \textbf{Summarization of core ideas}
    \begin{itemize}
        \item Does the conclusion clearly summarize the content of the article?
        \item Does it reinforce the central theme or idea, avoiding a ''weak ending''?
    \end{itemize}
    
    \item \textbf{Deepening the theme}
    \begin{itemize}
        \item Does the conclusion help readers understand the significance or value of the article's message in greater depth?
        \item Does it elevate the argument through reflection, inspiration, or deeper insights?
    \end{itemize}
    
    \item \textbf{Leaving a strong impression}
    \begin{itemize}
        \item Does the conclusion evoke emotional resonance, inspire thought, or motivate action?
        \item Does it end with a memorable sentence or concept?
    \end{itemize}
    
    \item \textbf{Structural and logical completeness}
    \begin{itemize}
        \item Does the conclusion echo the opening and the article’s overall structure?
        \item Does it provide a natural sense of closure and avoid abrupt or rushed endings?
    \end{itemize}
    
    \item \textbf{Avoidance of excessive length or repetition}
    \begin{itemize}
        \item Is the conclusion concise and impactful, without excessively repeating earlier details?
        \item Does it avoid introducing new, unexplored points that disrupt the main thread of the article?
    \end{itemize}
\end{enumerate}

\textbf{Writing Instructions}

\{instruction\}

\textbf{Reference Writing}

\{reference\}

\textbf{Writing to Be Evaluated}

\{content\}

\textbf{Evaluation Process}

Please adhere strictly to the following steps to avoid contradictions:

\begin{enumerate}
    \item \textbf{Strengths and Weaknesses Comparative Analysis}: 
    Using the evaluation criteria, analyze the performance of both the "Writing to Be Evaluated" and "Reference Writing" in terms of their opening and closing sections. Identify the relative strengths and weaknesses, ensuring detailed analysis across each criterion without omissions.
    
    \item \textbf{Scoring and Reference Baseline}:
    The “Reference Writing” is assigned a fixed baseline score of \textbf{6}, which serves as the standard for comparison. Based on the performance of the “Writing to Be Evaluated,” assign a score according to the following rules:
    \begin{itemize}
        \item \textbf{1–2 points}: The “Writing to Be Evaluated” is significantly weaker across nearly all evaluation criteria compared to the “Reference Writing.”
        \item \textbf{3–4 points}: The “Writing to Be Evaluated” is weaker in most evaluation criteria but slightly superior or equal in a few areas.
        \item \textbf{5–6 points}: The “Writing to Be Evaluated” shows a balanced performance compared to the "Reference Writing," being slightly better in certain aspects but generally equivalent overall.
        \item \textbf{7–8 points}: The “Writing to Be Evaluated” is stronger in most evaluation criteria compared to the “Reference Writing,” with only minor shortcomings.
        \item \textbf{9–10 points}: The “Writing to Be Evaluated” excels across nearly all criteria and demonstrates exceptional quality overall.
    \end{itemize}
    
    \item \textbf{Output Format}:
    Please present the evaluation outcome in the following format:
\end{enumerate}

Comparative Analysis: 

1. Opening Section: Analysis content…… 

2. Closing Section: Analysis content…… 

Score: [[X]] 

Reasons for the Score: …… 

\textbf{Important Notes}:
\begin{itemize}
    \item Summarize key points concisely while maintaining strict logical coherence.
    \item Use double square brackets (e.g., [[6]]) to indicate the score. Final scores must be whole numbers between 1 and 10.
\end{itemize}

Please proceed with the evaluation.
\end{tcolorbox}

\subsection{Metaphor}

\begin{tcolorbox}[title = {Metaphor Prompt}, breakable]
\textbf{Input}: Instruction, Reference, Content
\tcblower

Please evaluate the "Writing to Be Evaluated" and the "Reference Writing" in terms of language richness and appropriateness of rhetoric. Your task is to analyze the strengths and weaknesses of the "Writing to Be Evaluated" based on the provided assessment criteria, ultimately assigning it a score from 1 to 10 and briefly explaining your reasoning.

You will be provided with the following items:
\begin{itemize}
    \item \textbf{Assessment Criteria}: Guidelines on language richness and appropriate rhetoric, divided into positive and negative scenarios.
    \item \textbf{Writing Instructions}: Requirements, background, and main themes for both pieces of writing.
    \item \textbf{Reference Writing and Writing to Be Evaluated}: Two writing excerpts for comparison.
\end{itemize}

[Assessment Criteria Start]

\textbf{I. Positive Scenarios}

\begin{enumerate}
    \item \textbf{Language Richness}
    \begin{itemize}
        \item Uses a diverse range of vocabulary, avoiding repetition or monotony, which demonstrates flexibility in written expression.
        \item Exhibits expressiveness, precisely depicting scenes (e.g., landscapes, characters, psychological activities, and actions) and rendering the content vivid and lively.
        \item Shows meticulous attention to language, enhancing the cultural or artistic appeal of the writing through deliberate word choices.
    \end{itemize}

    \item \textbf{Excellence and Appropriateness of Rhetoric}
    \begin{itemize}
        \item Proper use of rhetorical devices contributes to vivid expression, depth of thought, or emotional impact (e.g., metaphors, repetition, parallelism, personification).
        \item Rhetorical devices align with the logical flow of the content, avoiding excessive embellishment and enhancing the power of communication.
        \item Writing techniques are not overdone; rhetoric is seamlessly integrated, blending naturally with the context and theme.
    \end{itemize}

    \item \textbf{Structure and Logic}
    \begin{itemize}
        \item The opening captures the reader's attention with clear and compelling language and ideas.
        \item The conclusion is impactful and summarizes effectively, elevating the main theme or inspiring further thought.
    \end{itemize}
\end{enumerate}

\textbf{II. Negative Scenarios}

\begin{enumerate}
    \item \textbf{Language Deficiency}
    \begin{itemize}
        \item Monotonous vocabulary or overuse of repetitive words (e.g., frequent use of simple synonyms), making the expression weak or immature.
        \item Sentences are poorly constructed, or grammatical errors noticeably disrupt the reading flow.
        \item Generic or meaningless information dominates the content (e.g., vague discussions lacking specific details).
    \end{itemize}

    \item \textbf{Inappropriate Use of Rhetoric}
    \begin{itemize}
        \item Lack of rhetorical devices or reliance on only one type, resulting in an overly flat or uninspired expression.
        \item Improper application of rhetorical devices (e.g., forced metaphors or overly complex sentences) lowers the overall quality.
        \item Awkwardly inserted rhetoric disrupts the logical flow or diverges from the main theme.
    \end{itemize}

    \item \textbf{Structural Problems Affecting Expression}
    \begin{itemize}
        \item Overusing simple connectors (e.g., "firstly, secondly, lastly") where the structure weakly relates to the logical content.
        \item Failing to echo the main theme (e.g., conclusions that do not summarize critical points or openings that lack appeal).
        \item Expression is limited to purely narrative progression or listing points, lacking deeper analysis or detailed depiction (e.g., bland storytelling or redundant argument repetition).
    \end{itemize}

    \item \textbf{Mixed Language Issues}
    \begin{itemize}
        \item Entirely mixed language styles (e.g., "我 say 这个不行" or "highlight 了这页 slide"); deduct 2 points.
        \item Unnecessarily retaining translatable English abbreviations (e.g., using "WC" directly instead of translating it to Chinese); deduct 1 point.
        \item Note: Conventional names like "KFC" and "NBA" left untranslated are acceptable unless they poorly match the context.
    \end{itemize}

    \item \textbf{Content Limitations in Argumentative or Fiction Writing}
    \begin{itemize}
        \item \textbf{Fiction}: Lack of detailed portrayal of characters or psychological depth, relying solely on surface-level narrative; merits deductions based on the importance of the plot.
        \item \textbf{Essay}: Circular reasoning with no progressive analysis (e.g., merely listing pros and cons without further comparative summaries); merits a 1–2 point deduction.
    \end{itemize}
\end{enumerate}

[Assessment Criteria End]

\textbf{Writing Instructions}

\{instruction\}

\textbf{Reference Writing}

\{reference\}

\textbf{Writing to Be Evaluated}

\{content\}

\textbf{Evaluation Process}

Follow the steps below to complete the evaluation. Avoid contradictory logic:

\begin{enumerate}
    \item \textbf{Strengths and Weaknesses Comparative Analysis}: Compare the performance of the "Writing to Be Evaluated" and the "Reference Writing" based on the assessment criteria, analyzing their relative strengths and weaknesses step by step.
    
    \item \textbf{Scoring Based on Reference Benchmark}: The fixed score for the "Reference Writing" is \textbf{6 points}, which serves as the baseline. Use the scoring guidelines below to evaluate the writing:
    \begin{itemize}
        \item \textbf{1–2 Points}: The "Writing to Be Evaluated" exhibits significantly more negative scenarios compared to the "Reference Writing," with almost no positive scenarios. It appears overly simplistic or lacks concrete examples.
        \item \textbf{3–4 Points}: The "Writing to Be Evaluated" exhibits slightly more negative scenarios and fewer positive scenarios than the "Reference Writing." The content lacks vividness or convincing argumentation.
        \item \textbf{5–6 Points}: The "Writing to Be Evaluated" exhibits positive and negative scenarios on par with the "Reference Writing," demonstrating average performance. It uses some rhetoric, techniques, or examples, albeit less naturally.
        \item \textbf{7–8 Points}: The "Writing to Be Evaluated" exhibits negative scenarios similar to the "Reference Writing" but features more positive scenarios. The language flows naturally and is enriched with strong expressions or examples.
        \item \textbf{9–10 Points}: The "Writing to Be Evaluated" exhibits significantly more positive scenarios compared to the "Reference Writing," with little to no negative scenarios. It demonstrates rich, vibrant, and well-balanced expression while maintaining logical progression.
    \end{itemize}
    
    \item \textbf{Output Format}: Deliver the evaluation in the following format:
\end{enumerate}

Strengths and Weaknesses Analysis: 

1. Positive Scenarios Comparison: Analysis content…… 

2. Negative Scenarios Comparison: Analysis content…… 

Score: [[X]] 

Score Reasoning: …… 

\textbf{Notes}:
\begin{itemize}
    \item Summarize key points succinctly while ensuring logical consistency.
    \item The score must be enclosed in double square brackets (e.g., [[6]]). The final score must be a whole number between 1 and 10 inclusively.
\end{itemize}

Begin your evaluation.
\end{tcolorbox}

\subsection{Logics}

\begin{tcolorbox}[title = {Logics Prompt}, breakable]
\textbf{Input}: Instruction, Reference, Content
\tcblower

Please act as a professional writing reviewer and evaluate the reasoning logic in the "Writing to Be Evaluated" and the "Reference Writing" provided below. Your task is to analyze the strengths and weaknesses of the "Writing to Be Evaluated" based on the evaluation criteria, provide a score between 1 and 10, and briefly explain your reasoning.

You will receive the following:
\begin{itemize}
    \item \textbf{Evaluation Criteria}: Descriptions of ideal scenarios and poor performances regarding logical reasoning.
    \item \textbf{Writing Instructions}: Requirements, context, and main themes for the two pieces of writing.
    \item \textbf{Reference Writing and Writing to Be Evaluated}: Two writing fragments for comparison.
\end{itemize}

[Evaluation Criteria Start]

\textbf{The core focus of content logic:}
\begin{enumerate}
    \item \textbf{Consistency in Person/Point of View}: Does the entire piece maintain a consistent narrative style and perspective, avoiding abrupt shifts? If there are changes in person or perspective, is there prior groundwork or subsequent explanation?
    \item \textbf{Logical Coherence}: Is the reasoning process internally consistent? Do the ideas naturally connect without abrupt breaks or gaps in logic?
    \item \textbf{Consistency in Language Style and Tone}: Does the expression retain a consistent tone and style throughout, ensuring a smooth and natural reasoning process?
\end{enumerate}
\textit{Note}: Reasoning logic typically pertains to issues within a single paragraph or across a few adjacent paragraphs. The focus is on maintaining contextual continuity and logical consistency.

\textbf{A. Ideal Scenarios (Exemplars of Excellent Reasoning Logic):}

\textbf{1. Logical Reasoning in Argumentative Writing:}
\begin{itemize}
    \item The reasoning process is tightly interconnected, with clearly defined logical layers. Ideas progress naturally from common knowledge to deeper analysis, enabling readers to follow the step-by-step reasoning.
    \item The content includes both abstract theoretical analysis and concrete evidence to support conclusions, forming naturally persuasive arguments.
\end{itemize}

\textbf{2. Structure in Speeches or Addresses:}
\begin{itemize}
    \item The logic is clear and straightforward: identify the core issue, present viewpoints for addressing the issue, and then explain steps or solutions with specificity.
    \item Ideas transition from macro-level problem analysis to specific actionable methods, culminating in an inspiring conclusion with layered content.
\end{itemize}

\textbf{3. Writing in Application Letters or Summary Reports:}
\begin{itemize}
    \item Reports unfold content logically following a structure like "Objective $\rightarrow$ Problem Analysis $\rightarrow$ Key Challenges $\rightarrow$ Solutions $\rightarrow$ Achieved Results," reflecting clear thought processes and reasoning.
    \item Application letters discuss the attributes and significance of the requested entity while precisely aligning it with the applicant's needs, forming a tight connection and enhancing the reasonableness and persuasiveness of the content.
\end{itemize}

\textbf{B. Poor Performances (Issues and Deduction Standards):}

If a piece displays the following logical problems, scores should be deducted accordingly:

\textbf{1. Issues in Argumentative Writing:}
\begin{itemize}
    \item The content merely lists opinions without reasoning or evidence (e.g., “We should XXXX, we must XXXX”), lacking justification or examples. This glaring lack of logic warrants at least a 3-point deduction.
\end{itemize}

\textbf{2. Problems in Speeches or Addresses:}
\begin{itemize}
    \item Analysis or solutions lack depth or breadth. For example, overly abstract discussions without actionable plans, or overly specific ideas without high-level insights. Deduct 3–6 points based on severity.
\end{itemize}

\textbf{3. Flaws in Application Letters or Summary Reports:}
\begin{itemize}
    \item Work summaries only state actions performed without analysis or reasoning, making the content superficial.
    \item Application letters fail to establish alignment between the requester's needs and the requested entity, resulting in vague or disconnected content. Deduct 3–6 points based on the severity of the issues.
\end{itemize}

[Evaluation Criteria End]

\textbf{[Writing Instructions Start]}

\{instruction\}

\textbf{[Writing Instructions End]}

\textbf{[Reference Writing Start]}

\{reference\}

\textbf{[Reference Writing End]}

\textbf{[Writing to Be Evaluated Start]}

\{content\}

\textbf{[Writing to Be Evaluated End]}

\textbf{Evaluation Process}

Please strictly follow the steps below to complete the evaluation, avoiding contradictions in logic:

\begin{enumerate}
    \item \textbf{Comparative Analysis of Strengths and Weaknesses}: Compare the performance of the "Writing to Be Evaluated" against the "Reference Writing" based on the evaluation criteria, gradually analyzing their relative strengths and weaknesses across all points.

    \item \textbf{Scoring with Reference Baseline}: Use the "Reference Writing" as the \textbf{8-point standard of excellence}. Start with 8 points for the "Writing to Be Evaluated" and adjust the score as follows:
    
    \textbf{Deduction Rules:}
    \begin{itemize}
        \item For 0–1 minor weakness or logical flaw, deduct 0–3 points.
        \item For 1–2 minor weaknesses, deduct 3–6 points.
        \item For 2 or more minor weaknesses, or 1 significant logical problem, deduct 6–8 points. Severe issues may justify further deductions.
    \end{itemize}
    
    \textbf{Additional Points:}
    \begin{itemize}
        \item If the evaluated writing demonstrates clear advantages over the reference writing, you can award additional points, up to a maximum of 2 points.
    \end{itemize}
    
    \textit{Other Notes}: A score between 1–4 does not necessarily mean poor reasoning logic overall. Simply base your score strictly on the weaknesses analyzed.

    \item \textbf{Output Format}: Deliver your evaluation result in the following format:
\end{enumerate}

Strengths and Weaknesses Analysis: 

1. Comparison of Positive Aspects: Analysis content… 

2. Comparison of Negative Aspects: Analysis content… 

Score: [[X]] 

Reason for Score: … 

\textbf{Notes}:
\begin{itemize}
    \item Briefly summarize key points while ensuring logical rigor.
    \item The "Score" section must use double square brackets to enclose the number (e.g., [[6]]). The final score must be a whole number between 1 and 10.
\end{itemize}

Please begin your evaluation.
\end{tcolorbox}

\subsection{Emotion}

\begin{tcolorbox}[title = {Emotion Prompt}, breakable]
\textbf{Input}: Instruction, Reference, Content
\tcblower

Please act as a professional writing reviewer and evaluate the emotional expression quality of the "Writing to Be Evaluated" and the "Reference Writing." Your task is to analyze the strengths and weaknesses of the "Writing to Be Evaluated" based on the given evaluation criteria. Then, provide a score between 1 and 10 and briefly explain your reasoning.

You will be given the following content:
\begin{itemize}
    \item \textbf{Evaluation Criteria}: Guidelines on emotional expression, divided into good and poor conditions.
    \item \textbf{Writing Instructions}: The requirements, background, and main themes of the two pieces of writing.
    \item \textbf{Reference Writing and Writing to Be Evaluated}: Two writing excerpts for comparison.
\end{itemize}

[Evaluation Criteria Start]

Emotional expression must connect to the intended audience. This means the author should clearly consider the target readers before conveying emotions.

\textbf{Good Conditions:}
\begin{enumerate}
    \item Emotion is successfully integrated into various descriptions (e.g., events, scenery, character portrayals), making the writing warm and layered while enhancing emotional tension.
    \item Skillful use of rhetorical devices (such as metaphor or personification) improves emotional expressiveness, creating greater visual appeal and emotional impact.
    \item Tone, vocabulary, and sentence structures match the target reader's style. The emotions are genuine and fluid, capable of resonating deeply with readers or sparking contemplation.
    \item Emotional expression aligns closely with the main theme, helping to drive the narrative forward or deepen core points.
\end{enumerate}

\textbf{Poor Conditions:}
\begin{enumerate}
    \item Ineffective or missing emotional expression that fails to convey the intended feeling or is disconnected from the emotional context.
    \item Misuse of rhetorical devices (e.g., improper metaphors or unrelated comparisons) that weakens the emotional expression.
    \item Emotional expression is superficial, exaggerated, or unnatural, making it difficult for readers to truly relate or empathize.
\end{enumerate}

[Evaluation Criteria End]

\textbf{Writing Instructions}

\{instruction\}

\textbf{Reference Writing}

\{reference\}

\textbf{Writing to Be Evaluated}

\{content\}

\textbf{Evaluation Process}

Please strictly follow the steps below and avoid contradictory logic:

\begin{enumerate}
    \item \textbf{Strengths and Weaknesses Analysis}: Analyze the relative strengths and weaknesses of the "Writing to Be Evaluated" compared to the "Reference Writing" based on the evaluation criteria, systematically addressing each point.
    
    \item \textbf{Scoring and Reference Baseline}: The "Reference Writing" is fixed at a score of \textbf{6}, which serves as the baseline for comparison. Assign a score to the "Writing to Be Evaluated" using the following scoring rules:
    \begin{itemize}
        \item \textbf{1–2 Points}: The "Writing to Be Evaluated" displays significantly more poor conditions than the reference. It lacks good conditions, leading to a monotonous, shallow text with weak emotional portrayal or insufficient author sentiment.
        \item \textbf{3–4 Points}: The "Writing to Be Evaluated" displays somewhat more poor conditions than the reference, and fewer good conditions. The text has some emotional portrayal but remains mediocre, or it contains abrupt, unsuitable elements.
        \item \textbf{5–6 Points}: The "Writing to Be Evaluated" showcases a balance of good and poor conditions similar to the reference. It performs adequately and achieves a baseline level of reader emotional engagement.
        \item \textbf{7–8 Points}: The "Writing to Be Evaluated" demonstrates an emotional quality close to the reference but features additional good conditions. It portrays a variety of complementary emotions or utilizes techniques like environmental descriptions to convey feelings appropriately.
        \item \textbf{9–10 Points}: The "Writing to Be Evaluated" exhibits significantly more good conditions than poor ones, displaying rich and nuanced emotions. Readers are deeply moved by the author's sentiment, or emotions are masterfully conveyed through elements like environmental descriptions.
    \end{itemize}
    
    \item \textbf{Output Format}: Use the following format for your evaluation result:
\end{enumerate}

Strengths and Weaknesses Analysis: 

1. Comparison of Good Conditions: Analysis content... 

2. Comparison of Poor Conditions: Analysis content... 

Score: [[X]] 

Reason for the Score: ... 

\textbf{Notes}:
\begin{itemize}
    \item Concisely summarize key points, ensuring logical rigor.
    \item Use double square brackets to denote the final score (e.g., [[6]]). The final score must be an integer between 1 and 10.
\end{itemize}

You may now proceed.
\end{tcolorbox}

\subsection{Plots}

\begin{tcolorbox}[title = {Plot Prompt}, breakable]
\textbf{Input}: Instruction, Reference, Content
\tcblower

Please act as a professional writing evaluator and assess the appropriateness of the plot design and development in the "Writing to Be Evaluated" and the "Reference Writing." Your task is to analyze the strengths and weaknesses of the "Writing to Be Evaluated" based on the provided evaluation criteria, then assign a score from 1 to 10, and briefly explain your reasoning.

You will receive the following content:
\begin{itemize}
    \item \textbf{Evaluation Criteria}: Explanation of plot design and development, divided into good and bad scenarios.
    \item \textbf{Writing Instructions}: Requirements, background, and central themes for both pieces of writing.
    \item \textbf{Reference Writing and Writing to Be Evaluated}: Two writing samples for comparison.
\end{itemize}

[Evaluation Criteria Start]

\textbf{I. Good Scenarios}

\begin{enumerate}
    \item \textbf{Structure \& Logic}:
    \begin{itemize}
        \item The overall structure is clear, well-organized, and logically sound.
        \item Paragraph transitions are smooth and coherent, using lively inter-sentence transitions instead of relying on mechanical connectors (e.g., "firstly, secondly, then").
        \item The plot unfolds in an orderly manner with internal consistency, including foreshadowing that is later resolved or explained.
    \end{itemize}
    
    \item \textbf{Narrative Techniques}:
    \begin{itemize}
        \item Skillful use of techniques such as flashbacks or insertions, resulting in a concise narrative with a clear focus.
        \item Well-paced progression, demonstrating the ability to create tension at key plot points, capturing reader interest, or building suspense.
        \item Narrative logic is reasonable and guided by causality or thematic relevance; the plot does not appear abrupt or forced.
    \end{itemize}
    
    \item \textbf{Thematic Value}:
    \begin{itemize}
        \item The thematic content is novel and meaningful, able to resonate with readers or provoke deeper thought.
        \item The narrative conveys insightful perspectives or elevated value orientations.
    \end{itemize}
    
    \item \textbf{Fit to Writing Type}:
    \begin{itemize}
        \item \textbf{Argumentative essay}: Clear thesis, well-organized points, sufficient and relevant evidence.
        \item \textbf{Fiction}: Characters and plot serve the theme; the narrative is tight and vivid.
        \item \textbf{Prose}: Beautiful imagery, a unified theme, and lively, evocative language.
        \item \textbf{Speech, Application, Report}: Concise, focused, logically flowing, and persuasive.
    \end{itemize}
\end{enumerate}

\textbf{II. Bad Scenarios}

\textit{(Evaluation basis: Minor issues merit a 3-point deduction; significant or severe problems merit a 6-point deduction.)}

\begin{enumerate}
    \item \textbf{Structure \& Organization Issues}:
    \begin{itemize}
        \item The structure is disorganized with no hierarchical progression.
        \item Paragraphs or sections lack transitions and connections, with abrupt jumps or content piling up, leading to confusion.
        \item Plot progression is flat and tedious, lacking attraction or logical coherence.
    \end{itemize}
    
    \item \textbf{Narrative Issues}:
    \begin{itemize}
        \item Improper use of flashbacks/insertions, resulting in confusion or breaking reading immersion.
        \item Plot design is loose or fragmented, failing to establish tension or draw interest.
        \item Plot development lacks causality or necessary foreshadowing, presenting unconvincing or abrupt logic.
    \end{itemize}
    
    \item \textbf{Thematic Issues}:
    \begin{itemize}
        \item The theme is cliché, dull, or lacks originality, failing to engage or resonate with readers.
        \item The intention remains superficial, lacking depth or meaningful development.
        \item Lacks thematic unity; content is scattered, making it hard to consolidate into an overall viewpoint or main idea.
    \end{itemize}
    
    \item \textbf{Appropriateness Issues}:
    \begin{itemize}
        \item \textbf{Argumentative essay}: Thesis deviates from the topic, points are simplistic or drawn out, and evidence is insufficient or weak.
        \item \textbf{Fiction}: Characters are stiff and lack appeal; plot design has no bearing on the theme or is disconnected from character behavior.
        \item \textbf{Prose}: Imagery is bland and lacks vividness; fragmented content fails to support the central theme.
        \item \textbf{Speech, Application, Report}: Sentences are verbose or too casual, lack focus, and suffer from logical disconnections.
    \end{itemize}
    
    \item \textbf{Language Issues}:
    \begin{itemize}
        \item Language is simplistic or monotonous, lacking vitality and expressiveness.
        \item Writing is obscure or hard to understand, or the style does not match the intent.
        \item Excessive use of mechanical or formulaic sentences, lacking the writer’s individual voice.
    \end{itemize}
\end{enumerate}

[Evaluation Criteria End]

\textbf{Writing Instructions}

\{instruction\}

\textbf{Reference Writing}

\{reference\}

\textbf{Writing to Be Evaluated}

\{content\}

\textbf{Evaluation Process}

Please strictly follow these steps for evaluation and avoid contradictory logic:

\begin{enumerate}
    \item \textbf{Comparative Analysis}: Using the evaluation criteria, analyze the strengths and weaknesses of the "Writing to Be Evaluated" in comparison to the "Reference Writing," covering all relevant points.
    
    \item \textbf{Scoring and Reference Baseline}: The "Reference Writing" is assigned a fixed score of \textbf{6}, serving as the baseline for assessment. Based on the performance of the "Writing to Be Evaluated," use the following rules to determine your score:
    \begin{itemize}
        \item \textbf{1–2 Points}: The "Writing to Be Evaluated" displays clearly more bad elements than good compared to the reference. The narrative is largely a dry recounting, lacking linguistic richness and rhetorical depth, with thin content, weak logic, and significant flaws.
        \item \textbf{3–4 Points}: The "Writing to Be Evaluated" displays somewhat more bad elements than the reference. The narrative is plain and unengaging, the plot is relatively bland, with weak rhetorical and detailed descriptions, but the overall logic is basically coherent.
        \item \textbf{5–6 Points}: The "Writing to Be Evaluated" and the "Reference Writing" have roughly equivalent numbers of good and bad elements. The language is mostly fluent, the plot is complete with no major flaws, and the structure is unified with some internal cohesion, but highlights are lacking.
        \item \textbf{7–8 Points}: The "Writing to Be Evaluated" displays a similar number of bad elements as the reference but clear strengths as well. There are highlights in language or plot, such as successfully catching the reader's attention, engaging twists, tightly structured logical progression, and a layered, impactful narrative.
        \item \textbf{9–10 Points}: The "Writing to Be Evaluated" displays obviously more good elements than the reference. The plot is full of tension, the logic is rigorous and concise, with distinctive strengths across multiple aspects, demonstrating high-level writing and artistry.
    \end{itemize}
    
    \item \textbf{Output Format}: Please present your evaluation in the following format:
\end{enumerate}

Strengths and Weaknesses Analysis:

1. Comparison of Good Elements: Analysis...

2. Comparison of Bad Elements: Analysis...

Score: [[X]]

Reason for Score: ...

\textbf{Notes}:
\begin{itemize}
    \item Briefly summarize key points, ensuring logical clarity.
    \item The score should be wrapped in double square brackets, e.g., [[6]]. The final score must be an integer between 1 and 10.
\end{itemize}

Begin your evaluation.
\end{tcolorbox}

\subsection{Paragraphing}

\begin{tcolorbox}[title = {Paragraphing Prompt}, breakable]
\textbf{Input}: Instruction, Reference, Content
\tcblower

Please evaluate the rationality of the chapter and paragraph divisions in the "Writing to Be Evaluated" according to the following principles:

\begin{enumerate}
    \item \textbf{Balance Between Sections}: Avoid overly long paragraphs (exceeding 500 characters) or having a single chapter that is disproportionately longer than all other chapters combined. Ideally, the length of each section should not vary significantly.
    
    \item \textbf{Necessity of Chapter Division}: Dividing content by chapters using titles is necessary for texts with strong structural requirements, such as lengthy novels, reports, or speeches. Avoid meaningless chapter divisions, such as splitting text into separate chapters when it could logically be merged into one. Ideally, each chapter should aim for a word count exceeding 800 characters.
    
    \item \textbf{Necessity of Paragraph Division}: All writing (aside from poetry) should avoid excessive paragraphing, such as having a majority of paragraphs consisting of only a single sentence. For poetry, stanzas or paragraphs can be divided according to rhythm. Additionally, avoid failing to segment thoughts—do not lump too many distinct themes or arguments into a single paragraph. Ideally, each paragraph should exceed 50 characters, though briefer paragraphs can be used for rhythmic or tonal purposes.
\end{enumerate}

Based on the three points above, score the text as follows:
\begin{itemize}
    \item \textbf{3 Points}: Division is reasonable and well-structured.
    \item \textbf{2 Points}: Division is somewhat flawed in its chapters or paragraphs.
    \item \textbf{1 Point}: Division is clearly uncoordinated, lacks structured planning, or the evaluated content is missing.
\end{itemize}

\textbf{Writing Instructions}

\{instruction\}

\textbf{Writing to Be Evaluated}

\{content\}

\textbf{Evaluation Process}

After assigning a score, provide a brief explanation of your rating. Provide your review with the score enclosed in double square brackets. For example:

Overall Score: [[2]]

Reason: ...

\end{tcolorbox}

The final scores are projected to a 1-10 scale by $y = 5 \times(x-1)$

\subsection{Impression}

\begin{tcolorbox}[title = {Impression Prompt}, breakable]
\textbf{Input}: Instruction, Reference, Content
\tcblower

Please provide an overall impression score based on the provided writing instructions and content. When evaluating, focus on the following aspects:

\begin{itemize}
    \item The attractiveness and engagement of the opening and closing sections.
    \item The overall fluency and logical coherence of the text.
    \item The appropriateness of the paragraph structure.
    \item The avoidance of excessive reliance on bullet points and lists.
\end{itemize}

\textbf{Writing Instructions}

\{instruction\}

\textbf{Reference Writing}

\{reference\}

\textbf{Writing to Be Evaluated}

\{content\}

\textbf{Evaluation Process}

Please provide an overall impression score as an integer between 1 and 10, and briefly explain your reasoning. You must format the final score within double square brackets. For example:

Overall Score: [[8]]

Reason: ...

\end{tcolorbox}

\subsection{Heading Parsing Regex}
\label{sec:appendix-heading-parser}

The implementation details of our hierarchical heading parsing algorithm are provided in Listing \ref{lst:parse_headings}.

\begin{lstlisting}[caption={Python implementation of the heading parsing algorithm.}, label={lst:parse_headings}]
def parse_headings(text):
    # regex for headings
    patterns = [
        (r'^(#{1,6})\s*(.*)$', 'markdown'),                   # Markdown title (#, ##, ### etc.)
        (r'^([一二三四五六七八九十]+[、.])\s*(.*)$', 'chinese'),  # Chinese title (一、二、三 etc.)
        (r'^（([一二三四五六七八九十])）\s*(.*)$', 'chinese_second'),  # Chinese title (（一）、（二）、（三）等)
        (r'^(\d{1,2}\.)\s*(.*)$', 'ordered'),                # ordered list (1. 2. 3.) 
        (r'^(\-)\s*(.*)$', 'unordered'),                # unordered list (-)
    ]
    
    # Initial hierarchical structure
    root = {'title': 'Root', 'subtopics': [], 'type': 'root'}
    stack = [root]  # Initialize stack; root level node holds all top-level nodes

    base_hash = -1
    
    # Split the text by lines and parse
    for line in text.splitlines():
        line = line.strip()  # Remove extra spaces from both ends
        if not line:
            continue  # Skip blank lines
        
        for pattern, kind in patterns:
            match = re.match(pattern, line)
            if match:
                content = ""
                if kind == 'markdown':
                    hashes, content = match.groups()
                    if base_hash == -1:
                        base_hash = len(hashes)
                    level = len(hashes) - base_hash + 1 + 1  # The number of # indicates level
                elif kind == 'chinese':
                    prefix, content = match.groups()
                    level = 2
                elif kind == 'chinese_second':
                    prefix, content = match.groups()
                    level = 3
                elif kind == 'unordered':
                    prefix, content = match.groups()  # Unordered list is always one level down
                    if stack[-1]['type'] != kind:
                        level = len(stack) + 1
                    else:
                        level = len(stack) 
                elif kind == 'ordered':
                    prefix, content = match.groups()  # Ordered list next to unordered list
                    if stack[-1]['type'] == 'ordered':
                        level = len(stack)
                    elif stack[-1]['type'] == 'unordered':
                        if stack[-2]['type'] == 'ordered': # Unordered is a child of ordered
                            level = len(stack) - 1
                        else:   # Unordered and ordered are at the same level
                            level = len(stack)
                    else:
                        level = len(stack) + 1
                        
                # Create the current node
                node = {'title': content.strip(), 'subtopics': [], 'type': kind}

                # Adjust the stack: ensure the parent node is at the top
                while len(stack) >= level:
                    stack.pop()

                # Add the current node to the parent node's subtopics
                stack[-1]['subtopics'].append(node)
                stack.append(node)
                break

    return root
\end{lstlisting}
\section{Data Examples for Each Tasks} \label{sec:data_examples}

We show three examples for Completion, Guide, Open tasks in Table~\ref{ex:completion},~\ref{ex:guide},~\ref{ex:open}. Their English translations are presented in Table~\ref{ex:completion_en},~\ref{ex:guide_en},~\ref{ex:open_en}.

\begin{table*}[!ht]
\scriptsize
    \centering
    \begin{tabular}{p{40pt}p{380pt}}
    \toprule
    \textbf{Setting}&\textbf{Prompt}\\
    \midrule
    \textbf{Instruction} & \begin{CJK}{UTF8}{gbsn} \tiny  请根据上下文补全以下文章中用 [fill in the blank] 特殊符号标记出的内容。\end{CJK} \\ \midrule
    \textbf{Information} & \begin{CJK}{UTF8}{gbsn} \tiny 最近，有媒体盘点出了中国超级工程里的“世界之最”，在网络上引发了一大波热议： \newline 白鹤滩水电站是目前世界在建规模最大、技术难度最高的水电工程；港珠澳大桥是世界上总体跨度最长的跨海大桥；新疆和若铁路开通运营，让世界首条环沙漠铁路线完成“最后一块拼图”……\newline 有网友称，“中国制造就是中国骄傲”。\newline 而如果我们往深了扒一扒，超级工程的背后，实际上凝聚了大量自主研发的新科技，科技自强自立的背后，最终则是创新的驱动。\newline 习近平总书记在党的二十大报告中22次提到创新，并深刻指出：坚持创新在我国现代化建设全局中的核心地位。报告中还有一处提到：创新是第一动力。\newline 我们来理一理，“核心地位+第一动力”，创新的分量为何这么重？\newline 一\newline 从人类历史来看，社会生产力的每一次发展、科学技术的每一次进步，无不是通过创新实现的。\newline 欧美几个发达国家就是抓住了科技和产业革命的创新机会而一跃跨入现代化行列，实现大国崛起和民族振兴，并引领时代的走向和世界的发展。\newline 有创新就会有发展，谋创新就能谋未来。涅槃于一穷二白旧社会的中国式现代化，也经历了无数次以创新求发展的浴火重生。\newline 特别是新时代以来，在创新驱动发展战略的指引下，我国的“创新型国家”的建设稳步加快。从2012年到2021年，全社会研发投入从1.03万亿元增长到2.79万亿元，全球创新指数排名从第34位上升到第12位。\newline 科技创新在企业壮大、产业升级、区域发展、重大工程建设等方面发挥了重要作用，有力支撑了高质量发展，带动一些关键核心技术相继实现突破，取得了载人航天、探月探火、深海深地探测、超级计算机等重大成果。\newline 九天之外传来的“感觉良好”，深潜海底万米的“妙不可言”，乘坐“复兴号”飞驰万里，睁开“天眼”仰观浩渺宇宙......这些，都成了网民心中中国科技创新的“名场面”，成了我们心中升腾起的自信和自豪。\newline \newline 二\newline [fill in the blank]\newline 这样的故事还有不少。这些年，我们在科技“从模仿到创新”的转型过程中遭遇了“追赶的极限”，关键领域核心技术被“卡脖子”的问题愈发突出。\newline 特别是中美贸易摩擦中，我国“缺芯少核”的科技短板暴露了出来。美西方国家利用技术优势地位一方面禁止关键技术流入中国，推动高科技产业链的“对华脱钩”；另一方面阻碍我国核心技术研发，企图将我国彻底压制在产业链中低端。\newline 在激烈的国际竞争中，惟创新者进，惟创新者强，惟创新者胜。正是因为我国科技实力和世界领先水平的差距在不断缩小，一些领域实现了从“跟跑”到“并跑”甚至“领跑”，才引发了美西方国家的战略焦虑，并招致不惜成本的封锁和打压。\newline 然而，我们的目标绝不是跟着西方国家亦步亦趋。我们要开拓出中国式现代化路径，这是一条从未有人走过的路。为人类实现现代化提供新选择，科技创新在其中的核心作用无疑更加凸显。\newline \newline 三\newline 东部沿海省份浙江，为创新之路探了路。\newline 早在2006年，习近平同志在浙江工作时就为浙江定下了用15年时间进入创新型省份行列，基本建成科技强省的目标。当年的“全省科学技术大会”这个会议名称，被习近平同志修改为“全省自主创新大会”。几字之变，意图更加清晰，导向更加明确。\newline 一路走来，“自主创新”这面旗帜始终在之江大地上高高飘扬。今天的浙江，已经拥有良好的科创环境和氛围，三大科创高地加速打造。很多人一提到科创大走廊、之江实验室、西湖大学就想到浙江，这些高能级的平台不仅是浙江的“标签”，也正成为创新的沃土。\newline 有活力就有人才，浙江也越来越成为顶尖人才的向往之地。截至今年8月，全省研发人员总量已达77.58万人，这就意味着大概每1000个浙江人中就有12个科研人员。\newline 而这些科研平台、科研技术、创新力量，则前所未有地融入到百姓的日常当中。在全国率先启动数字化改革一年多来，浙江打造出一批实用、管用的重大应用。“海外智慧物流”“浙农服”“健康码”“政采云”……一个个有着鲜明浙江烙印的数字化应用，便企惠民，香飘墙外、飞向万家。\newline 每个时代，都有打开创新之门的钥匙。比如第一次工业革命是蒸汽机，第二次工业革命是电气化。今天，浙江则以“数”谋“新”，做第一个吃螃蟹的人。\newline \newline 四\newline 今天的世界瞬息万变。大变局之下，唯一的“不变之道”就是以变应变、以新应变。\newline 创新，该怎么创？如何新？\newline “必须坚持科技是第一生产力、人才是第一资源、创新是第一动力，深入实施科教兴国战略、人才强国战略、创新驱动发展战略，开辟发展新领域新赛道，不断塑造发展新动能新优势。”党的二十大报告中的这段话，为创新之路擘画了清晰的领域和路径。\newline \newline 此外，笔者认为，以创新驱动发展还要坚持好以下几个关键点。\newline 创新靠不得别人，还得靠自己。创新能力，关乎一个国家在世界格局中的地位，甚至关乎着国家安全。在世界竞技赛中，跟着别人跑随时可能会被绊倒，只有把创新的自主权、技术的所有权、发展的主动权紧紧攥在自己手中，才能跑出速度、跑到前列。\newline 创新的重要目的之一，是整合资源，打通链条、畅通循环。中国已经是全球第二大经济体，依靠传统的土地、资源和低成本人力来驱动发展已经没有竞争力，也不会有出路。只有用好新型举国体制优势，发挥创新的核心作用，打通不受制于人的产业链、供应链，才能在经济发展中涌现出无数“风口”，在国际竞争中站稳脚跟。  \newline 真正的创新，最终要落脚于民。近年来，我国科技创新能力不断提升，越来越多的创新成果广泛应用于民生领域。高铁网络、电子商务、移动支付、互联网+、共享经济……正在深刻改变着人们的衣食住行。不过，实现“人的现代化”也还有很多空白领域，如何围绕老百姓的切身需要，填补这些空白，是需要瞄准的“靶子”。\newline 赶考路上，需要创新来“澎湃”。坚持创新在我国现代化建设全局中的核心地位，坚持创新是第一动力，不仅要让1不断地递增出N，也要探索如何让更多的0实现1的突破。\end{CJK} \\ \midrule
    \textbf{Reference} & \begin{CJK}{UTF8}{gbsn} \tiny 科技创新是大国竞争的核心领域。一个国家科技创新能力的高低，决定了其在国际竞争中的水平。\newline 一个经典的故事是，1960年前后，一套重量为3公斤的精密光学坐标镗床主轴轴承，外商对我们的要价竟相当于和轴承同等重量的黄金或6吨重的对虾。直到我们通过自主创新成功攻关，才不再需要依赖进口。\newline 这至少告诉我们两个道理：\newline 第一，关键核心技术要不来、买不来、讨不来。只有把它牢牢攥在自己手中，才能从根本上保障国家总体安全。\newline 第二，在现代世界体系中，不同国家有着不同的分工。位于“中心地区”的发达国家享有先进技术和高附加值产业，而位于“边缘地区”的欠发达国家只能提供原材料、自然资源和廉价劳动力。这一格局让资本和价值源源不断地向“中心地区”聚集并导致严重的两极分化。 \end{CJK} \\
    \bottomrule
    \end{tabular}
    \caption{Example for \textbf{Completion} in Chinese.}
    \label{ex:completion}
\end{table*}

\begin{table*}[!ht]
\scriptsize
    \centering
    \begin{tabular}{p{40pt}p{400pt}}
    \toprule
    \textbf{Setting}&\textbf{Prompt}\\
    \midrule
    \textbf{Instruction} & \tiny  Complete the contents whose position is marked with [fill in the blank] according to contexts. \\ \midrule
    \textbf{Information} & \tiny Recently, the media compiled a list of “world’s best” super projects in China, sparking lively discussions online:  
\newline Baihetan Hydropower Station is currently the largest under-construction hydropower project in the world, with the highest technical difficulty; the Hong Kong–Zhuhai–Macau Bridge is the longest cross-sea bridge in the world; the opening and operation of the Xinjiang Hotan-Ruoqiang Railway has completed the “last piece of the puzzle” for the world’s first desert-circling railway line...  
\newline Some netizens remarked, "Made in China is China's pride."  
\newline However, when we dig deeper, we find that behind these super projects lie significant new technologies developed independently, backed by the drive for technological self-reliance and self-strengthening, which in turn is fueled by innovation.  
\newline In the report to the 20th National Congress of the Communist Party of China, General Secretary Xi Jinping mentioned innovation 22 times and profoundly emphasized: Innovation must occupy the core position in China's overall modernization strategy. The report also stated: Innovation is the primary driving force.  
\newline Let’s unpack this—“core position + primary driving force.” Why does innovation weigh so heavily?  
\newline  
\newline **I**  
\newline In human history, every advancement in social productivity and every progress in science and technology has always been achieved through innovation.  
\newline Several developed Western countries, such as those in Europe and North America, managed to seize the opportunities brought by technological and industrial revolutions, propelling themselves into the ranks of modernized nations, achieving national rejuvenation and rise to prominence, and leading the trajectory of their times and global progress.  
\newline Where there is innovation, there is development; where there is a plan for innovation, there is a plan for the future. China's modernization, which rose from a once-impoverished and backward society, has also undergone countless “phoenix-like rebirths” driven by innovation to seek development.  \newline Especially since the advent of the new era, under the guidance of the innovation-driven development strategy, China has been steadily accelerating its progress as an “innovative nation.” From 2012 to 2021, nationwide R\&D expenditures increased from 1.03 trillion yuan to 2.79 trillion yuan, and the global innovation index ranking rose from 34th to 12th.  
\newline Technological innovation has played a vital role in driving business growth, industrial upgrades, regional development, and the construction of major projects. It strongly supports high-quality development, enabling breakthroughs across critical core technologies in areas such as manned spaceflight, lunar and Mars exploration, deep-sea and deep-earth exploration, and supercomputers.  
\newline The “feeling good” phrase transmitted from outer space, the “beyond words” achievement of deep-sea dives exceeding 10,000 meters, the miles sped through on the “Fuxing” bullet train, and the vast universe explored using the “Sky Eye”... All these iconic moments of China's technological innovation have captured netizens’ imaginations, igniting pride and confidence in all our hearts.  
\newline  
\newline **II**  
\newline [fill in the blank]  
\newline There are many more stories like this. In recent years, during China's transformative journey from “imitation” to “innovation,” we have encountered the “limits of catching up,” with challenges in critical core technologies increasingly coming to the forefront.  
\newline Particularly during the U.S.-China trade friction, the technological shortcomings labeled as China’s “chip deficiency” and “lack of core technologies” were laid bare. Western countries, leveraging their technical dominance, simultaneously imposed bans on transferring critical technologies to China and tried to “decouple” high-tech industrial chains from China. They also sought to obstruct China's R\&D of core technologies in an attempt to suppress China to the lower ends of the industrial chain.  
\newline In the fierce international competition, only those who innovate advance, only those who innovate become stronger, and only those who innovate win. It is precisely because the gap between China's technological strength and world-leading levels is narrowing, with some fields accomplishing shifts from “running behind” to “running alongside” or even “leading,” that strategic anxiety has arisen among Western countries, prompting them to resort to cost-no-object blockades and suppression.  
\newline However, our goal is not to follow in the footsteps of Western nations. Our aim is to pioneer a uniquely Chinese path to modernization—a road never before taken. Offering humanity an alternative modernization model makes the core role of technological innovation even more prominent.  
\newline  
\newline **III**  
\newline The eastern coastal province of Zhejiang has been a trailblazer in the journey of innovation.  
\newline Back in 2006, while working in Zhejiang, Comrade Xi Jinping set the goal of making Zhejiang an innovation-oriented province within 15 years and essentially building it into a province strong in science and technology. The conference, originally named the “Provincial Science and Technology Conference,” was renamed by Xi Jinping as the “Provincial Independent Innovation Conference.” This subtle change in wording carried a clearer intent and a more focused objective.  
\newline Over time, the flag of “independent innovation” has flown high across the land of Zhejiang. Today, Zhejiang boasts an excellent environment and atmosphere for scientific and technological innovation, with three major innovation centers being rapidly developed. Mentioning the Innovation Corridor, the Zhijiang Laboratory, or Westlake University immediately brings Zhejiang to mind. These high-caliber platforms are not only among Zhejiang’s prominent “labels” but are also becoming fertile ground for innovation.   \newline Where there is vitality, there is talent. Zhejiang has increasingly become a magnet for top-tier talent. As of this August, the total number of R\&D personnel in the province had reached 775,800, meaning that approximately 12 out of every 1,000 people in Zhejiang work in research.   \newline These research platforms, technologies, and innovation resources have also been unprecedentedly integrated into the daily lives of ordinary people. Thanks to Zhejiang's bold steps in launching its digitization reform efforts, practical and user-friendly applications have emerged, such as “Overseas Smart Logistics,” “Zhejiang Agricultural Services,” “Health Code,” and “Government Procurement Cloud.” These digital services, bearing the unmistakable imprint of Zhejiang, have benefited businesses and citizens alike, extending their influence far beyond the region.   \newline Every era has its own key that unlocks the door to innovation. For instance, the steam engine in the First Industrial Revolution and electrification in the Second Industrial Revolution. Today, Zhejiang is creating the new with “data,” becoming the first to “try new things.”   \newline   \newline **IV**   \newline The world today is undergoing rapid changes. In an age of great transformations, the only “constant” is to adapt to change with change, and to respond to the new with the new.   \newline How should we innovate? What constitutes “new”?   \newline “We must uphold the principle that science and technology are the primary productive forces, talent is the primary resource, and innovation is the primary driving force. We must intensively implement the strategies of rejuvenating the country through science and education, strengthening the nation through talent, and driving development through innovation. We must continuously open new fields and tracks for development and create new momentum and new advantages for growth.” This excerpt from the 20th National Congress report outlines a clear roadmap for the path of innovation.  
\newline  
\newline Moreover, the author believes that driving development through innovation requires adhering to the following key principles:  
\newline Innovation cannot rely on others; it must depend on ourselves. Innovation capacity determines a nation’s standing in the global landscape and even its national security. In the global competition arena, following others always carries the risk of being tripped. Only by firmly grasping the autonomy of innovation, ownership of core technologies, and initiative in development can we achieve speed and move to the forefront.  
\newline One of the primary objectives of innovation is to integrate resources, streamline the chain, and ensure smooth circulation. As the world's second-largest economy, China can no longer depend on traditional drivers such as land, resources, and low-cost labor for competitiveness or growth. By utilizing the advantages of the new nationwide system and emphasizing the role of innovation, China can build an autonomous and robust industrial and supply chain, generate numerous “opportunities” for economic growth, and secure its position in international competition.  
\newline True innovation must ultimately focus on people. In recent years, China’s technological innovation prowess has steadily improved, leading to the widespread application of many innovative achievements in the realm of public welfare. High-speed rail networks, e-commerce, mobile payments, Internet+, the shared economy... all these have profoundly transformed people’s livelihoods. However, there remain many gaps in achieving “human modernization.” Addressing these gaps and meeting the genuine needs of ordinary people becomes the target to aim for.  
\newline On this challenging journey, innovation serves as the driving force that powers us forward. By maintaining innovation as the core position in China’s modernization strategy and upholding it as the primary driving force, we must not only ensure the continuous transformation of 1 into N but also explore how to turn more zeros into breakthroughs of 1.   \\ \midrule
    \textbf{Reference} & \tiny Technological innovation is the core arena of competition among major powers. The level of a country's capacity for technological innovation determines its standing in international competition. \newline A classic story goes that, around 1960, a set of precision optical coordinate boring machine spindle bearings weighing 3 kilograms was offered to us by foreign sellers at a price equivalent to either the same weight in gold or 6 tons of shrimp. It was not until we achieved a breakthrough through independent innovation that we no longer needed to rely on imports. \newline
At least two lessons can be drawn from this: \newline
First, key and core technologies cannot be obtained by asking, buying, or begging. Only by firmly holding them in our own hands can we fundamentally ensure the overall security of the nation. \newline
Second, in the modern world system, different countries have different roles. Developed countries in the “core regions” enjoy advanced technology and high value-added industries, while less developed countries in the “peripheral regions” can only supply raw materials, natural resources, and cheap labor. This structure causes capital and value to continuously flow toward the “core regions,” resulting in severe polarization. \\
    \bottomrule
    \end{tabular}
    \caption{Example for \textbf{Completion} translated to English by GPT-4.1-2025-0414.}
    \label{ex:completion_en}
\end{table*}

\begin{table*}[!ht]
\scriptsize
    \centering
    \begin{tabular}{p{40pt}p{380pt}}
    \toprule
    \textbf{Setting}&\textbf{Prompt}\\
    \midrule
    \textbf{Instruction} & \begin{CJK}{UTF8}{gbsn} \tiny 我想写一篇2000字左右的散文，主题是关于主人公在四月的雨季里，对理想与现实的思考和挣扎，以及对自我和人生真谛的探索。你能帮我写一下吗？ 故事背景设定在四月的雨季。 \end{CJK} \\ \midrule
    \textbf{Information} & \begin{CJK}{UTF8}{gbsn} \tiny 语言与表现 : 要求语言富有诗意和哲理性，多用比喻、象征等修辞手法，营造一种沉思和感悟的氛围。句子节奏舒缓，富有韵律感，体现主人公内心的波动和思考的深度。\newline 主要人物 : 主人公：一个敏感、内向，富有理想主义色彩的年轻人，对未来充满憧憬，却又常常感到迷茫和焦虑，在现实的压力下不断地进行自我反思和探索。\newline 环境设定 : 四月雨季，持续的阴雨天气，潮湿、阴冷的环境，象征着主人公内心的迷茫和压抑。\newline 主题与思想 : 探索理想与现实的平衡点，如何在现实的压力下保持内心的理想，以及对人生意义的追寻和思考。\newline \tiny \end{CJK} \\ \midrule
    \textbf{Reference} & \begin{CJK}{UTF8}{gbsn} \tiny 人间四月\newline 清明的雨还是搅乱了春色，湖中迭起的涟漪泛开，天郁沉的像一部默片。四月沉默了不说话，放映着刻有划痕的影碟，像是几十年代的声音断断续续，悲剧中的主角走得磕磕绊绊。\newline 是谁用蜡黄的胶卷，把人间留在四月，把四月给了人间。\newline 我以为世界真的如诗，符号不那么清晰，句句牵丝带意，篇篇浮华，好像四月天里雨如烟，山花正开。可我越往前走，句点变得明显，世界摘掉诗化的帽子，露出真容，头上实是一片贫瘠，倒不是无人开垦的荒地，而是挖得太深，变成了沙漠。我变成了神话里勇敢的人，却没那么聪明，伏下身开始数这些砂砾。高高在上的人望着我，颇有耐心，仿若天上星不眨眼。\newline 也许人间没有四月，人间四月只在理想主义者的朦胧诗里。如是这样，那些费力博人一笑的花儿也许就要抱憾走进土里，带着对人间的绝望，不堪地变得灰黄，那悲哀到极点的眼神没有换来分毫同情，反而是在泥土亲吻后大雨的恣意谩骂、大风的浇灌和捶打。\newline 我闭上眼睛，希望用一个夜替代白昼，用一个梦逃离现实。想象确有万千星辰躺在暮河里，梨花万千地开了，雨和风也都舒缓，像安静岛屿上安静的轻音乐。成长了多少多少年的沙滩上没有一个脚印，在那儿我不用一腔愚勇地数着沙子。 黄鹂的歌声拥抱我，我拥抱了春天，真正的春天，真正的人间四月，我的春天，我的人间四月。在那儿我不会庄周梦蝶，不会为了一个背影徘徊在河边。我也许在船上，也许在酒炉旁，柴火是新添的，沸着的炉子咕咕冒汽。我不会真的酩酊大醉，即使会了，也有人搀着我的肩膀，边骂边笑，走过梧桐路，走过芦苇荡，在湿腾腾的夜里影子被打乱又重组，重组又打乱。是汩汩的泉水、是半开的窗、是刚长出的月牙、是我把目光放进夜色里，深深的凝望。\newline 如果我知道要活得现实，我就不会在人间四月里醉生梦死。异如飞蛾扑火，我亦把那屏后灯看作了云上月。人间已是四月，只我不认，只我不觉。\newline 我只记得第一抹花色，野草中的白，碧绿浮萍中央的天鹅。不是蒲公英，风一吹飞不起来，它只是无名地生于荒芜之上，不带着迎春的使命，只是恰好张扬到了四月。慢慢的我的荒原也有植物破土抽芽，我像从某个阴暗的角落里撕扯蛛网，慢慢地由暗到明，眼睛由迷蒙到清晰。利刃往空气里劈了一下，世界的伤口裂开又愈合，也许疼只有一瞬，可疼只在这一瞬。\newline 哗啦啦落了场雨，淋湿人间，淋湿山丘和野树，淋湿四月，淋湿灵魂，淋湿行人有些脆弱的肩膀。人间多了林黛玉，一滴雨水就要撑破那娇颜，几阵风那垂柳就被摇碎。\newline 这是我的人间四月吗，这么轻，这么薄，像宣纸一样。\newline 南方正落叶，我走得快了些，走出四月的夜。还以为到了秋天，风也长出了霜，走过我的皮肤，一点清凉。可如今是四月天，再凉一阵恐就燥热了，惹人厌的虫象征着夏天的蠢蠢欲动。\newline 我独想要我的人间四月，不在永夜，不在幻梦，不在眨眼间，只完完整整地在我眼前，在不远处霄灯下，在层峦的山后面。可能我会在某处走廊找到吧，夕阳同样地斜照，驳杂的光影在地板砖，是四月的脚丫子。也可能藏在楼梯的拐角，我的一瞥；藏在几张纸片里，在花绿的伞下，雨把伞沿砸下，我的头也低下来。\newline 保温杯蒸蒸的热气、老旧的电话机。嘶哑的声音、明媚的声音、清脆的声音、沉默的声音、鸟的声音虫的声音、我的声音你的声音。红纸上的油墨字迹、一颗没有破损的心。光的影水的影、风的影雪的影、你的影我的影。如果在春天一定要有事物消亡，我会把自己献上。如果在春天一定要有一个答案，那我把自己剖开也找不到、把心脏给予也填不满。因为我早已空空如也，在那千疮百孔的十二月。\newline 从青铜到铁器，人类用了多少年；从冬春至秋夏，流云走了多少圈。从我睁开眼，到我再一次走进黑夜，再一次找到光明，还要耗费多少血液。猛然地我把头抬了起来，看见一张张漠然的脸，刹那我知道人间的命运只在一秒不到的时间里重新落向我。\newline 还有些不屑。\newline 依旧是四月天里，街巷人影绰绰，我风尘仆仆穿行而过，雨后的水洼溅起打湿了赶路人的鞋和裤管，人群那么密，步子那么紧，不同的目的却使人与人此刻相聚。我毅然决然地向前然后扑空，扑空然后退后。终于我明白一些事情刻意不得，不如停一停，尝尝巷头的茶水、巷尾的酒。还好，我走的不远，没错过太多。\newline 事实是这样，我也曾试着走到四月的尽头，看看半枯半盛的树，看看支起一半的太阳，看看也会犹豫的时间，看看寄出的信，在不高的楼台上，看看我的承诺，我的誓言。黑色的笔耕出一片片田，却写出了四月飞雪。\newline 人间四月已回，只我所想，只我所念。不是我的春天，不是我的人间四月。永远吧，我活在四月里，永远地活在四月里，永远地不欺骗。\newline 我觉着你像云，好轻好轻。 \end{CJK} \\ \bottomrule
    \end{tabular}
    \caption{Example for \textbf{Guide} in Chinese.}
    \label{ex:guide}
\end{table*}

\begin{table*}[!ht]
\scriptsize
    \centering
    \begin{tabular}{p{40pt}p{400pt}}
    \toprule
    \textbf{Setting}&\textbf{Prompt}\\
    \midrule
    \textbf{Instruction} & \tiny I want to write an essay of about 2,000 words, with the theme focusing on the protagonist’s reflections and struggles between ideals and reality during the rainy season in April, as well as their exploration of self and the true meaning of life. Can you help me write this? The story is set against the backdrop of the rainy season in April. \\ \midrule
    \textbf{Information} & \tiny Language and Expression: The language should be rich in poetic and philosophical qualities, making frequent use of metaphors, symbolism, and other rhetorical devices to create an atmosphere of contemplation and insight. The sentences should flow slowly, with a rhythmic cadence, reflecting the protagonist’s inner fluctuations and the depth of their thoughts.\newline
Main Character: Protagonist: A sensitive, introverted young person imbued with idealism; although filled with longing for the future, they are often beset by confusion and anxiety, continuously engaging in self-reflection and exploration under the pressures of reality.\newline
Setting: The rainy season in April, with persistent overcast and rainy weather; the damp, cold environment symbolizes the protagonist’s internal confusion and oppression.\newline
Theme and Ideas: The search for a balance between ideals and reality; how to maintain one’s inner ideals under the pressures of the real world, as well as the quest for and contemplation of the meaning of life. \\ \midrule
    \textbf{Reference} & \tiny The Human World in April\newline  
The Qingming rains still disrupt the spring colors, ripples rising and spreading across the lake, the sky overcast as if it were a silent film. April grows silent and speaks no words, playing scratched discs like the stop-and-go voices of past decades, the protagonist of a tragedy stumbling along.\newline  
Who used the waxy yellow film to keep the human world in April, and give April to the human world?\newline  
I thought the world truly resembled poetry, the symbols not so clear, every line connected, each passage resplendent, as if the rains in April were misty and wildflowers were blooming. But the more I walked forward, the more the periods stood out, the world took off its poetic cap and revealed its true face, barren above—not an uncultivated wasteland, but a place dug too deep, turned to desert. I became a hero from myth, brave but not clever, bending down to start counting these grains of sand. High above, someone watches me, very patient, untouched, like an unblinking star in the sky.\newline  
Perhaps there is no April in the human world—April in the human world exists only in the hazy poems of idealists. If so, then the flowers that struggle to make people smile might have to return to the earth in regret, taking their disappointment in the world with them, turning gray and yellow. Those eyes, sorrowful to the extreme, find not an ounce of sympathy, but after kissing the dirt are lashed and battered by the rain’s tirade and the wild wind’s pouring and blows.\newline  
I close my eyes, wishing to replace the day with a night, to escape reality through a dream. I imagine thousands of stars lying in the twilight river, pear blossoms blooming everywhere, rain and wind all gentle, like soft music on a quiet island. No footprints mark the shore that has grown for so many years; there, I don’t have to count the sand with my foolish courage. The song of the oriole embraces me, and I embrace spring, the real spring, the real human-world April, my spring, my human-world April. There, I won't be Zhuang Zhou dreaming of butterflies, won't linger by the riverside for a departing back. I might be on a boat, or beside a wine stove, the fire newly kindled, the boiling pot bubbling with steam. I won’t really be drunk, or even if I am, there will be someone holding my shoulder, scolding and laughing, walking with me past the Phoenix trees, through reeds, with our shadows reshuffling on humid nights, breaking and reforming, reforming and breaking. It is the gurgling spring, the half-opened window, the just-emerging crescent moon, it is my gaze let slip into the night, gazing deeply.\newline  
If I knew I had to live in reality, I wouldn’t live in a dream in the human world’s April. Like a moth to a flame, I mistake the lamp behind the screen for the moon above the clouds. It is already April in the human world, only I won’t admit it, only I haven’t noticed.\newline  
I just remember the first tinge of flowers, the white amid the weeds, the swan in the center of emerald duckweed. Not a dandelion that would fly away on the wind, but something nameless growing on the desolation, without a mission to herald spring, only blooming loud by chance in April. Slowly, my wasteland too bursts forth with shoots, as if I’m tearing cobwebs from a dark corner, emerging slowly from shadow to light, my eyes going from blurred to clear. The blade slices the air, the wound in the world splits open then heals, maybe pain is but a moment, and pain exists only in that moment.\newline  
A sudden rain falls, soaking the world, soaking hills and wild trees, soaking April, soaking the soul, soaking travelers’ somewhat fragile shoulders. The world has gained many Lin Daiyus—one drop of rain might burst that delicate face, a few gusts might shatter those drooping willows.\newline  
Is this my human-world April, so light, so thin, like rice paper?\newline  
In the south, the leaves are falling. I walk a little quicker, out from April’s night. I almost believed it was autumn, the wind frosted, brushing my skin with a coolness. But now is April, one more chilly spell and it’ll be stifling, annoying insects foreshadow summer’s restless approach.\newline  
I only want my own human-world April, not in never-ending night, not in dreams, not in a blink, but there intact before my eyes, under lanterns not far ahead, behind layered hills. Maybe I’ll find it in a corridor somewhere, with the sun slanting in as always, dappled light and shadow on the tiles—the footsteps of April. Or maybe it’s hiding around a stairwell, in a passing glance; on some scraps of paper, beneath a colorful umbrella as rain drums down its edge, and I too lower my head.\newline  
Steam from a thermos, an old-fashioned telephone. Hoarse voices, radiant voices, crisp voices, silent voices—birdsong, insect song, my voice, your voice. The ink marks on red paper, a heart as yet unbroken. Shadows of light, shadows of water, shadows of wind, shadows of snow, your shadow, my shadow. If something must perish in spring, I’d offer myself. If there must be an answer in spring, even if I cut myself open I wouldn’t find it, nor would giving my heart fill the void. For I have long been empty, ever since the battered December.\newline  
From bronze to iron, how many years for humankind? From winter and spring to autumn and summer, how many rounds have trailing clouds made? From the moment I open my eyes to the moment I return again to night and again find light, how much more blood must I spend? Suddenly, I look up, and see indifferent faces—at that instant, I know the fate of the human world falls back toward me in less than a second.\newline  
And some disdain.\newline  
It is still an April day, figures flickering in alleyways, I pass hurried and travel-worn, puddles splash and dampen the hurried shoes and pant cuffs of passersby, the crowd so dense, their steps so tight. With all these different purposes, people happen to meet at this moment. Resolutely I press forward only to reach empty air, and after missing, retreat. At last I realize some things cannot be forced; better to pause, taste a cup of tea at the entrance of the alley, some wine at its end. Luckily, I haven’t gone far, haven’t missed too much.\newline  
So it is: I tried once to walk to April’s end, to see the half-withered, half-thriving tree, the half-risen sun, time itself hesitating, the letters sent from balconies not high, my promises, my vows. The black pen plows paddies on paper, but ends up writing of snow flying in April.\newline  
April in the human world has returned, only what I think, only what I yearn for. Not my spring, not my human-world April. Forever, let me live in April, live in April forever, never deceive.\newline  
I think you are like a cloud, so light, so light.  \\ \bottomrule
    \end{tabular}
    \caption{Example for \textbf{Guide} in translated to English by GPT-4.1-2025-0414.}
    \label{ex:guide_en}
\end{table*}

\begin{table*}[!ht]
\scriptsize
    \centering
    \begin{tabular}{p{40pt}p{380pt}}
    \toprule
    \textbf{Setting}&\textbf{Prompt}\\
    \midrule
    \textbf{Instruction} & \begin{CJK}{UTF8}{gbsn} \tiny 请创作一篇分析影视作品中“配角上桌”现象的文章，探讨配角走红的原因和影响，阐述这一现象对影视创作的启示。核心观点是配角走红反映了观众审美的提升和对优质影视作品的追求。 \end{CJK} \\ \midrule
    \textbf{Information} & \begin{CJK}{UTF8}{gbsn} \tiny \end{CJK} \\ \midrule
    \textbf{Reference} & \begin{CJK}{UTF8}{gbsn} \tiny 近年来，影视剧中“配角上桌”的现象愈发明显，俨然成为影视界的新潮流，相关话题多次登上热搜，引发网友的关注和讨论。\newline 像《狂飙》中的陈书婷、李有田，《长月烬明》中的叶冰裳，《长相思》中的相柳……这些配角不再只是影视剧中的“点缀”，而频频在观众心中留下深刻印象。这一现象被形象地称为“配角上桌”，而当配角的热度超过主角时，便升级为“配角掀桌”。\newline 这不禁让人好奇，配角何以逆袭“出圈”？“配角上桌”是喜还是忧？\newline 一\newline “配角上桌”虽是一个新词，但这样的现象在影视行业中早已不是新鲜事。“考古”早年的电视剧会发现，像《金粉世家》中的白秀珠、《逆水寒》中的顾惜朝、《天下第一》里的上官海棠、《伪装者》中的汪曼春等配角，都收获了观众的喜爱。\newline 而近年来，这种现象似乎变得越来越常见，许多配角演员的影响力大幅提升，2023年更是被网友称为“配角元年”。那么，配角何以逆袭“出圈”？都有哪些“招式”？\newline 配角出彩一定程度上离不开好剧本。为了满足观众越发挑剔的眼光，编剧对配角的打造投入更多心血。配角不再只是简单的“好人”或“坏人”，而是具有多面的性格、丰富的背景故事和独特的人生观。其人设更为“带感”，富有张力，以多元化的设定引发观众共情。比如《狂飙》编剧为主角、配角都精心设置了“对照组”，像安欣与李响、孟德海与安长林等，以此凸显时代风云中有人迷失、有人坚守，既强化了戏剧性和冲击力，也让不同角色有了各自的命运沉浮，加深了观众的记忆点。正如编剧所说：“在李响身上，在曹闯身上，甚至在程程身上，其实都有对他们的追求和命运的讨论。”\newline 演员本身演技精湛、实力过硬。在深入理解角色的基础上，许多配角演员以其精湛的演技，将角色演绎得立体生动，让观众赞不绝口。比如，《繁花》播出后，大家纷纷被游本昌饰演的爷叔这一角色深深折服。这位90岁高龄的老戏骨，仅通过一个眼神就能展现角色复杂的内心世界。正如网友留言所说的，他的表演赋予角色独特的魅力，堪称全剧的“定海神针”。\newline “二创”破圈，有梗有料。与主角相比，配角的戏份虽然相对较少，却拥有“留白”的空间，给“二创”提供了较大发挥空间。许多配角的“出圈”正是源于这些“再创作”的短视频。比如，《我的人间烟火》中男三号孟宴臣的“出圈”，正是从B站一个“二创”视频开始的，像孟宴臣反手开车门、蝴蝶墙前凝望等“名场面”，也被剪辑成短视频在社交平台大量传播。\newline 当影视剧中的配角在“二创”作品里“晋升”为主角，观众对角色的想象和期待在“二创”中得到满足，有时一句台词甚至一个饱含情绪的眼神就可能让配角脱颖而出。\newline 二\newline 配角“出圈”受欢迎，对作品来说本是件好事。但当配角人气大幅超越主角，演变成“配角掀桌”，似乎就偏离了影视剧创作者的初衷，也引发不少忧虑。\newline 在笔者看来，当我们讨论配角“上桌”或是“掀桌”时，实则默认了这个“桌”只能是主角的。而如今，演技好、业务精的配角们有更多机会成为观众心目中的“主角”，这便是在呼唤“以业务论英雄”的良性演艺生态。对于观众而言，能看见更多凭本事吃饭、与角色融为一体的优秀演员，这才是真正的“喜”，无关主角还是配角。\newline 配角之所以能够“逆袭”，有时候得益于主角的“衬托”。当主角的演技、人设或者剧情设计不够出色时，配角便有机会在对比中凸显出来。比如，在一些影视剧中，主演的演技稍显生硬，而戏份又非常多，那么在观众的“火眼金睛”下，这些缺点就会被放大、受到批评。还有的时候，主角表现已经达到了“及格分”，但配角的表演更加出彩，甚至盖过了主角，哪怕只有一两集或者十几分钟的出场，也已经足够凭借其鲜明的个性和深入人心的表演吸引观众注意。\newline 如今的观众不再满足于脸谱化的主角设定，而更加关注角色的多样性和真实性。这“倒逼”影视剧创作者创新人物塑造方式，赋予每一个角色独特的意义和价值。比如，电视剧《繁城之下》不单单讲述一个人，而是用一群人的故事来书写一个时代，从官差衙役到市井小民都拥有自己的鲜明特征。\newline 然而，在一些影视剧中，“主角光环”可谓照耀全场，似乎没有挑战可以难倒主角，导致剧情缺乏冲突和悬念。对于这样的“主角绝对压制”，很多观众可能会对“主角光环”产生审美疲劳，而将目光投向一些演技“在线”的配角。特别是随着阅历增加，一些观众逐渐感受到多数的人生剧本都不是“开了挂”的主角，进而“移情”配角。\newline 三\newline 配角火过主角的现象以及角色间的“争奇斗艳”也给影视行业带来一定启示。笔者想到三句话。\newline “角色无大小，全当正戏唱”。这句话源自京剧，意思是说不管什么角色，全情的投入、扎实的演技是获得观众喜爱的根本途径。细数近年来“出圈”的配角，他们各自戏份不一、人设不同，共同点都是用演技说话，让角色的“血肉”变得丰满，而出色的演技往往让角色和演员都闪闪发光、互相成就。\newline 比如电视剧《漫长的季节》的剧情虽然围绕主角王响展开，但剧中的不少人物都给人留下比较深刻的印象。像剧中镜头并不多的李巧云，“活脱脱”一个坚韧不拔的母亲形象，而傅卫军这一悲情角色虽然没有台词，但通过眼神和动作的表演就在观众心中激起波澜。\newline “把配角当主角写，把主角当人写”。没有一个角色的存在是毫无意义的，每个角色都有自己的舞台。“配角上桌”“配角掀桌”现象的出现，其实透露出观众对鲜活人物、动人故事的渴望。在流量时代，打磨剧本、提高作品质量更应是影视剧创作者的“必修课”，创作需要力求久久为功、精益求精，而非敷衍了事、粗制滥造。\newline 对编剧来说，打磨出更有深度和内涵的剧本，才能让每个角色有更强的生命力；对导演来说，则需要把握好整部剧的节奏和氛围，让每一个角色都能在适合的时刻展现出自己的魅力。正如一位青年编剧所说：“每个人物都有自己的一条命运线，要把配角当主角写，把主角当人写。”比如，《甄嬛传》之所以令人“久看不厌”，不仅得益于全员“演技在线”，还在于其剧情让主角和配角都有各自的闪光点，并衍生出值得重新解读的价值。\newline “与观众共情，而不是把他们劝退”。配角频频走红，也说明社会心态在悄然发生变化。随着观众审美的提高和影视市场竞争的加剧，套路化、模板化的剧情正在失去观众。正如有人说，这届观众很“逆反”，也拒绝被安排。现在有的古装剧，不仅演员演技浮夸，而且剪辑混乱、台词粗浅，分分钟就把观众劝退了，评分直线下降。因此，创作者们也需要跟上观众，呈现更多鲜活饱满、熠熠生辉的角色，启发观众对现实生活的理解和感悟。\newline 一位哲学家曾说：“人是目的而不是手段。”无论是主角还是配角，以全心全意、尽职尽责的态度去演绎，这不仅是对自己的尊重，也是对观众的尊重，唯有这样，才能塑造经典、收获认可。 \end{CJK} \\ \bottomrule
    \end{tabular}
    \caption{Example for \textbf{Open} in Chinese.}
    \label{ex:open}
\end{table*}

\begin{table*}[!ht]
\scriptsize
    \centering
    \begin{tabular}{p{40pt}p{380pt}}
    \toprule
    \textbf{Setting}&\textbf{Prompt}\\
    \midrule
    \textbf{Instruction} & Please write an article analyzing the phenomenon of "supporting characters taking center stage" in film and television works. Discuss the reasons and impacts behind the rising popularity of supporting roles, and explain the insights this phenomenon offers for film and television creation. The central idea is that the popularity of supporting characters reflects the improvement of audience aesthetics and their pursuit of high-quality film and television productions. \\ \midrule
    \textbf{Information} &  \\ \midrule
    \textbf{Reference} & \tiny In recent years, the phenomenon of “supporting roles taking the center stage” in films and TV dramas has become increasingly prominent, evidently turning into a new trend in the entertainment industry. Related topics have repeatedly made headlines and sparked widespread audience interest and discussion.\newline  
Characters like Chen Shuting and Li Youtian in *The Knockout*, Ye Bingshang in *Till The End Of The Moon*, and Xiang Liu in *Lost You Forever*… These supporting roles are no longer mere “decorations” within the story; instead, they often leave a deep impression on the audience. This phenomenon has been vividly termed “supporting roles taking the center stage,” and when the popularity of supporting roles exceeds that of the protagonists, it escalates into “supporting roles flipping the table.”\newline  
This inevitably raises curiosity: how are supporting roles able to rise and “break through”? Is the phenomenon of “supporting roles taking the center stage” a blessing or a concern?\newline  
**One**\newline  
Although “supporting roles taking the center stage” is a relatively new term, this phenomenon has long existed in the entertainment industry. “Digging into” older dramas would reveal that supporting characters like Bai Xiuzhu in *Romance in the Rain*, Gu Xichao in *The Story of a Noble Family*, Shangguan Haitang in *The Legend of the First*, and Wang Manchun in *The Disguiser* have all gained substantial audience appreciation.\newline  
In recent years, however, this phenomenon seems to have become even more prevalent, with the influence of many supporting role actors significantly increasing. In particular, 2023 has been dubbed by netizens as “The Year of Supporting Roles.” So, what has contributed to supporting roles breaking through and taking the spotlight? What are the “tactics” involved?\newline  
Excellent scripts often play a crucial role in making supporting roles shine. To satisfy the increasingly discerning audience, screenwriters are putting more effort into crafting compelling supporting roles. Supporting characters are no longer simply “good people” or “bad people” but are instead multi-dimensional individuals with rich backstories, unique perspectives, and layered personalities. Their designs are becoming more impactful and tension-filled, with diversified settings that evoke audience empathy. For instance, in *The Knockout*, the screenwriters meticulously created “comparative pairs” for both protagonists and supporting roles—such as An Xin and Li Xiang or Meng Dehai and An Changlin—to emphasize the dynamics of how some people stay true to their principles while others become lost amidst the turbulent times. This not only enhances dramatic conflict and impact but also gives each character their own fate and identity, deepening audience memory. As the screenwriter noted, “In Li Xiang, in Cao Chuang, and even in Cheng Cheng, there are discussions about their pursuits and destinies.”\newline  
The actors themselves also play a significant role, bringing solid performances and superior acting skills. By deeply understanding their roles, many supporting actors have delivered vivid and dynamic portrayals that leave audiences in awe. For example, after the release of *Blossoms Shanghai*, viewers were deeply moved by 90-year-old veteran actor You Benchang’s portrayal of Grandpa Shu. With just a glance, he conveyed the character’s complex inner world. As one comment online stated, his performance endowed the character with a unique charm and served as the “anchor” of the show.\newline  
Creative “secondary creations” (fan edits or reimaginings) further amplify the appeal of supporting roles. Unlike protagonists, who typically have more screen time and detailed arcs, supporting roles often have moments of “blank space,” leaving room for fan creatives to explore. Many supporting characters “break through” thanks to popular short videos made by fans. For instance, the breakout popularity of Meng Yancheng, the third male lead in *Fireworks of My Heart*, began with a viral video on Bilibili. Iconic moments such as Meng Yancheng’s reverse hand gesture to open a car door or his thoughtful gaze in front of the butterfly mural were edited into short clips and widely circulated on social media platforms.\newline  
When supporting roles ascend to “lead status” in secondary creations, viewers’ imaginations and expectations surrounding these characters are often fulfilled. Sometimes a single line of dialogue, or an emotionally charged glance, can make a supporting role stand out.\newline  
**Two**\newline  
Supporting roles “breaking through” and gaining popularity is undoubtedly a positive for the work as a whole. However, when the popularity of supporting roles significantly surpasses that of the protagonists and evolves into “supporting roles flipping the table,” it may deviate from the original purpose of film and TV creators, raising certain concerns.\newline  
In the author’s view, when discussing “supporting roles taking the center stage” or “flipping the table,” there’s an implicit assumption that the “table” is exclusively reserved for protagonists. Today, supporting actors with superior skills and high professionalism have more opportunities to become audiences’ “favorites,” reflecting a demand for a healthier entertainment ecosystem based on merit. For viewers, seeing more outstanding actors who earn recognition with their talent and fully embody their characters is truly a “win,” regardless of whether the performer is playing a lead or secondary role.\newline  
Supporting roles often rise due in part to the contrast provided by the protagonists themselves. When a protagonist’s acting, character design, or storyline fails to impress, supporting roles can stand out by comparison. For instance, in some dramas, main actors deliver overly stiff performances or dominate the screen time excessively, magnifying their shortcomings under the audience’s critical gaze. In other cases, while the protagonist’s performance is “passable,” a supporting role’s exceptional portrayal can overshadow them. Even a brief appearance of just one or two episodes or several minutes may suffice for a supporting role to captivate audiences with their distinct personality and memorable acting.\newline  
Today’s audiences are no longer satisfied with cookie-cutter protagonists; instead, they seek diverse and realistic characters. This “reverse pressure” pushes creators to innovate character development and endow every role with unique meaning and value. For example, *Beneath The City’s Light* doesn’t only tell one person’s story but uses the experiences of a group to portray an era—every character, from government officers to ordinary citizens, has distinct traits.\newline  
However, in certain dramas, the “protagonist halo” overwhelms the scene. It may seem as though no challenge is insurmountable for the lead, resulting in a lack of conflict and suspense in the storyline. This type of “absolute protagonist dominance” often leads to audience fatigue with such setups, prompting them to shift focus toward well-acted supporting roles. Particularly as viewers grow more mature, they come to realize that most life paths aren’t “cheat-mode” protagonist scripts, thus leading them to empathize more with supporting roles.\newline  
**Three**\newline  
The phenomenon of supporting roles eclipsing protagonists, as well as the competition between characters to “shine,” provides significant lessons for the entertainment industry. Three phrases come to the author’s mind.\newline  
“Every role matters; each deserves full effort.” Originating from Chinese opera, this saying highlights that regardless of the size of a role, wholehearted dedication and solid acting are the keys to earning audience love. Recent “breakout” supporting roles differ in their screen time and character designs but share one commonality: masterful performances that breathe life into characters, making them compelling. Great acting often elevates both the character and the actor, bringing mutual success.\newline  
For instance, while the story in *The Long Season* primarily revolves around the protagonist Wang Xiang, several other characters leave enduring impressions on the audience. From Li Qiaoyun’s portrayal of an indomitable mother to Fu Weijun’s tragic character, their performances, even with few lines, create notable emotional ripples through expressions and gestures.\newline  
“Write supporting roles as main characters; write protagonists as flawed humans.” No role exists without purpose; each character deserves their own spotlight. The rise of phenomena like “supporting roles taking the center stage” reveals audiences’ eagerness for lifelike characters and engaging narratives. In an age of excessive focus on social media metrics, crafting deeper scripts and improving production quality must remain essential “courses” for creators. Achieving excellence demands sustained effort and perfection rather than rushed, low-quality outputs.\newline  
For screenwriters, creating multi-dimensional characters with depth ensures that roles possess strong vitality. For directors, orchestrating the narrative flow and atmospheric tone allows every role to shine at the right moment. As one young screenwriter emphasized, “Every character follows their fate line; treat supporting roles as leads and protagonists as humans.” The timeless appeal of *Empresses in the Palace* stems not only from universally strong performances but also from its capacity to endow both main and supporting characters with distinctive traits, creating content deserving of constant reinterpretation.\newline  
“Connect with audiences through empathy, not alienation.” The popularity of supporting roles also reflects shifts in societal attitudes. Heightened audience expectations and fierce market competition mean formulaic narratives and stereotypical templates are losing public favor. As someone remarked, today’s viewers are rebellious and resist being spoon-fed. Some costume dramas feature exaggerated acting, chaotic editing, and shallow dialogue, quickly alienating viewers and leading to plummeting ratings. Thus, creators must align with audience values by presenting vibrant, multi-layered characters capable of inspiring insights into real-life experiences.\newline  
A philosopher once said, “Humans are ends, not means.” Whether portraying protagonists or supporting roles, a wholehearted, responsible performance is a sign of respect—not only for oneself but also for the audience. Only with this mindset can timeless classics be created, gaining genuine recognition and enduring appreciation. \\ \bottomrule
    \end{tabular}
    \caption{Example for \textbf{Open} in translated to English using GPT-4.1-2025-0414.}
    \label{ex:open_en}
\end{table*}

\section{LLM usage in this paper}

ChatGPT and Gemini are used in the preparation of this work as a general-purpose assistance tool. Specifically, they are employed in the following ways:
\begin{itemize}
    \item \textbf{Translation Assistance}: Converting expressions and sentences from the author’s native language into English.
    \item \textbf{Language Polishing and Grammar Revision}: Improving clarity, fluency, and grammatical correctness of the text, and ensuring that phrasing is natural in academic English.
    \item \textbf{Draft Review and Critique}: Providing feedback on drafts, including identifying unclear passages, suggesting improvements in structure, and flagging potential ambiguities.
\end{itemize}

They are not used for generating original research ideas, performing data analysis, or writing substantive technical content. All core research contributions, results, and argumentative structure were developed by the authors. The role of LLMs was limited to translation, linguistic polishing, and non-substantive editorial suggestions to improve presentation.

\end{CJK*}

\end{document}